\documentclass[a4paper,fleqn,preprint]{cas-dc}

\usepackage[utf8]{inputenc}
\usepackage{textcomp}
\usepackage{amsmath}
\usepackage{amssymb}
\usepackage[numbers]{natbib}
\usepackage{indentfirst}
\usepackage{xpatch}
\usepackage{orcidlink}
\usepackage{comment}
\usepackage{caption}
\usepackage{afterpage}
\usepackage{supertabular}
\usepackage{geometry}
\usepackage{microtype}
\usepackage{hyperref}
\hypersetup{
    colorlinks=true,
    linkcolor=[RGB]{70,130,180},
    urlcolor=[RGB]{70,130,180},
    citecolor=[RGB]{70,130,180},
    bookmarks=true,
    bookmarksopen=true,
    pdfstartview={FitH},
    breaklinks=true
}

\usepackage{longtable}
\usepackage{multirow}
\usepackage{makecell}
\usepackage{booktabs}
\usepackage{array}
\newcolumntype{L}{l}
\newlength{\tblwidth}
\usepackage{graphicx}
\usepackage{stfloats}
\usepackage{float}
\usepackage[section]{placeins}
\usepackage[justification=raggedright,labelfont=bf]{caption}

\usepackage{lineno}

\usepackage{xcolor}
\usepackage{ulem}  
\newcommand{\minew}[1]{{\color{black}{#1}}}
\newcommand{\miold}[1]{\iffalse{#1}\fi}

\makeatletter
\def\ps@pprintTitle{%
  \let\@oddhead\@empty
  \let\@evenhead\@empty
  \let\@oddfoot\@empty
  \let\@evenfoot\@odd@foot
  \def\@oddhead{\footnotesize\textit{Revised Manuscript (with Changes Marked)}\hfill}%
  \let\@mkboth\markboth
}
\makeatother

\makeatletter
\def\fnum@figure{Fig.~\thefigure}
\makeatother

\makeatletter
\let\oldsubsection\subsection
\renewcommand\subsection{\@afterindenttrue\oldsubsection}
\renewcommand\subsection{\@startsection{subsection}{2}{\z@}%
  {-3.25ex\@plus -1ex \@minus -.2ex}%
  {1.5ex \@plus .2ex}%
  {\normalfont\normalsize}}

\let\oldsubsubsection\subsubsection
\renewcommand\subsubsection{\@afterindenttrue\oldsubsubsection}
\renewcommand\subsubsection{\@startsection{subsubsection}{3}{\z@}%
  {-3.25ex\@plus -1ex \@minus -.2ex}%
  {1.5ex \@plus .2ex}%
  {\normalfont\normalsize}}
\makeatother

\usepackage{xspace}

\newcommand{\eg}{\textit{e.g.}\xspace}
\newcommand{\etal}{\textit{et al.}\xspace}

\newcommand{\etc}{\textit{etc.}\xspace}

\makeatletter
\def\changeBibColor#1{%
  \in@{#1}{Z-PNN
  lambda-PNN,10102101,8894479,9153037,liu2020psgan,zhou2022unified,XU202331,Lin_Dong_Ding_Liu_Liu_2024,GAN,MA2020110,10552904
  10477939,10506713,LIU2023292
  }
  \ifin@\color{black}\else\normalcolor\fi
}
 
\xpatchcmd\@bibitem
  {\item}
  {\changeBibColor{#1}\item}
  {}{}
 
\xpatchcmd\@lbibitem
  {\item}
  {\changeBibColor{#2}\item}
  {}{}
\makeatother

\makeatletter
\def\@cormark#1{%
  \ifnum#1=2\textsuperscript{$\dagger$}%
  \else\textsuperscript{*}%
  \fi}
\makeatother

\begin{document} 

\begin{sloppypar}
\let\WriteBookmarks\relax
\def\floatpagepagefraction{1}
\def\textpagefraction{.001}
\shorttitle{A Decade of Deep Learning for Remote Sensing Spatiotemporal Fusion: Advances, Challenges, and Opportunities}
\shortauthors{E. Sun, et~al.}

\title [mode = title]{A Decade of Deep Learning for Remote Sensing Spatiotemporal Fusion: Advances, Challenges, and Opportunities}

\author[1]{Enzhe Sun}[style=chinese]\cormark[2]
\ead{senzhe1113@hotmail.com}
\address[1]{School of Computer Science, China University of Geosciences (Wuhan), Wuhan 430078, China}

\author[2,3]{Yongchuan Cui}[style=chinese]\cormark[2]
\ead{yongchuancui@gmail.com}
\address[2]{Aerospace Information Research Institute, Chinese Academy of Sciences, Beijing 100094, China}
\address[3]{School of Electronic, Electrical and Communication Engineering, University of Chinese Academy of Sciences, Beijing 101408, China}

\author[2,3]{Peng Liu}[style=chinese,orcid=0000-0003-3292-8551]\cormark[1]
\ead{liupeng202303@aircas.ac.cn}

\author[1]{Jining Yan}[style=chinese,orcid=0000-0003-0680-5427]\cormark[1]
\ead{yanjn@cug.edu.cn}

\cortext[cor1]{Corresponding author}
\cortext[cor2]{These authors contributed equally to this work.}

\begin{abstract}
    \minew{Remote sensing spatiotemporal fusion (STF) addresses the fundamental trade-off 
    between temporal and spatial resolution by combining high temporal-low spatial 
    and high spatial-low temporal imagery. This paper presents the 
    first comprehensive survey of deep learning advances in remote sensing STF over 
    the past decade. We establish a systematic taxonomy of deep learning architectures 
    including Convolutional Neural Networks (CNNs), Transformers, Generative Adversarial 
    Networks (GANs), diffusion models, and sequence models, revealing 
    significant growth in deep learning adoption for STF tasks. Our analysis reveals 
    that CNN-based methods dominate spatial feature extraction, while Transformer 
    architectures show superior performance in capturing long-range temporal dependencies. 
    GAN and diffusion models demonstrate exceptional capability in detail reconstruction, 
    substantially outperforming traditional methods in structural similarity and spectral 
    fidelity. Through comprehensive experiments on seven benchmark datasets comparing 
    ten representative methods, we validate these findings and quantify the performance 
    trade-offs between different approaches. We identify five critical challenges: 
    time-space conflicts, limited generalization across datasets, computational 
    efficiency for large-scale processing, multi-source heterogeneous fusion, and 
    insufficient benchmark diversity. The survey highlights promising opportunities 
    in foundation models, hybrid architectures, and self-supervised learning approaches 
    that could address current limitations and enable multimodal applications. 
    The specific models, datasets, and other information mentioned in this article 
    have been collected in: \url{https://github.com/yc-cui/Deep-Learning-Spatiotemporal-Fusion-Survey}.
    }
\end{abstract}


\maketitle

\section{Introduction}\label{sec:intro}  
\begin{figure}[t]
    \centering
    \includegraphics[width=.42\textwidth]{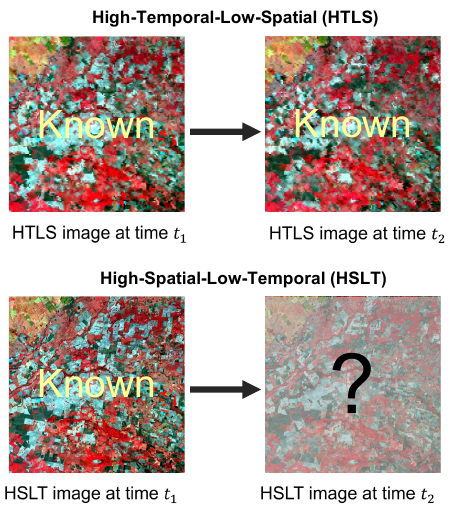}
    \caption{Fusion of remote sensing images.}
    \label{fig:fusion}
\end{figure}

\minew{
Spatiotemporal fusion (STF) integrates data with different spatial and temporal resolutions to generate high-quality imagery 
with enhanced spatiotemporal characteristics. STF has proven versatile across multiple domains: in computer vision for video 
analysis and action recognition~\cite{feichtenhoferConvolutionalTwoStreamNetwork2016}, in urban planning for traffic flow 
integration and congestion prediction~\cite{niuNovelSpatioTemporalModel2019}, and in medical imaging for visualizing dynamic 
processes such as tumor growth through multi-temporal image fusion~\cite{dafniroseLungCancerDiagnosis2021}.

In remote sensing (RS), STF addresses the inherent trade-off between spatial and temporal 
resolution in Earth observation. As illustrated in \autoref{fig:fusion}, STF combines high 
temporal but low spatial resolution (HTLS) data with high spatial but low temporal resolution 
(HSLT) data to generate synthetic imagery maintaining both characteristics~\cite{tan2022flexible}. 
While \autoref{fig:fusion} shows a typical example using one pair of HTLS-HSLT images from a 
previous time point and an HTLS image from the target time point, STF methods can flexibly 
incorporate varying numbers of input pairs—some approaches utilize three or more temporal 
pairs to enhance fusion accuracy. This capability is crucial for land surface monitoring, 
environmental research, and agricultural management.
}

\minew{
The fundamental challenge in STF stems from the spatiotemporal trade-off~\cite{liuRemoteSensingData2022a}. HTLS sensors like 
MODIS provide frequent temporal snapshots with reduced spatial resolution, while HSLT imagery offers detailed spatial information 
with limited temporal coverage. This inherent contradiction makes generating fused products that maintain both high temporal and 
spatial resolution challenging~\cite{belgiu2019spatiotemporal,walterskirchenTakingTimeSeriously2024}. STF techniques address 
this by combining these complementary data sources to produce enhanced products that capitalize on their respective strengths 
(\autoref{fig:fusion}).
}

Traditional STF methods can be systematically classified based on their underlying 
mathematical modeling principles~\cite{zhu2010enhanced,liuRemoteSensingData2022a}, 
which fall into five main categories:

\noindent\textbf{Bayesian-based methods.} The Bayesian fusion framework accounts for the temporal correlations in 
  image time series and employs 
  a maximum a posteriori estimator to produce the fused predictions. Representative implementations include the Bayesian Maximum Entropy 
  model~\cite{liBlendingMultiresolutionSatellite2013} and the unified fusion model~\cite{huangUnifiedFusionRemotesensing2013}. In practice, 
  the accuracy of such probabilistic fusion methods is assessed by co-registering the outputs with reference high-resolution imagery in both 
  space and time, and then computing conventional error metrics (e.g., RMSE, MAE). Moreover, the reliability of the predicted uncertainty distributions 
  is evaluated by comparing nominal confidence intervals against their empirical coverage rates, and cross-validation schemes are employed to ensure 
  the stability of the results. Bayesian approaches can effectively accommodate heterogeneous landscapes and thus are particularly valuable in 
  dynamic and complex applications such as vegetation monitoring and climate change 
  analysis~\cite{belgiu2019spatiotemporal, xiao2023review, li2020overview, zhuSpatiotemporalFusionMultisource2018}.

\noindent\textbf{Unmixing-based methods.} These methods estimate the high-resolution 
   pixel values by decomposing low-resolution pixels into endmembers based on linear 
   spectral mixing theory~\cite{zhuSpatiotemporalFusionMultisource2018,
   jiangUnmixingBasedSpatiotemporalImage2022,huang2014spatio}. 
   Well-known models include MMT (Multiresolution Spectral-Matching 
   Technique)~\cite{zhukovUnmixingbasedMultisensorMultiresolution1999} and others.
   Although they are limited by the linear assumptions in heterogeneous 
   areas, these approaches are useful for sub-pixel analysis, especially in 
   multisource remote sensing data fusion~\cite{li2020overview,xiao2023review,
   huang2021enhanced}.

\noindent\textbf{Learning-based methods.} These methods aim to represent the 
   relationship between coarse and fine resolution images, predicting the final 
   high-resolution image based on patterns learned from previous observations~\cite{
   huangSpatiotemporalReflectanceFusion2012}. Learning-based approaches 
   establish models that simulate the relationship between images of different 
   resolutions, capturing features that may not be directly observed in the 
   final images. These methods can effectively process images with similar 
   characteristics to those in the training set, automatically extracting
   features from large datasets~\cite{liuFastAccurateSpatiotemporal2016}.

\noindent\textbf{Weight function-based methods.} These methods estimate high-resolution 
   pixel values by combining information from multiple input images using weighted 
   functions. Prominent techniques include STARFM (Spatial and Temporal Adaptive 
   Reflectance Fusion Model)~\cite{fenggaoBlendingLandsatMODIS2006a},
   ESTARFM (Enhanced STARFM)~\cite{zhuEnhancedSpatialTemporal2010a}, STAARCH 
   (Spatiotemporal Adaptive Reflectance Correction for High-Resolution RS 
   Images)~\cite{hilker2009new}, and others. These methods perform well in homogeneous 
   areas but are less effective when dealing with nonlinear changes or complex 
   terrain~\cite{zhuSpatiotemporalFusionMultisource2018,belgiu2019spatiotemporal}.

\noindent\textbf{Hybrid methods.} Hybrid methods enhance fusion performance by integrating 
   advantages from different techniques. For example, Flexible Spatiotemporal 
   DAta Fusion (FSDAF)~\cite{zhu2016flexible} combines the strengths of unmixing 
   and weight function approaches to process complex spatiotemporal data, significantly 
   improving fusion accuracy and environmental adaptability, especially suitable for 
   high-resolution RS image processing scenarios~\cite{li2020overview,xiao2023review}.

\minew{
Traditional STF methods, despite their advancements, face significant limitations that 
constrain their practical applicability~\cite{lianRecentAdvancesDeep2025a}. Specifically, 
unmixing-based methods are constrained by linear spectral mixing assumptions that often 
fail in heterogeneous landscapes and require prior classification, greatly limiting their 
applicability~\cite{lianRecentAdvancesDeep2025a,zhuSpatiotemporalFusionMultisource2018}. 
Weight function-based methods rely heavily on prior knowledge and predefined spatial 
similarity assumptions, resulting in reduced stability and poor performance when these 
assumptions are violated~\cite{lianRecentAdvancesDeep2025a,belgiu2019spatiotemporal}. 
Bayesian methods suffer from prohibitive computational costs when processing large-scale 
or high-resolution images, making them impractical for operational applications requiring 
near real-time processing~\cite{lianRecentAdvancesDeep2025a}. Learning-based traditional 
methods depend on complex hand-crafted features that are time-consuming to design and 
often fail to capture the full complexity of spatiotemporal relationships, exhibiting 
poor stability across diverse datasets~\cite{lianRecentAdvancesDeep2025a}. Hybrid methods, 
while attempting to combine strengths of multiple approaches, increase computational 
complexity and parameter tuning difficulty, often resulting in error propagation through 
the fusion pipeline~\cite{lianRecentAdvancesDeep2025a}. These limitations collectively 
demonstrate that traditional methods struggle with computational efficiency, scalability 
to large datasets, adaptability to diverse environmental conditions, and the ability to 
capture complex nonlinear spatiotemporal relationships, thereby motivating the adoption 
of deep learning approaches~\cite{zhuSpatiotemporalFusionMultisource2018,belgiu2019spatiotemporal}.
}

\minew{
The application of deep learning (DL) technology in RS STF research has attracted considerable attention in recent years, arising 
from the rapid coordinated development of RS technology and DL methods~\cite{zhu2016flexible}. Unlike traditional methods that often 
rely on manually designed features and assume simple surface change patterns~\cite{zhukovUnmixingbasedMultisensorMultiresolution1999,wang2018spatio,wu2015integrated}, 
DL offers distinct advantages through: (1) automatic extraction of multi-level spatiotemporal features via large-scale data training; 
(2) effective modeling of non-linear relationships in complex scenarios; (3) higher tolerance for data noise and missing information, 
generating more stable fusion images; and (4) better adaptability to multi-source, multi-modal RS data fusion requirements. 
These capabilities have established deep learning as a key driving force in spatiotemporal fusion 
technology development~\cite{li2022deeplearning,jiang2024cnntransformer}.
}

\begin{figure*}[t]
    \centering
    \includegraphics[width=.65\textwidth]{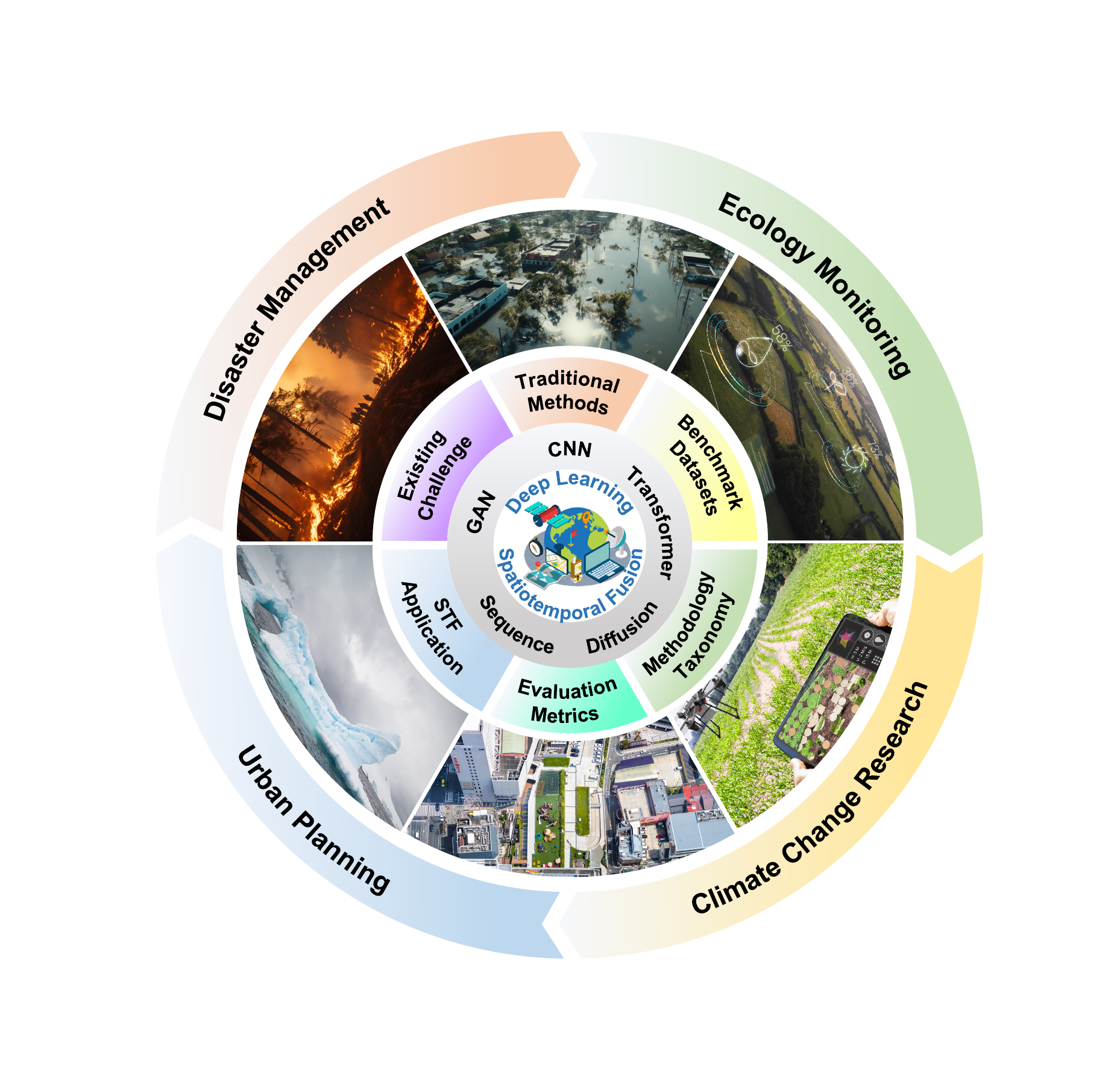}
    \caption{\minew{Overview of Deep Learning-based Remote Sensing Spatiotemporal Fusion
    framework and applications.} This figure illustrates the comprehensive landscape 
    covered in this review, including: (a) the network architectures examined (Convolutional Neural Network, 
    Transformer, Generative Adversarial Network, Diffusion, and sequence models, \textit{etc.}); (b) the thematic progression 
    of the paper sections from analysis of research trends to future opportunities; 
    (c) key application domains of spatiotemporal fusion technologies; and (d) visual representations 
    of the methodological framework. This graphical abstract serves as a roadmap for 
    understanding the decade-long evolution of deep learning approaches in remote 
    sensing spatiotemporal fusion discussed throughout the paper.}
    \label{fig:graphical_abstract}
\end{figure*}

This paper systematically reviews the application of deep learning in remote 
sensing spatiotemporal fusion over the past decade, highlighting the unique 
advantages and development potential of deep learning methods while analyzing 
current technical limitations and future development directions.
\autoref{fig:graphical_abstract} provides a visual framework illustrating this
survey's structure.

\subsection{Analysis of Research Trends of Deep Learning for 
Remote Sensing Spatiotemporal Fusion}\label{sec:trends}
\begin{figure}[htbp!]
    \centering
    \includegraphics[width=.5\textwidth]{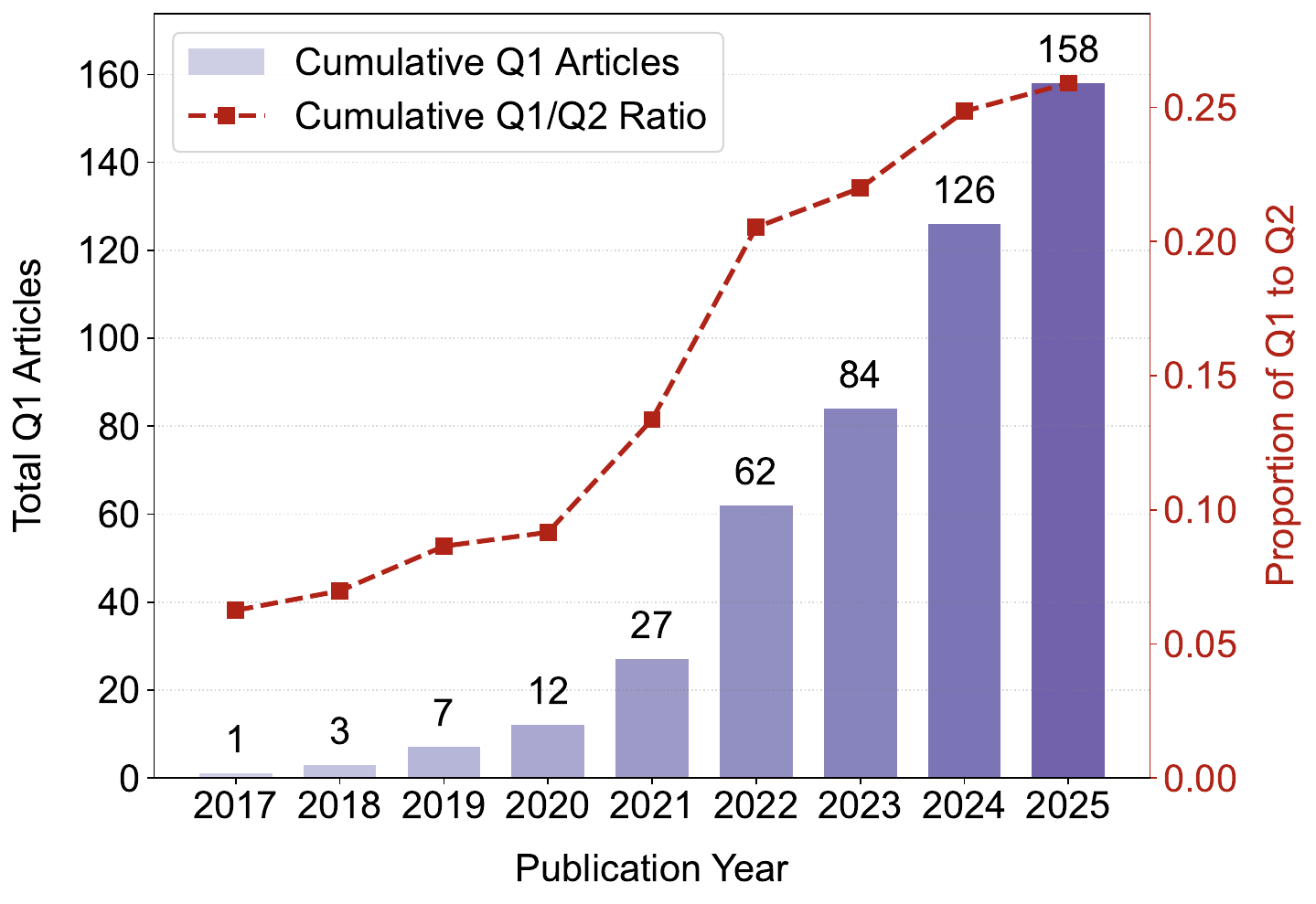}
    \caption{\minew{Number of published articles annually in Query 1 
    and its proportion on Query 2.}}
    \label{fig:trend_analysis}
\end{figure}

To visually demonstrate research trends and scope, we conducted 
statistical analysis of relevant literature in the Web of Science 
(WOS) database, collecting data on different research methods through 
two advanced searches, providing data support for exploring the 
evolution and future development trends in this field. Query Q1 was 
used to filter literature closely related to \textbf{\textit{deep 
learning}}, \textbf{\textit{remote sensing}}, and \textbf{\textit{
spatiotemporal fusion}} with the specific search topic: \texttt{(TS=
remote sensing) AND (TS=spatiotemporal fusion) AND (TS=deep learning)}. 
Query Q2 was broader and not limited to deep learning methods: 
\texttt{(TS=remote sensing) AND (TS=spatiotemporal fusion)}. 
For clarity, in the WOS database, \texttt{TS} refers to topic search, 
which searches for specified terms in titles, abstracts, author 
keywords, and keywords plus. Based on these two queries, we selected 
recent literature on these topics and conducted visualization analysis 
(see \autoref{tab:trends} and \autoref{fig:trend_analysis}).

\begin{table*}[htbp!]
\centering
\caption{Web of Science data retrieval results (2016-2025).}
\label{tab:trends}
\footnotesize
\begin{tabular*}{0.8\textwidth}{@{\extracolsep{\fill}} p{0.1\textwidth} 
p{0.6\textwidth} c @{}}
\toprule
\textbf{Query} & \textbf{Topic} & \textbf{Results} \\
\midrule
Q1 & (TS=remote sensing) AND (TS=spatiotemporal fusion) AND 
(TS=deep learning) & 158 \\
Q2 & (TS=remote sensing) AND (TS=spatiotemporal fusion) & 610 \\
\bottomrule
\end{tabular*}
\end{table*}

\autoref{fig:trend_analysis} shows that the cumulative number of Q1 
literature increased from 1 article in 2017 to 158 articles in 2025, 
with the ratio of Q1 to Q2 rising from 0.06 to 0.25. This indicates 
a significant growth in the proportion of deep learning methods in the 
spatiotemporal fusion field over the past ten years, highlighting its 
emergence as an important research direction in this domain.

\begin{figure*}[htbp!]
    \centering
    \includegraphics[width=\textwidth]{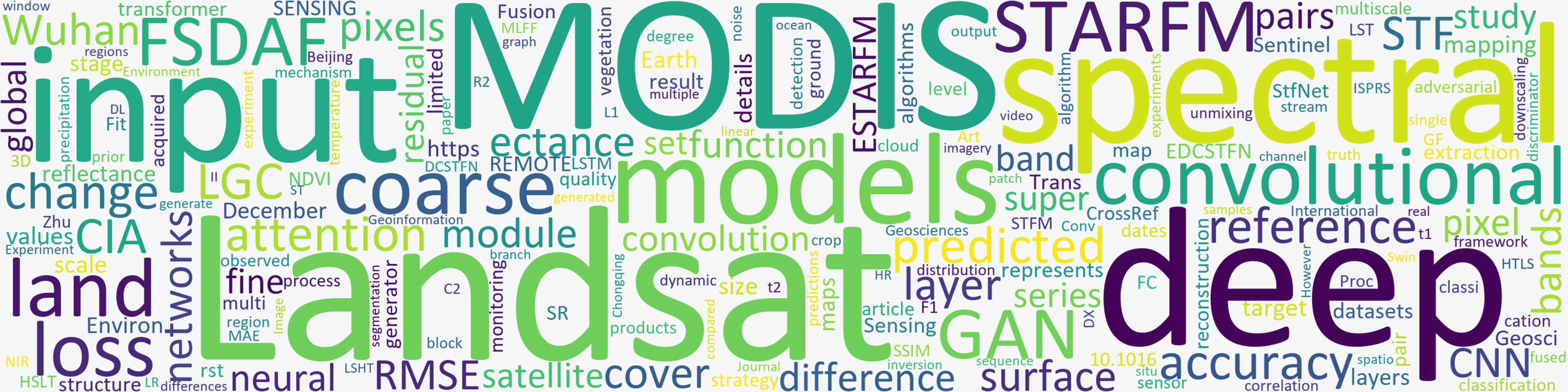}
    \caption{Word cloud visualization of keywords.}
    \label{fig:word_cloud}
\end{figure*}

To gain a more comprehensive understanding of research hotspots in DL for STF, we also generated a keyword word 
cloud using the all collected literatures, as shown in \autoref{fig:word_cloud}, from which high-frequency keywords 
and their distribution characteristics can be observed. The word cloud shows \textbf{\textit{Landsat}} and \textbf{\textit{MODIS}} as the most commonly used 
remote sensing data sources, highlighting their core role in STF research. 
The frequent appearance of deep learning-related technical terms underscores 
their importance in spatiotemporal fusion. \textbf{\textit{Deep}} and \textbf{\textit{Models}} reflect 
the widespread application of deep learning models in this field, while 
\textbf{\textit{Convolutional}} and \textbf{\textit{GAN}} demonstrate the application trends of 
convolutional neural networks~\cite{krizhevsky2012imagenet} and generative adversarial networks~\cite{goodfellow2014gan,arjovsky2017wasserstein}. In recent 
years, attention mechanisms and Transformer~\cite{vaswaniAttentionAllYou2017} models have gradually become new 
trends in STF research, as evidenced by related words like 
\textbf{\textit{Attention}} and \textbf{\textit{Transformer}}. Additionally, the word cloud highlights 
key research areas related to remote sensing image features, such as 
\textbf{\textit{Spectral}} and \textbf{\textit{Pixels}}, emphasizing the importance of multispectral 
features and pixel-level processing, while \textbf{\textit{Land}} and \textbf{\textit{Vegetation}} 
reflect research directions in STF concerning land cover and ecological 
monitoring. This word cloud demonstrates the extensive application of deep 
learning in spatiotemporal fusion and the rapid development of emerging 
technologies, clearly outlining the research framework and key directions 
in this field.

\subsection{Previous Surveys and Scope}\label{sec:surveys}
Previous surveys have played an important role in summarizing basic 
knowledge, technological developments, and typical applications in the remote 
sensing STF field~\cite{belgiu2019spatiotemporal}. However, most reviews primarily focus on traditional 
spatiotemporal fusion methods and have not comprehensively covered the latest 
advances in deep learning for remote sensing spatiotemporal fusion~\cite{li2022deeplearning}.

Zhu \etal~\cite{zhu2018spatiotemporal} systematically reviewed 
spatiotemporal fusion methods for multi-source remote sensing data and 
proposed a classification framework including several major categories such as 
regression methods, dictionary learning methods, and physical model methods, 
but did not deeply explore the potential of deep learning methods. Li \etal~\cite{li2020overview} 
focused on spatiotemporal fusion techniques for remote sensing data and 
conducted experimental comparisons of different models, but mainly concentrated 
on traditional machine learning methods, lacking detailed analysis of deep 
learning models that have emerged in recent years. Belgiu and Stein~\cite{belgiu2019spatiotemporal} mentioned 
some studies using neural networks but did not conduct in-depth analysis of 
these models' structures, data requirements, and application scenarios.

Lian \etal~\cite{lianRecentAdvancesDeep2025a} provided a detailed review of recent advances in 
deep learning-based remote sensing image spatiotemporal fusion methods.
The paper summarized the advantages and disadvantages of traditional
spatiotemporal fusion methods and analyzed the application of deep
learning models in spatiotemporal fusion, demonstrating the advantages
of these methods in improving fusion accuracy, processing efficiency,
and adaptability to complex scenarios. However, several shortcomings
remain in the work. The article lacks quantitative comparisons between
different methods; despite mentioning the pros and cons of various
approaches, it fails to provide detailed experimental data and
performance comparisons, making it difficult for readers to
comprehensively evaluate the performance of these methods across
different tasks and datasets. The description of current spatiotemporal
fusion methods' limitations in complex scenarios is rather simplistic,
without in-depth discussion of challenges that deep learning methods
face in practical applications, such as processing large-scale datasets,
high computational costs, and insufficient generalization capabilities.
While introducing various deep learning models, the analysis of specific
working mechanisms, training methods, and optimization strategies for
each method is relatively brief, lacking thorough discussion of
principles and implementation, which limits readers' deeper
understanding of these methods.

Despite covering different techniques and application scenarios of 
spatiotemporal fusion, existing reviews still have several limitations. 
\minew{They mostly focus on traditional methods, lacking systematic 
coverage of recent deep learning developments~\cite{xiao2023review,belgiu2019spatiotemporal}.} Additionally, only a few studies 
mention datasets and evaluation metrics for spatiotemporal fusion but fail to 
systematically compile commonly used datasets, evaluation frameworks, and 
related open-source code~\cite{li2022deeplearning}. Most critically, many new deep learning models 
introduced in recent years have not been comprehensively covered and discussed 
in existing reviews~\cite{wu2021land,zhu2018spatiotemporal,li2022deeplearning}.

\subsection{Our Contributions}\label{sec:contributions}
To provide a more comprehensive and systematic review of deep learning applications in 
remote sensing spatiotemporal fusion, we have completed the following work:

\begin{itemize}
   \item \textbf{Review of deep learning applications in remote sensing spatiotemporal fusion.} 
   We systematically review core deep learning models and their 
   application scenarios in spatiotemporal fusion. We examine these models' potential 
   in remote sensing image reconstruction, prediction, and enhancement through 
   analysis of practical cases.

   \item \textbf{Proposed classification framework and model timeline.} Based on existing 
   research, we construct a classification framework for deep learning 
   spatiotemporal fusion methods, categorized by network architecture, and present 
   the development trajectory of these models using timeline charts.

   \item \textbf{Summary of common datasets, evaluation metrics, and open-source code.} 
   We compile commonly used public datasets and evaluation metrics in the 
   spatiotemporal fusion field, provide comparative analysis, and catalog related 
   open-source code repositories.

   \item \textbf{Discussion of technical bottlenecks and future opportunities.} 
   We analyze technical bottlenecks faced by deep learning in remote sensing 
   spatiotemporal fusion and explore promising future research directions.
\end{itemize}

\subsection{Organization of This Survey}
To help readers efficiently navigate this review, 
we outline the organizational structure of this paper:
\textbf{Section \ref{sec:intro}} introduces basic concepts, importance, research trends, 
existing reviews, and the main contributions of this paper; \textbf{Section \ref{sec:advances}} 
reviews the latest research results, including benchmark datasets, evaluation metrics, 
and method classification; \textbf{Section \ref{sec:experiments}} presents comprehensive 
experimental comparisons of representative methods on multiple benchmark datasets; 
\textbf{Section \ref{sec:challenges}} discusses major technical bottlenecks and challenges; 
\textbf{Section \ref{sec:opportunities}} explores future research directions and potential 
applications; \textbf{Section \ref{sec:conclusion}} summarizes key points and provides 
a long-term outlook on the development of deep learning in remote sensing spatiotemporal fusion.

\autoref{tab:abbreviations} presents commonly used abbreviations in this survey 
with their explanations. These abbreviations cover methods, structures, and application 
scenarios fundamental to research in this field.

\begin{table*}[htbp!] 
\centering
\caption{Core abbreviations and explanations in remote sensing STF and deep learning.}
\label{tab:abbreviations}
\footnotesize
\begin{tabular*}{\textwidth}{@{\extracolsep{\fill}} p{0.15\textwidth} p{0.3\textwidth} p{0.15\textwidth} p{0.3\textwidth} @{}}
\toprule
\textbf{Abbreviation} & \textbf{Description} & \textbf{Abbreviation} & \textbf{Description} \\
\midrule
CNN & Convolutional Neural Network & LSTM & Long Short-Term Memory \\
CV & Computer Vision & MLP & Multi-Layer Perceptron \\
DEM & Digital Elevation Model & MTL & Multi-task Learning \\
DL & Deep Learning & OLI & Operational Land Imager \\
GAN & Generative Adversarial Network & RNN & Recurrent Neural Network \\
GNN & Graph Neural Network & RS & Remote Sensing \\
GRU & Gated Recurrent Unit & RSFM & Remote Sensing Foundation Model \\
HTLS & High Temporal Low Spatial & STF & Spatiotemporal Fusion \\
HSLT & High Spatial Low Temporal & ViT & Vision Transformer \\  
\bottomrule
\end{tabular*}
\end{table*}

\section{Advances}\label{sec:advances}
Recent years have witnessed significant advancements in \minew{spatiotemporal fusion (STF)} techniques, particularly through the integration of deep learning 
approaches. The field has evolved from traditional statistical methods to sophisticated neural network architectures, enabling 
more accurate and efficient fusion of remote sensing data across different spatial, temporal, and spectral resolutions.

\subsection{Key Trends}
\begin{figure*}[htb!]
    \centering
    \includegraphics[width=\textwidth]{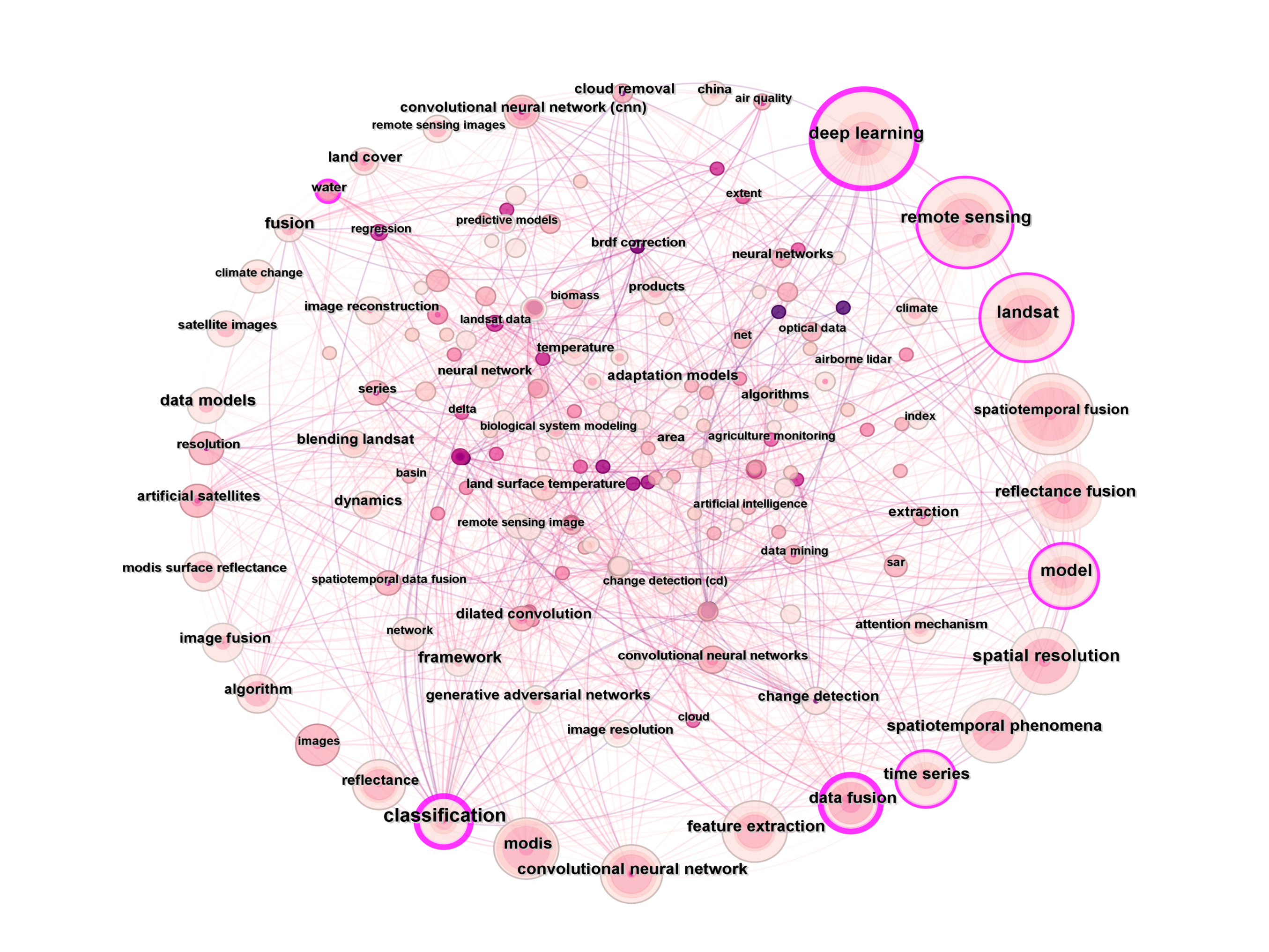}
    \caption{Keyword co-occurrence network.}
    \label{fig:keyword_network}
\end{figure*}

The analysis of literature volume and related word clouds clearly reveals the rising 
significance of deep learning in spatiotemporal fusion. By examining the keyword 
co-occurrence network created from relevant literature published between 2019 and 
2024, we gain deeper insights into core technologies, key trends, and application 
directions. As illustrated in \autoref{fig:keyword_network}, \textbf{\textit{spatiotemporal fusion}}, \textbf{\textit{deep learning}}, 
and \textbf{\textit{remote sensing}} emerge as the central concepts in the spatiotemporal 
fusion domain, while deep learning network architectures such as \textbf{\textit{Transformer}} 
and \textbf{\textit{GAN}} are being widely adopted throughout the field.

\begin{figure*}[!htbp]
    \centering
    \includegraphics[width=0.9\textwidth]{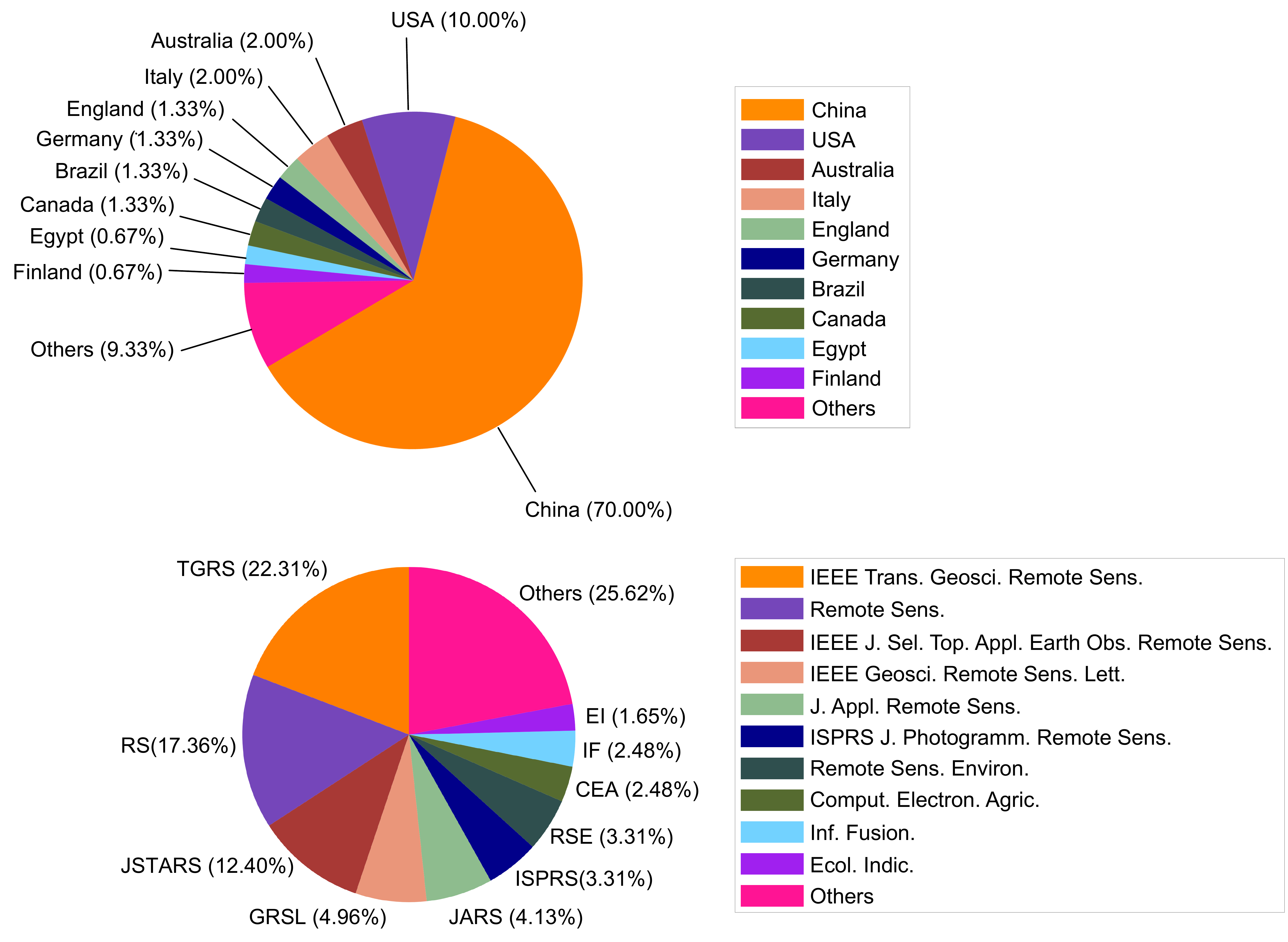}
    \caption{Distribution of published articles by top countries and journals.}
    \label{fig:country_journal_dist}
\end{figure*}

Looking at the distribution of author nationalities and relevant publications provides additional perspective 
on research contributions. As shown in \autoref{fig:country_journal_dist}, 
from a national contribution standpoint, China dominates 
this field, accounting for approximately 70\% of Q1 literature matching the search 
criteria (\texttt{(TS=``remote sensing'') AND (TS=``spatiotemporal fusion'') AND 
(TS=``deep learning'')}). The United States follows with roughly 10\% of publications, 
while countries like Australia, Italy, and the United Kingdom make up the remainder. 
This distribution clearly highlights the technological advantages and research 
foundations that China and the US possess in remote sensing and deep learning domains. 
The concentration of research efforts globally is accelerating technological 
development in this integrated field of remote sensing and deep learning.

From a journal distribution perspective, top-tier remote sensing publications have 
become the primary platforms for papers focused on deep learning and spatiotemporal 
fusion. \textit{IEEE Transactions on Geoscience and Remote Sensing} (TGRS) leads with the 
highest publication rate at 22.31\%, followed by \textit{Remote Sensing} and 
\textit{IEEE Journal of Selected Topics in Applied Earth Observations and Remote Sensing} (JSTARS). The 
concentration of papers in these high-impact journals, \eg, TGRS, \textit{Information Fusion} (IF), reflects both the academic value and technical depth of research in 
this field, while also indicating the increasingly specialized nature of studies in 
this domain.

\begin{figure*}[p]
\centering
\includegraphics[width=0.9\textwidth]{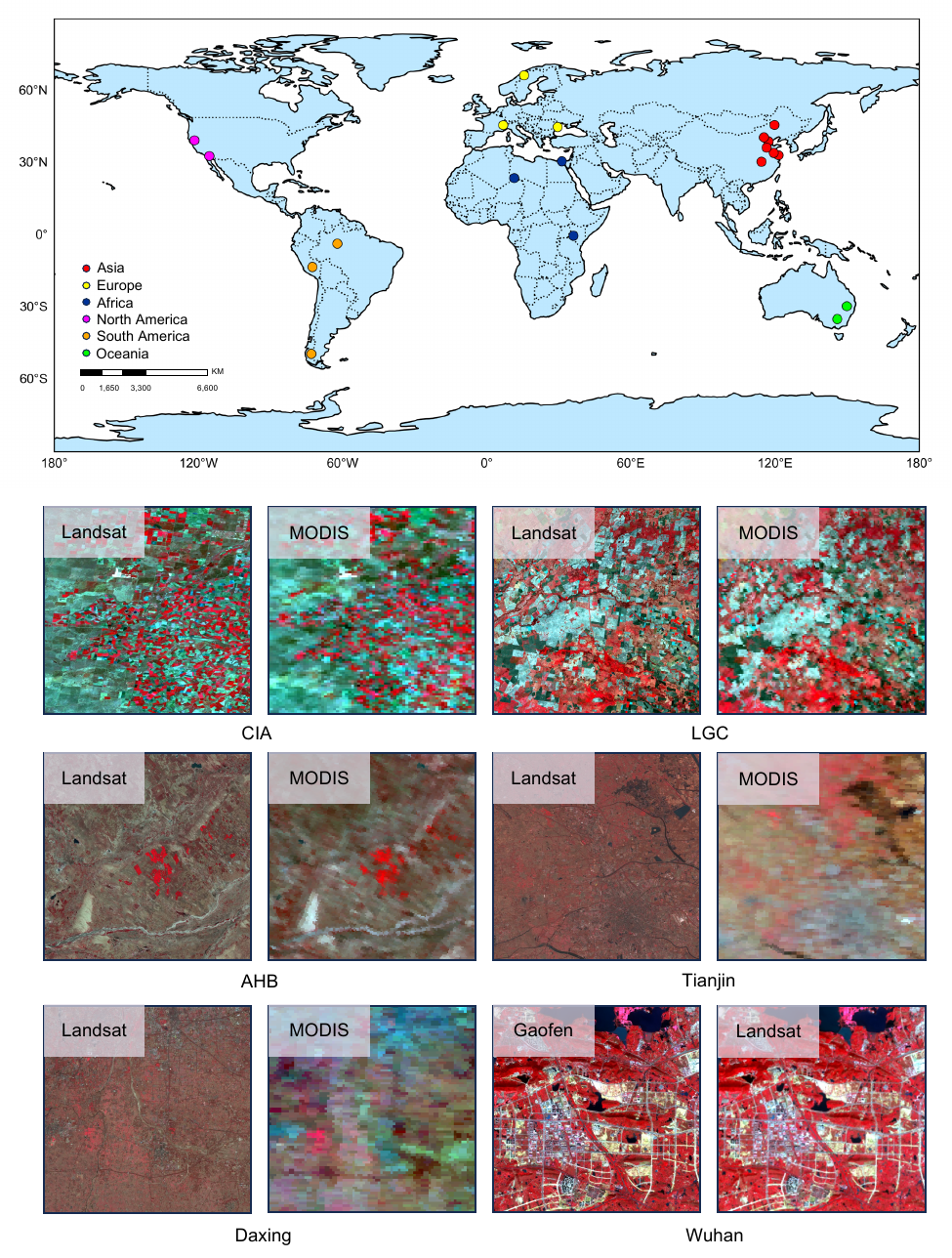}
\caption{Global distribution and data samples of spatiotemporal fusion datasets.}
\label{fig:dataset_dist}
\end{figure*}

\subsection{Benchmark Datasets}\label{subsec:benchmark_datasets}
Benchmark datasets play a crucial role in remote sensing spatiotemporal fusion research, 
providing unified standards for model training, evaluation, and performance comparison 
that are essential for assessing deep learning algorithms and promoting technological 
advancement~\cite{huang2024stfdiff,wang2024stinet,weng2024spatially}.
After more than a decade of development, a relatively complete dataset system has 
been established~\cite{mu2024spatiotemporal}, as summarized in \autoref{tab:datasets}.

These datasets display a clear hierarchical structure when viewed from their 
spatiotemporal characteristics~\cite{xie2024gan,xu2024hidri}. Terrestrial observation datasets typically employ 
complementary configurations of \textit{high spatial - low temporal} resolution (\eg, 
Landsat's 30m/16 days) and \textit{low spatial - high temporal} resolution (\eg, MODIS's 
500m/1 day), creating standardized frameworks for evaluating the spatiotemporal 
reconstruction capabilities of fusion algorithms. In contrast, oceanic observation 
datasets (\eg, GOCI-II) offer unique high temporal resolution (1 hour)~\cite{zhang2024spatiotemporal,jia2024forecasting}. 
As shown in \autoref{fig:dataset_dist}, the spatial distribution of existing benchmark datasets exhibits 
notable regional clustering, primarily concentrated in Asia and Oceania. The figure also 
presents image comparisons between Landsat and MODIS for five representative datasets 
(CIA~\cite{zhang2024enhanced}, LGC~\cite{ma2024conditional}, AHB, Tianjin~\cite{li2020overview}, and Daxing~\cite{li2020overview}), as well as between Gaofen and Landsat for the 
Wuhan~\cite{zhang2024dcdgan} dataset, collectively covering six distinct geographical regions.

\subsubsection{Regional Dataset Distribution}\label{subsubsec:regional_datasets}

\minew{Over years of extensive development, spatiotemporal fusion datasets have become globally distributed, covering diverse geographical 
scenarios and exhibiting a wide variety of land surface changes~\cite{cuiPansharpeningPredictiveFiltering2025}.}

\paragraph{Australia Datasets}
In Australia, the most widely utilized 
and representative datasets are CIA~\cite{zhang2024enhanced,ao2022deep,song2018spatiotemporal} and LGC~\cite{ma2024conditional,ao2022deep,song2018spatiotemporal}.
The CIA dataset is situated within the Coleambally Irrigation Area in southern New South Wales, Australia. It utilizes 
Landsat 7 ETM and MODIS (MOD09GA) imagery with spatial resolutions of 30 m and 500 m respectively. Comprising 17 image 
pairs captured between October 2001 and May 2002, each image measures 1720 × 2040 pixels. The dataset predominantly includes 
irregularly shaped irrigated fields, drylands, and forested areas, making it particularly suitable for evaluating the performance 
of fusion models in heterogeneous agricultural landscapes. The LGC dataset, located in the Lower Gwydir Catchment of northern New 
South Wales, Australia, incorporates Landsat 5 TM and MODIS (MOD09GA) images with identical spatial resolutions of 30 m and 500 m. 
It contains 14 cloud-free image pairs collected from April 2004 to April 2005, each with dimensions of 3200 × 2720 pixels. 
Due to frequent sudden flooding events, this dataset is well-suited for assessing fusion 
methodologies in dynamically evolving landscapes~\cite{li2022deeplearning,lei2024hpltsgan}.

\paragraph{China Datasets}
In China, currently prevalent benchmark datasets include AHB, TJ~\cite{li2020overview}, and DX~\cite{li2020overview}. The AHB dataset, located in Ar Horqin Banner, 
Inner Mongolia Autonomous Region, primarily captures agricultural and pastoral landscapes using Landsat 8 OLI (30 m) and 
MODIS (MOD09GA, 500 m) imagery~\cite{li2020overview}. It spans over five years, from May 2013 to December 2018, featuring 27 cloud-free image 
pairs. This extensive temporal coverage encapsulates pronounced phenological variations, making it ideal for examining fusion model 
efficacy in rural regions characterized by heterogeneous land-cover types. The TJ dataset, focused on Tianjin municipality, 
integrates Landsat 8 OLI (30 m) and MODIS (MOD02HKM, 500 m) data. With 27 pairs of cloud-free images collected between 
September 2013 and September 2019, this dataset systematically documents rapid urban expansion alongside seasonal vegetation changes, 
thus providing a robust benchmark for validating fusion methods in rapidly urbanizing environments. Similarly, 
the DX dataset pertains to Beijing's Daxing district, utilizing Landsat 8 OLI (30 m) and MODIS (MOD02HKM, 500 m) imagery. 
Comprising 29 cloud-free image pairs captured from September 2013 to November 2019, it effectively records substantial 
transformations in land cover during the construction period of Beijing Daxing International Airport, thereby serving as a 
valuable resource for assessing the performance of fusion approaches amid significant land-use transitions~\cite{xiao2023review, zhang2024enhanced}.

\paragraph{United States Datasets}
In the United States, two recent datasets, IC~\cite{guo2024obsum} and BC~\cite{guo2024obsum}, both composed of Sentinel-2 MSI (10 m) and Sentinel-3 OLCI (300 m) imagery, 
have emerged, each encompassing five cloud-free image pairs within their respective regions~\cite{guo2024obsum}. Specifically, the IC dataset is 
located in California's Imperial County, an area characterized by heterogeneous agricultural practices including cultivation 
of alfalfa, sugar beet, durum wheat, and onions. This region is marked by fragmented farmland parcels with rapidly shifting 
land-use dynamics, making it suitable for evaluating model performance within complex agricultural landscapes. Conversely, 
the BC dataset situated in southwestern Butte County, California, predominantly features rice cultivation areas. It exhibits 
clear seasonal land-cover variations, presenting a prototypical cyclic agricultural phenological scenario, thereby being 
particularly apt for testing fusion methodologies aimed at capturing periodic patterns of land surface change.

\subsubsection{Dataset Evolution and Trends}\label{subsubsec:dataset_trends}

\minew{Overall, the global spatiotemporal fusion datasets exhibit clear regional differences and geographical diversity. 
Initially, the datasets focused more on agricultural scenes, but they have gradually shifted towards urban landscapes. 
This undoubtedly reflects the recent trend in spatiotemporal fusion research, which is evolving from classic medium-resolution 
agricultural areas to more complex, finer, and higher spatial resolution urban and diverse landscapes.}

\subsubsection{Current Limitations and Future Directions}\label{subsubsec:dataset_limitations}

\minew{
However, the currently used benchmark datasets for spatiotemporal fusion have some limitations. 
Most benchmarks rely on Landsat–MODIS image pairs, but the spatial resolution gap between the two 
can be as large as sixteenfold, making some image textures virtually impossible to predict. 
Undoubtedly, this issue undermines the evaluation of a model's detail-recovery capability. 
In addition, cloud cover interference introduces errors during the fusion process. Although large 
cloud masses and noise can be manually removed, residual thin clouds, shadows, and striping 
still affect the fusion results~\cite{xiao2023review}.

To overcome these limitations, we can draw on the approach taken by Zhang and his team~\cite{zhang2024wuhan}. They selected 
a region with abundant surface features and significant change at the junction of Hongshan and 
Jiangxia Districts in Wuhan, China, and constructed the Wuhan benchmark dataset—a high-resolution 
spatiotemporal fusion dataset comprising GF (Gaofen) and Landsat imagery. This is the first 
dataset to combine high-resolution GF images (2~m panchromatic, 8~m multispectral) with 
Landsat-8 multispectral images (30~m), and it contains eight cloud-free image pairs spanning 
more than seven years. Unlike the traditional low-to-medium resolution MODIS–Landsat pairs, 
whose spatial resolution difference can reach sixteenfold, the GF–Landsat combination reduces 
this gap to between four and ten times, enabling a more realistic evaluation of a model's 
detail-recovery capabilities.
As for cloud cover interference, the following measures can be adopted: pre-screen remote sensing 
images for cloud cover below 10\% and select image pairs whose acquisition dates are identical 
or as close as possible; combine multi-source data (e.g., Sentinel-2, meteorological satellite 
cloud maps) and deep-learning cloud-removal algorithms to automate and improve the accuracy 
of cloud masking, thereby reducing the burden of manual screening~\cite{zhang2024wuhan, juAvailabilityCloudfreeLandsat2008}.
}

\renewcommand{\arraystretch}{1.3}  
\begin{table*}[htb!]
\caption{Common Spatiotemporal Fusion Datasets.}
\label{tab:datasets}
\small 
\begin{tabular*}{\textwidth}{@{\extracolsep{\fill}} m{0.13\textwidth} m{0.22\textwidth} m{0.2\textwidth} m{0.2\textwidth} m{0.1\textwidth} @{}}
\toprule
\textbf{Dataset} & \textbf{Source} & \textbf{Resolution} & \textbf{Region} & \textbf{Link} \\
\midrule
CIA~\cite{zhang2024enhanced} & Landsat, MODIS & 30m, 500m, 16 days & Coleambally, AU & \href{https://data.csiro.au/collections/collection/CIcsiro\%3A5846v3}{Link} \\

LGC~\cite{ma2024conditional} & Landsat-5 TM, MODIS & 30m, 500m, 16 days & Lower Gwydir, AU & \href{https://data.csiro.au/collections/collection/CIcsiro:5847v003}{Link} \\

MOD09GA~\cite{chen2023robot} & MODIS & 500m, Monthly & North China Plain & \href{https://ladsweb.modaps.eosdis.nasa.gov}{Link} \\

BC~\cite{guo2024obsum} & Sentinel-2 MSI, S3 OLCI & 10m, 300m, Monthly & SW Butte County, CA & \href{https://www.esa.int/Applications/Observing_the_Earth/Copernicus/Sentinel-2}{Link} \\

IC~\cite{guo2024obsum} & Sentinel-2 MSI, S3 OLCI & 10m, 300m, Monthly & Imperial County, CA & \href{https://www.esa.int/Applications/Observing_the_Earth/Copernicus/Sentinel-2}{Link} \\

OISST~\cite{kim2023multisource} & AVHRR, Buoy, Ship & 0.25° × 0.25°, Daily & Global Ocean & \href{https://www.ncei.noaa.gov/products/optimum-interpolation-sst}{Link} \\

OSTIA~\cite{kim2023multisource} & Multi-sat IR, MW, Buoy & 0.05° × 0.05°, Daily & Global Ocean & \href{http://marine.copernicus.eu/services-portfolio/access-to-products/}{Link} \\

G1SST~\cite{kim2023multisource} & Geo, Polar Sats & 0.01° × 0.01°, Daily & Global Ocean & \href{https://podaac.jpl.nasa.gov/dataset/JPL_OUROCEAN-L4UHfnd-GLOB-G1SST}{Link} \\

EARS~\cite{kim2023multisource} & ECMWF Model & 0.25° × 0.25°, Hourly & Global Ocean & \href{https://rda.ucar.edu}{Link} \\

In-situ~\cite{kim2023multisource} & Buoy Measurements & 16 Stations, Hourly & Korean waters & \href{http://www.nifs.go.kr/kodc}{Link} \\

TRMM~\cite{wu2020spatiotemporal} & NASA GSFC PPS & 0.25°, 3-hourly & 50°N--50°S & \href{https://pmm.nasa.gov/data-access/downloads/trmm}{Link} \\

GridSat~\cite{wu2020spatiotemporal} & NOAA & 0.07°, 3-hourly & 70°S--70°N & \href{https://www.ncdc.noaa.gov/gridsat/gridsat-index.php}{Link} \\

DEM~\cite{wu2020spatiotemporal} & USGS, NASA SRTM & 90m, N/A & 60°N--56°S & \href{http://srtm.csi.cgiar.org/srtmdata/}{Link} \\

Rain~\cite{wu2020spatiotemporal} & CMDC (China) & Point, 12-hourly & China & \href{http://data.cma.cn/}{Link} \\

S2~\cite{gu2024muddy} & Sentinel-2 & 10m, 5-day revisit & Dafeng, China & \href{https://scilb.copernicus.eu}{Link} \\

GOCI-II~\cite{gu2024muddy} & GOCI-II Satellite & 500m, hourly & Dafeng, China & \href{https://www.nosc.go.kr/}{Link} \\

Wuhan~\cite{zhang2024dcdgan} & GF, Landsat & 8m, 30m & Wuhan, China & \href{https://github.com/lixinghua5540/Wuhan-dataset}{Link} \\

Daxing~\cite{li2020overview} & LS8 OLI, MODIS & 30m, 500m, 8 days & Daxing, Beijing & \href{https://pan.baidu.com/disk/main#/index?category=all&path=\%2FDatasets}{Link} \\

AHB~\cite{li2020overview} & LS8 OLI, MOD09GA & 30m, 500m, 16 days & Ar Horqin Banner & \href{https://pan.baidu.com/disk/main#/index?category=all&path=\%2FDatasets}{Link} \\

Tianjin~\cite{li2020overview} & LS8 OLI, MOD02HKM & 30m, 500m, 16 days & Tianjin, China & \href{https://pan.baidu.com/disk/main#/index?category=all&path=\%2FDatasets}{Link} \\

Terra~\cite{chenTerraMultimodalSpatioTemporal2024} & Multi-source & 0.1°, 3-hourly & Global & \href{https://www.selectdataset.com/dataset/5813a4f2da636c4f1c24dac729fab5e0}{Link} \\
\bottomrule
\end{tabular*}
\end{table*}
\renewcommand{\arraystretch}{1.0}

\subsection{Methodology Taxonomy}
The introduction of deep learning technology in the research of remote sensing spatiotemporal 
fusion has significantly enhanced the models' ability to model complex spatiotemporal 
features~\cite{zhao2024cfformer,yu2025mgsfformer}. Based on network architecture and 
technical characteristics, existing methods can be classified into the following categories: 
Convolutional Neural Networks (CNNs)~\cite{kattenborn2021cnn}, 
Transformer~\cite{zhao2024enso}, Generative models~\cite{zhang2024dcdgan}, Sequence 
models~\cite{sun2022lunet}, and other innovative architectures (such as graph neural 
networks~\cite{shi2020pointgnn}, dual-branch fusion networks~\cite{sun2023dbfnet}, 
multi-layer perceptrons~\cite{chen2023stfmlp}, \etc). These models exhibit distinct 
features in terms of spatiotemporal modeling capabilities, performance optimization, and 
applicable scenarios, driving technological innovation in the remote sensing 
field~\cite{li2022deeplearning,li2020overview}.

\begin{figure*}[htbp!]
    \centering
    \includegraphics[width=1\textwidth]{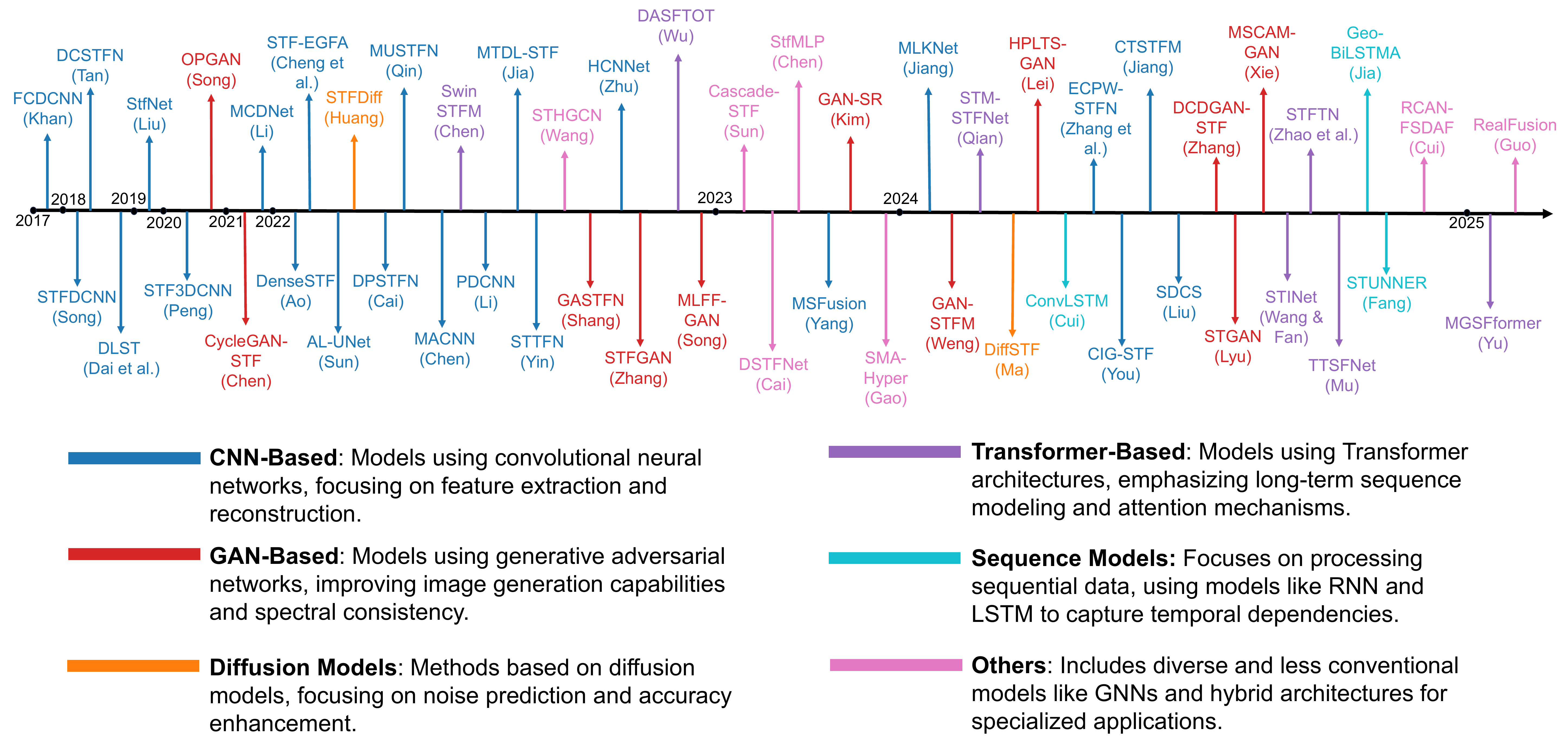}
    \caption{Timeline of deep learning-based spatiotemporal fusion methods and their classifications.}
    \label{fig:method_timeline}
\end{figure*}

As shown in \autoref{fig:method_timeline}, a summary of representative methods in remote 
sensing spatiotemporal fusion and their development over time is provided. Different colored 
timelines indicate representative models and corresponding years for CNN, GAN (Generative Adversarial Network), Diffusion, 
Transformer, sequence models, and other architectures~\cite{xiao2023review,zhu2018spatiotemporal}. 
From 2017 to 2025, the application of various network architectures in this field has shown 
trends of diversification and refinement~\cite{xiao2023review}. Meanwhile, 
\autoref{tab:model_characteristics} systematically summarizes the specific characteristics, 
loss functions, and performance metrics of different models, providing important references 
for in-depth analysis of the advantages, disadvantages, and usability of these methods.

Furthermore, \autoref{tab:model_comparison_long} presents a comprehensive comparison of deep learning models for 
spatiotemporal fusion, detailing their architectures, performance highlights, and limitations across various application 
scenarios, which offers crucial insights for researchers to select appropriate methods based on specific requirements.

\subsubsection{Convolutional Neural Networks}\label{subsubsec:cnn}

Convolutional Neural Networks (CNNs)~\cite{liSurveyConvolutionalNeural2022} have become the 
core architecture in spatiotemporal fusion research in remote sensing since their 
breakthrough in computer vision~\cite{maggioriConvolutionalNeuralNetworks2017}. The fundamental 
operation in CNNs is convolution, which can be mathematically expressed as:
\begin{equation}
y(i,j) = \sum_{m=0}^{M-1} \sum_{n=0}^{N-1} x(i+m, j+n) \cdot w(m,n),
\end{equation}
\noindent where $x$ represents the input image or feature map, $w$ is the convolution kernel, 
$y$ is the output feature map, $i$ and $j$ are the spatial coordinates in the output feature map,
$m$ and $n$ are the indices for the kernel elements, and $M$ and $N$ represent the height and width 
of the convolution kernel, respectively. 

\onecolumn
\minew{
\setlength{\LTleft}{0pt plus 1fill minus 1fill}
\setlength{\LTright}{0pt plus 1fill minus 1fill}

{\footnotesize  
\begin{longtable}{>{\raggedright}p{2cm}>{\raggedright}p{4cm}>{\raggedright}p{5cm}>{\raggedright\arraybackslash}p{4.5cm}}

\caption{Comparison of Deep Learning Models for Spatiotemporal Fusion} \label{tab:model_comparison_long} \\

\toprule
\textbf{Model} & \textbf{Architecture} & \textbf{Performance Highlights} & \textbf{Limitations} \\
\midrule
\endfirsthead

\toprule
\textbf{Model} & \textbf{Architecture} & \textbf{Performance Highlights} & \textbf{Limitations} \\
\midrule
\endhead

\multicolumn{4}{r}{\textit{Continued on next page}} \\ 
\endfoot

\bottomrule
\endlastfoot

STFDCNN~\cite{song2018spatiotemporal} & 
Two-stage CNN: NLM CNN + SR CNN + fusion & 
Superior metrics; Strong spatial heterogeneity; Excellent spectral preservation & 
Missing spatial details; Complex land changes; High complexity \\
\midrule

STFNet~\cite{liu2019stfnet} & 
Dual-stream 3-layer CNN; Temporal dependency; Joint reconstruction loss & 
Outperforms STARFM/FSDAF; 8-16x scaling; Temporal coherence & 
High computational cost; Shallow structure; Limited parameters \\
\midrule

EDCSTFN~\cite{tan2019enhanced} & 
Encoder-merge-decoder; LTHS encoder + residual; Composite loss & 
Image clarity; Flexible training; Reduced parameters; Enhanced robustness & 
Data quality sensitive; Poor reference quality issues; Complex training \\
\midrule

STF3DCNN~\cite{peng2020fast} & 
4D 3D CNN; Multi-dimensional dataset; No pooling design & 
Improved speed; Large-scale datasets; Efficient structure & 
Abrupt change issues; Irregular change problems; Overfitting prone \\
\midrule

STTFN~\cite{yin2021spatiotemporal} & 
Multi-scale fusion CNN; Slim-WDSR; Local/global residual & 
Superior accuracy; Nonlinear mapping; Spatiotemporal coherence & 
Training sample dependent; Two image pairs required; Complex tuning \\
\midrule

MCDNet~\cite{li2021multi} & 
Multi-collaborative CNN; SRCNN + DRCNN; Multi-scale perception & 
Collaborative networks; Edge preservation; Feature + content loss & 
Multi-network burden; Parameter adjustment; Task-specific design \\
\midrule

MOST~\cite{wei2021enblending} & 
Cascaded EDSR; Two-stage 4x upsampling; 16x super-resolution & 
Mosaicking application; Spectral consistency; Offline pre-training & 
Domain gap; GPU dependence; Transfer limitations \\
\midrule

MUSTFN~\cite{qin2022mustfn} & 
Multi-level CNN; Feature pyramid; Channel attention; Mask coefficients & 
Large change scenarios; Cloud-contaminated training; Adaptive pooling & 
Texture loss; Less flexible; Temporal interval sensitive \\
\midrule

DenseSTF~\cite{ao2022deep} & 
DenseNet-based; Patch-to-pixel; Dense blocks; Weight normalization & 
Spatial heterogeneity; Abrupt changes; Flood scenarios & 
GPU required; Long training; Hidden changes unpredictable \\
\midrule

STF-EGFA~\cite{cheng2022stfegfa} & 
Dual encoder-decoder; Edge extraction; Feature attention (CA+PA) & 
Various scenarios; Clear edges; Road/field boundaries & 
Many parameters; Complex tuning; Poor color prediction \\
\midrule

HCNNet~\cite{zhu2022hcnn} & 
2D/3D hybrid CNN; CBAM attention; Multi-band structure & 
Multiple scenarios; Strong feature extraction; Stable results & 
High overhead; Large parameters; Hardware demanding \\
\midrule

MACNN~\cite{chen2022multiscale_srs} & 
Dual-stream FSRCNN; ASPP multi-scale; Spatial-channel attention & 
Flood detection; Vegetation details; Multi-scale features & 
GPU dependent; Blurry regions; Training sample limitations \\
\midrule

ECPW-STFN~\cite{zhang2024enhanced} & 
Wavelet transform; Four modules; High/low frequency separation & 
Only 2 images required; Complex scenarios; Frequency separation & 
Limited improvement; Stability issues; Transfer capability unclear \\
\midrule

MLKNet~\cite{jiang2024mlknet} & 
Three-module dual-branch; TFNet + LAM + TFM; Large kernel attention & 
Urban/complex features; Global perception; Multi-scale fusion & 
Large parameters; Computational overhead; Limited representation \\
\midrule

RCAN-FSDAF~\cite{cui2024novel} & 
RCAN super-resolution + FSDAF; Two-step reconstruction; RIR module & 
Heterogeneous regions; Land change areas; Spatial details & 
RCAN dependent; Registration errors; Additional training \\
\midrule

CTSTFM~\cite{jiang2024cnntransformer} & 
CNN-Transformer hybrid; Multi-kernel encoder; Cross fusion & 
Low input requirements; Small parameters; Global modeling & 
Boundary seams; Architecture efficiency; Overlapping needed \\
\midrule

SwinSTFM~\cite{chen2022swinstfm} & 
Swin Transformer; FEM + MFM; Shifted window; Unmixing theory & 
Severe land changes; Flood scenarios; Linear unmixing integration & 
High resources; Complex structure; Registration sensitive \\
\midrule

TTSFNet~\cite{mu2024spatiotemporal} & 
Dual-branch Transformer; TFEB + SFEB; Spatiotemporal fusion & 
33-layer inversion; Ocean temperature/salinity; Single model & 
Surface data dependent; Monthly averages; Regional limitations \\
\midrule

STINet~\cite{wang2024stinet} & 
Transformer U-shaped; FF blocks; ST-MSA; CNN+Transformer & 
Vegetation dynamics; Long-range dependencies; Skip connections & 
Time interval sensitive; Training-testing differences; Instability \\
\midrule

MLFF-GAN~\cite{song2022mlffgan} & 
U-net GAN; AdaIN + attention; Multi-level features & 
Temporal uncertainty; Resolution differences; Local/global fusion & 
High resources; Small dataset overfitting; GAN instability \\
\midrule

GAN-STFM~\cite{tan2022flexible} & 
Conditional GAN; ResNet blocks; SwitchNorm; Encoder-decoder & 
Simplified input; Small fluctuation; Easy data preparation & 
Large parameters; GPU required; Generalization unclear \\
\midrule

SS-STFM~\cite{weng2024spatially} & 
Spatial seamless stitching; Overlapping patches; Buffer removal & 
Discontinuity solution; Large-scale fusion; Image continuity & 
Slow processing; Complex buffer; Post-processing only \\
\midrule

HPLTS-GAN~\cite{lei2024hpltsgan} & 
GAN encoder-decoder; ASDT + MLFE + CSAFFIR; U-Net style & 
Temporal independence; Stable accuracy; Strong generalization & 
High complexity; Large parameters; Complex loss tuning \\
\midrule

DCDGAN-STF~\cite{zhang2024dcdgan} & 
Teacher-student GAN; Pyramid cascade; Deformable convolution & 
Large-scale super-resolution; Detail reconstruction; Multi-scene & 
More inference resources; Reference quality dependent \\
\midrule

STFDiff~\cite{huang2024stfdiff} & 
Diffusion model; DS-Unet; Dual-stream encoders; Feature difference & 
Triple uncertainty solution; Iterative refinement; Good generalization & 
Multiple iterations; DDIM acceleration; Common diffusion issues \\
\midrule

DiffSTF~\cite{ma2024conditional} & 
Conditional diffusion; ResBlock + TraBlock; Time encoding & 
Improved accuracy; Cloud coverage handling; Global timestamp & 
No manual parameters; Slow inference; Limited adjustability \\
\midrule

DiffSTSF~\cite{weiDiffusionModelsSpatiotemporalspectral2024} & 
Enhanced diffusion; U-Net denoising; SSC-Block; Wavelet + attention & 
Homogeneous platform; Spatial generalization; Spectral fusion & 
Pansharpening dependent; Sampling step trade-off; Quality balance \\
\midrule

MSFusion~\cite{yang2022msfusion} & 
Multi-stage dual-branch; Texture Transformer; VGG + CNN & 
Global temporal correlation; Self-attention; Adaptive fusion & 
Poor urban performance; Compression loss; Detail recovery issues \\
\midrule

StfMLP~\cite{chen2023stfmlp} & 
Dual MLP networks; Feature pyramid; Transductive learning & 
Lightweight design; Improved efficiency; Data-focused approach & 
Severe change limitations; Boundary effects; Complex structure issues \\
\midrule

SDCS~\cite{liu2024semiblind} & 
Two-stage bidirectional; Semi-blind sensing; PD-CNN + MAF & 
Physical interpretability; Feature learning; Large resolution handling & 
Semi-blind instability; Independent training; RIP conditions \\
\midrule

RealFusion~\cite{guoRealFusionReliableDeep2025} & 
Task decoupling; SIQRN + DSTFN; U-Net + information injection & 
Severe changes; Quality reconstruction; Disaster monitoring potential & 
Rare change types; Training set limitations; Missing type issues \\

\end{longtable}
}  
\setlength{\LTleft}{0pt}
\setlength{\LTright}{0pt}
}

\setlength{\LTleft}{0pt plus 1fill minus 1fill}
\setlength{\LTright}{0pt plus 1fill minus 1fill}


{\scriptsize
\begin{longtable}{@{} m{0.105\textwidth} m{0.153\textwidth} m{0.060\textwidth} m{0.155\textwidth} m{0.15\textwidth} m{0.139\textwidth} m{0.09\textwidth} @{}}

\caption{A taxonomy of deep learning models for spatiotemporal fusion.} \label{tab:model_characteristics} \\

\toprule
\textbf{Network} & \textbf{Model} & \textbf{Year} & \textbf{Dataset} & \textbf{Loss Function} & \textbf{Metrics} & \textbf{Code} \\ 
\midrule
\endfirsthead

\toprule
\textbf{Network} & \textbf{Model} & \textbf{Year} & \textbf{Dataset} & \textbf{Loss Function} & \textbf{Metrics} & \textbf{Code} \\ 
\midrule
\endhead

\bottomrule
\multicolumn{7}{r}{\textbf{\textit{(continued on next page)}}} \\
\endfoot

\bottomrule
\endlastfoot

\multirow{17}{*}{CNN} 
    & STFDCNN~\cite{song2018spatiotemporal} & 2018 & CIA, LGC & MSE Loss & \makecell[l]{RMSE, ERGAS\\SAM, SSIM} & N/A \\
    \cmidrule(lr){2-7}
    & STFNet~\cite{liu2019stfnet} & 2019 & Landsat, MODIS & \makecell[l]{MSE Loss,\\TC Loss} & \makecell[l]{RMSE, CC\\SSIM} & N/A \\
    \cmidrule(lr){2-7}
    & EDCSTFN~\cite{tan2019enhanced} & 2019 & MODIS, Landsat & \makecell[l]{Content Loss,\\Feature Loss,\\Vision Loss} & \makecell[l]{RMSE, ERGAS\\SAM, SSIM} & \href{https://github.com/theonegis/edcstfn}{Link} \\
    \cmidrule(lr){2-7}
    & STF3DCNN~\cite{peng2020fast} & 2020 & CIA, LGC, RDT & MSE Loss & \makecell[l]{CC, SAM\\PSNR, UIQI} & N/A \\
    \cmidrule(lr){2-7}
    & STTFN~\cite{yin2021spatiotemporal} & 2021 & Landsat, MODIS & Huber Loss & \makecell[l]{RMSE, SSIM} & N/A \\
    \cmidrule(lr){2-7}
    & MCDNet~\cite{li2021multi} & 2021 & CIA, LGC & \makecell[l]{Feature Loss,\\Content Loss} & \makecell[l]{SSIM, RMSE\\CC, $R^2$} & N/A \\
    \cmidrule(lr){2-7}
    & MOST~\cite{wei2021enblending} & 2021 & Landsat-8, MODIS & \makecell[l]{L1 Loss} & \makecell[l]{RMSE, SSIM\\SAM, ERGAS} & N/A \\
    \cmidrule(lr){2-7}
    & MUSTFN~\cite{qin2022mustfn} & 2022 & \makecell[l]{Landsat-7,\\ GaoFen-1} & \makecell[l]{MAW-MSE Loss,\\VI Loss,\\SSIM Loss} & \makecell[l]{RMSE, MAE\\rMAE} & \href{https://github.com/qpyeah/MUSTFN}{Link} \\
    \cmidrule(lr){2-7}
    & DenseSTF~\cite{ao2022deep} & 2022 & CIA, LGC & MSE Loss & \makecell[l]{CC, RMSE\\SSIM} & \href{https://github.com/sysu-xin-lab/DenseSTF}{Link} \\
    \cmidrule(lr){2-7}
    & STF-EGFA~\cite{cheng2022stfegfa} & 2022 & \makecell[l]{AHB, Daxing,\\Tianjin} & \makecell[l]{Content Loss,\\Feature Loss,\\Visual Loss} & \makecell[l]{SAM, PSNR\\CC, SSIM} & N/A \\
    \cmidrule(lr){2-7}                  
    & HCNNet~\cite{zhu2022hcnn} & 2022 & CIA, LGC, Daxing & \makecell[l]{MAE Loss,\\ MS-SSIM Loss} & \makecell[l]{RMSE, SAM\\ERGAS, CC} & N/A \\
    \cmidrule(lr){2-7}
    & MACNN~\cite{chen2022multiscale_srs} & 2022 & Landsat, MODIS & \makecell[l]{TC Loss,\\TD Loss} & \makecell[l]{RMSE, SSIM\\CC, AD} & N/A \\
    \cmidrule(lr){2-7}
    & ECPW-STFN~\cite{zhang2024enhanced} & 2024 & CIA, Daxing & \makecell[l]{Wavelet Loss,\\Feature Loss,\\Vision Loss} & \makecell[l]{RMSE, SSIM\\CC,SAM} & \href{https://github.com/lixinghua5540/ECPW-STFN}{Link} \\
    \cmidrule(lr){2-7}
    & MLKNet~\cite{jiang2024mlknet} & 2024 & CIA, DX, SW & \makecell[l]{Content loss,\\Feature loss,\\Visual loss} & \makecell[l]{PSNR, SSIM\\SAM, CC} & N/A \\
    \cmidrule(lr){2-7}
    & RCAN-FSDAF~\cite{cui2024novel} & 2024 & \makecell[l]{Landsat 8 OLI,\\ MODIS13Q1} & \makecell[l]{L1 Loss} & \makecell[l]{RMSE, R,\\AD} & N/A \\
    \cmidrule(lr){2-7}
    & CTSTFM~\cite{jiang2024cnntransformer} & 2024 & CIA, DX & L1 Loss & \makecell[l]{MAE, SAM,\\ PSNR, SSIM} & N/A \\
\midrule

\multirow{5}{*}{Transformer} 
    & SwinSTFM~\cite{chen2022swinstfm} & 2022 & CIA, LGC, AHB & \makecell[l]{Charbonnier Loss,\\ MS-SSIM Loss} & \makecell[l]{RMSE, SSIM,\\SAM, ERGAS} & \href{https://github.com/LouisChen0104/swinstfm}{Link} \\
    \cmidrule(lr){2-7}
    & TTSFNet~\cite{mu2024spatiotemporal} & 2024 & SST, SSS, SSHA, SSWA, SSTA & MSE Loss & \makecell[l]{$R^2$, RMSE\\NRMSE} & N/A \\
    \cmidrule(lr){2-7}
    & STINet~\cite{wang2024stinet} & 2024 & \makecell[l]{Landsat-8,\\ Sentinel-2} & \makecell[l]{RMSE Loss} & \makecell[l]{RMSE} & N/A \\
    \cmidrule(lr){2-7}
    & STFTN~\cite{zhao2024enso} & 2024 & \makecell[l]{CMIP6, SODA,\\ GODAS, NMME} & \makecell[l]{Weighted RMSE} & \makecell[l]{RMSE, PCC} & N/A \\
    \cmidrule(lr){2-7}
    & MGSFformer~\cite{yu2025mgsfformer} & 2025 & \makecell[l]{Beijingsites,\\ Chinacities} & \makecell[l]{MSE Loss} & \makecell[l]{MSE, MAE,\\CORR} & \href{https://github.com/ChengqingYu/MGSFformer}{Link} \\
\midrule

\multirow{5}{*}{GAN} 
    & MLFF-GAN~\cite{song2022mlffgan} & 2022 & CIA, LGC & \makecell[l]{GAN Loss,\\ L1 Loss,\\ Spectrum Loss} & \makecell[l]{MAE, RMSE,\\SSIM} & \href{https://github.com/songbingze/MLFF-GAN}{Link} \\
    \cmidrule(lr){2-7}
    & GAN-STFM~\cite{tan2022flexible} & 2022 & CIA, LGC & \makecell[l]{LSGAN Loss,\\ Feature Loss,\\ Vision Loss} & \makecell[l]{RMSE, SSIM} & \href{https://github.com/theonegis/ganstfm}{Link} \\
    \cmidrule(lr){2-7}
    & SS-STFM~\cite{weng2024spatially} & 2024 & LGC, BJGF6 & \makecell[l]{GAN Loss} & \makecell[l]{RMSE, AD} & N/A \\
    \cmidrule(lr){2-7}
    & HPLTS-GAN~\cite{lei2024hpltsgan} & 2024 & CIA, LGC & \makecell[l]{LSGAN Loss,\\L1 Loss,\\Vision Loss} & \makecell[l]{RMSE, SSIM} & N/A \\
    \cmidrule(lr){2-7}
    & DCDGAN-STF~\cite{zhang2024dcdgan} & 2024 & \makecell[l]{CIA, LGC,\\ Wuhan} & \makecell[l]{GAN Loss,\\Visual Loss,\\Spectral Loss} & \makecell[l]{SSIM, RMSE} & \href{https://github.com/zhangyanxa/DCDGAN-STF}{Link} \\
\midrule

\multirow{4}{*}{Diffusion} 
    & MFDGCN~\cite{cuiMFDGCNMultiStageSpatioTemporal2022} & 2022 & PeMS\_BAY & \makecell[l]{MAE Loss} & \makecell[l]{MAE, RMSE,\\MAPE} & N/A \\
    \cmidrule(lr){2-7}
    & STFDiff~\cite{huang2024stfdiff} & 2024 & CIA, LGC & \makecell[l]{Simple Loss} & \makecell[l]{RMSE, SSIM,\\ERGAS} & \href{https://github.com/prowDIY/STF}{Link} \\
    \cmidrule(lr){2-7}
    & DiffSTF~\cite{ma2024conditional} & 2024 & CIA, LGC & \makecell[l]{MSE Loss,\\Spectral Loss} & \makecell[l]{RMSE, SAM,\\RASE, SSIM} & N/A \\
    \cmidrule(lr){2-7}
    & DiffSTSF~\cite{weiDiffusionModelsSpatiotemporalspectral2024} & 2024 & \makecell[l]{Gaofen-1\\(PAN, MS, WFV)} & \makecell[l]{MSE Loss,\\ KL Divergence} & \makecell[l]{RMSE, SSIM,\\ERGAS} & \href{https://github.com/isstncu/gf1fusion}{Link} \\
\midrule

\multirow{3}{*}{\makecell[l]{Sequence\\ Models}}                                                         
    & CNN-LSTM~\cite{wu2020spatiotemporal} & 2020 & \makecell[l]{TRMM, GridSat-B1,\\ DEM, Rain Gauges} & \makecell[l]{MSE Loss} & \makecell[l]{RMSE, MAE,\\CC} & N/A \\
    \cmidrule(lr){2-7}
    & ConvLSTM~\cite{cui2024soil} & 2024 & \makecell[l]{ERA5-land, GPM} & \makecell[l]{N/A} & \makecell[l]{FCA, TCA,\\OCA, R} & N/A \\
    \cmidrule(lr){2-7}
    & Geo-BiLSTMA~\cite{jia2024forecasting} & 2024 & Xi'an & \makecell[l]{L2 Loss} & \makecell[l]{RMSE, $R^2$} & N/A \\
\midrule

\multirow{6}{*}{Others} 
    & STHGCN~\cite{wang2022sthgcn} & 2022 & \makecell[l]{METR-LA,\\ PEMS-BAY,\\ Solar Energy} & \makecell[l]{Huber Loss,\\ KL Divergence} & \makecell[l]{MSE, RMSE} & N/A \\ 
    \cmidrule(lr){2-7}
    & MSFusion~\cite{yang2022msfusion} & 2022 & CIA, LGC, DX & L1 Loss & \makecell[l]{SSIM, RMSE,\\ERGAS, SAM} & N/A \\ 
    \cmidrule(lr){2-7}
    & STFMLP~\cite{chen2023stfmlp} & 2023 & CIA, LGC & MSE Loss & \makecell[l]{RMSE, SAM} & \href{https://github.com/luhailaing-max/STFMLP-master}{Link} \\ 
    \cmidrule(lr){2-7}
    & DSTFNet~\cite{cai2023dualbranch} & 2023 & \makecell[l]{GF-2,\\ Sentinel-2} & \makecell[l]{Tanimoto Loss} & \makecell[l]{MCC, F1-score} & N/A \\ 
    \cmidrule(lr){2-7}
    & SDCS~\cite{liu2024semiblind} & 2024 & \makecell[l]{CIA, LGC,\\AHB, Tianjin} & \makecell[l]{L1 Loss, L2 Loss,\\SSIM Loss} & \makecell[l]{RMSE, SSIM,\\SAM} & \href{https://github.com/yc-cui/SDCS}{Link} \\ 
    \cmidrule(lr){2-7}
    & RealFusion~\cite{guoRealFusionReliableDeep2025} & 2025 & \makecell[l]{Kansas, Taihang,\\Poyang} & \makecell[l]{Charbonnier loss,\\visual loss,\\edge information loss} & \makecell[l]{RMSE, SSIM,\\CC, SAM} & N/A \\

\end{longtable}
}

\setlength{\LTleft}{0pt}
\setlength{\LTright}{0pt}

\twocolumn

This operation enables CNNs to effectively capture spatial 
features at various levels of abstraction. Following convolution layers, pooling operations 
reduce spatial dimensions while preserving essential information:
\begin{equation}
y(i,j) = \max_{0 \leq m < s, 0 \leq n < s} x(i \cdot s + m, j \cdot s + n),
\end{equation}
\noindent where $s$ denotes the pooling window size, $y(i,j)$ is the output feature at position $(i,j)$,
$x(i \cdot s + m, j \cdot s + n)$ represents the input feature at the corresponding position, and
$m$ and $n$ are indices within the pooling window. Initially, 
CNNs demonstrated excellent performance in image classification and recognition tasks 
with multi-layer convolution and pooling 
operations~\cite{sunEvolvingDeepConvolutional2020}. For example, 
AlexNet~\cite{krizhevsky2012imagenet} significantly improved classification accuracy in the 
ImageNet competition with its deep convolutional structure. Later, 
VGGNet~\cite{muhammadPretrainedVGGNetArchitecture2018} deepened the network layers further, 
enhancing feature extraction capability. These successes prompted researchers to apply CNNs 
to remote sensing image processing to leverage their powerful spatial feature extraction 
ability for complex tasks such as land cover classification and 
change detection~\cite{simonyan2015very}.

\minew{
Semantic segmentation is one of the primary tasks in the field of computer vision, with the purpose of classifying pixels 
and assigning labels to achieve a more refined understanding of image content. In the remote sensing field, semantic segmentation 
is widely applied to tasks such as land cover classification, urban remote sensing, and agricultural monitoring, enabling precise 
identification and analysis of complex ground objects in high-resolution remote sensing images. In recent years, the remote sensing 
field typically combines CNN and Transformer models for image classification to efficiently perform pixel-level segmentation. CNNs excel 
at capturing local spatial features in images, while Transformers effectively model global contextual information, and both have achieved 
excellent results in remote sensing image segmentation tasks. Among these, CNN models represented by 
Fully Convolutional Networks (FCN)~\cite{guoPixelWiseClassificationMethod2018} have performed outstandingly in remote sensing image 
semantic segmentation. After incorporating atrous convolution~\cite{liang-chiehSemanticImageSegmentation2015} and fully connected Conditional 
Random Fields (CRFs)~\cite{krahenbuhlEfficientInferenceFully2011}, the classification refinement and 
boundary clarity of the models have been significantly improved.
}

In early applications in remote sensing, STFDCNN~\cite{song2018spatiotemporal} achieved 
spatiotemporal fusion of MODIS and Landsat images using nonlinear mapping and 
super-resolution convolutional networks, significantly improving fusion accuracy and 
efficiency. However, this model faced issues with insufficient detail retention and poor 
spectral consistency when handling complex terrains and rapidly changing land 
cover~\cite{liNewSensorBiasdriven2020}. 

To address the inherent time-space conflict in 
spatiotemporal fusion, Liu~\etal proposed STFNet~\cite{liu2019stfnet}, which uses a 
dual-stream CNN structure to process spatial and temporal features separately, combining 
temporal dependence and consistency through adaptive feature fusion. This architecture 
mitigates the resolution trade-off between \minew{high spatial low temporal (HSLT)} and \minew{high temporal low spatial (HTLS)} data by explicitly modeling      
temporal dynamics in one stream while preserving spatial details in another. Specifically, 
STFNet~\cite{liu2019stfnet} reconstructs the target fine image $F_2$ through an adaptive weighting strategy:
\begin{equation}
F_2 = \alpha * (F_1 + F_{12}) + (1 - \alpha) * (F_3 - F_{23}),
\end{equation}
\noindent where $F_{12}$ and $F_{23}$ are fine difference images predicted from corresponding coarse 
ones and neighboring fine images, and $\alpha$ is a weighting parameter determined by the 
temporal similarity between coarse images. This adaptive fusion mechanism enables STFNet~\cite{liu2019stfnet} to 
effectively balance the contribution of temporal information from both before and after the 
prediction time, thereby improving the reconstruction quality in areas with complex terrain 
and rapid land cover changes.

CNNs extract features through a hierarchical process, where lower-level spatial 
details evolve into more abstract semantic concepts across successive network 
layers. This progressive abstraction enables the network to capture increasingly 
complex patterns. The fundamental convolution operation shows how local receptive 
fields interact with kernels to produce feature maps that become inputs to deeper 
layers. This feature extraction mechanism has made CNNs particularly effective 
for spatiotemporal fusion tasks, where both spatial details and temporal changes 
must be accurately represented. 

\minew{Building on this fundamental architecture, Tan~\etal proposed 
EDCSTFN~\cite{tan2019enhanced}, which represents a significant 
advancement in CNN-based spatiotemporal fusion through three key 
innovations: an encoder-merge-decoder architecture with residual 
learning, a compound loss function, and flexible reference data 
strategies. The EDCSTFN architecture employs a residual encoder to 
learn feature differences between reference and prediction dates. 
This design eliminates the restrictive assumption in DCSTFN that 
ground changes observed from different sensors are identical. To 
address the image blurriness issue common in reconstruction, EDCSTFN 
introduces a compound loss function that ensures pixel-level accuracy 
through MSE, preserves essential textures using a pre-trained 
AutoEncoder, and employs MS-SSIM (Multi-Scale Structural Similarity 
Index) to enhance image sharpness and perceptual quality. MS-SSIM 
evaluates structural similarity across multiple scales, providing a 
more comprehensive assessment of image structure and visual quality 
compared to single-scale SSIM. For cases with two reference pairs, 
EDCSTFN implements an adaptive weighting strategy based on inverse 
distance weighting, allowing the model to dynamically adjust 
contributions based on temporal change magnitude. The implementation 
adopts several key design principles to enhance performance: (1) 
maintaining original spatial resolution throughout the network to 
preserve texture details, (2) using bicubic interpolation instead 
of transposed convolution to avoid checkerboard artifacts, and (3) 
stacking multi-temporal inputs channel-wise to capture temporal 
correlations.}

Further advancing this approach, Cai~\etal introduced 
DPSTFN~\cite{cai2022progressive}, which adopted a progressive fusion 
framework that combined pan-sharpening, super-resolution, and 
spatiotemporal fusion modules. Leveraging Residual Dense Blocks 
(RRDB)~\cite{chenSuperResolutionSatelliteImages2022} and a decoupled 
spatial-spectral attention mechanism, DPSTFN~\cite{cai2022progressive} 
achieved enhanced fusion results by sequentially resolving spatial 
and temporal discrepancies. This hierarchical approach alleviates 
the time-space conflict through stage-wise refinement, where initial 
stages focus on spatial enhancement while subsequent layers handle 
temporal coherence.

Recent studies, such as CIG-STF~\cite{you2024cigstf} proposed by You~\etal, 
explicitly use land cover change detection in spatiotemporal fusion to enhance 
model performance. CIG-STF~\cite{you2024cigstf} introduced a Change Information-Guided Enhancement 
Module to enhance the generation of superior final prediction results. 
CIG-STF~\cite{you2024cigstf} designed a multi-scale dilated convolution feature extractor and a 
spatiotemporal fusion-change detection module, excelling in the 
reconstruction of sudden change areas and significantly enhancing the model's 
robustness in complex environments~\cite{lianRecentAdvancesDeep2025a}. By 
incorporating change-aware temporal modeling, this framework dynamically adjusts 
spatial reconstruction intensity based on temporal change magnitude, thereby 
balancing the resolution conflict through adaptive feature weighting.

\minew{
CNNs demonstrate superior spatial feature extraction through hierarchical representations~\cite{tan2022robust}. Their 
parallel processing capability enables efficient computation, while transfer learning from pre-trained models accelerates 
convergence. The automatic feature learning eliminates manual feature engineering, and residual connections effectively 
preserve both spatial details and temporal dynamics. These advantages make CNN-based methods particularly suitable for 
operational applications requiring rapid processing, such as agricultural yield estimation where timely predictions are 
crucial for market decisions, and routine land cover monitoring where computational efficiency is paramount. The robust 
spatial feature extraction also excels in urban expansion monitoring, where detecting subtle changes in building patterns 
and infrastructure development requires precise spatial detail preservation~\cite{zhang2024wuhan,cai2023dualbranch}.

Despite these advances, CNN-based methods face inherent constraints that limit their applicability in certain scenarios. 
The limited receptive field restricts long-range spatial context modeling, even with dilated convolutions, which becomes 
problematic in large-scale flood monitoring where understanding watershed-level patterns is essential~\cite{chen2022swinstfm}. 
Temporal modeling remains challenging as CNNs primarily excel at spatial processing, requiring specialized architectures 
for sequence modeling, which limits their effectiveness in phenological monitoring applications that demand accurate 
tracking of vegetation cycles across growing seasons. The methods demand substantial labeled training 
data and show limited generalization across different sensors and geographical regions, creating barriers for developing 
countries or remote areas where training data is scarce. Additionally, the computational requirements for training deep 
networks can be prohibitive for real-time disaster response applications, where rapid deployment and processing of 
multi-temporal imagery is critical for emergency management~\cite{guo2024obsum}.
}

\subsubsection{Transformer}
\begin{figure}[t]
    \centering
    \includegraphics[width=0.5\textwidth]{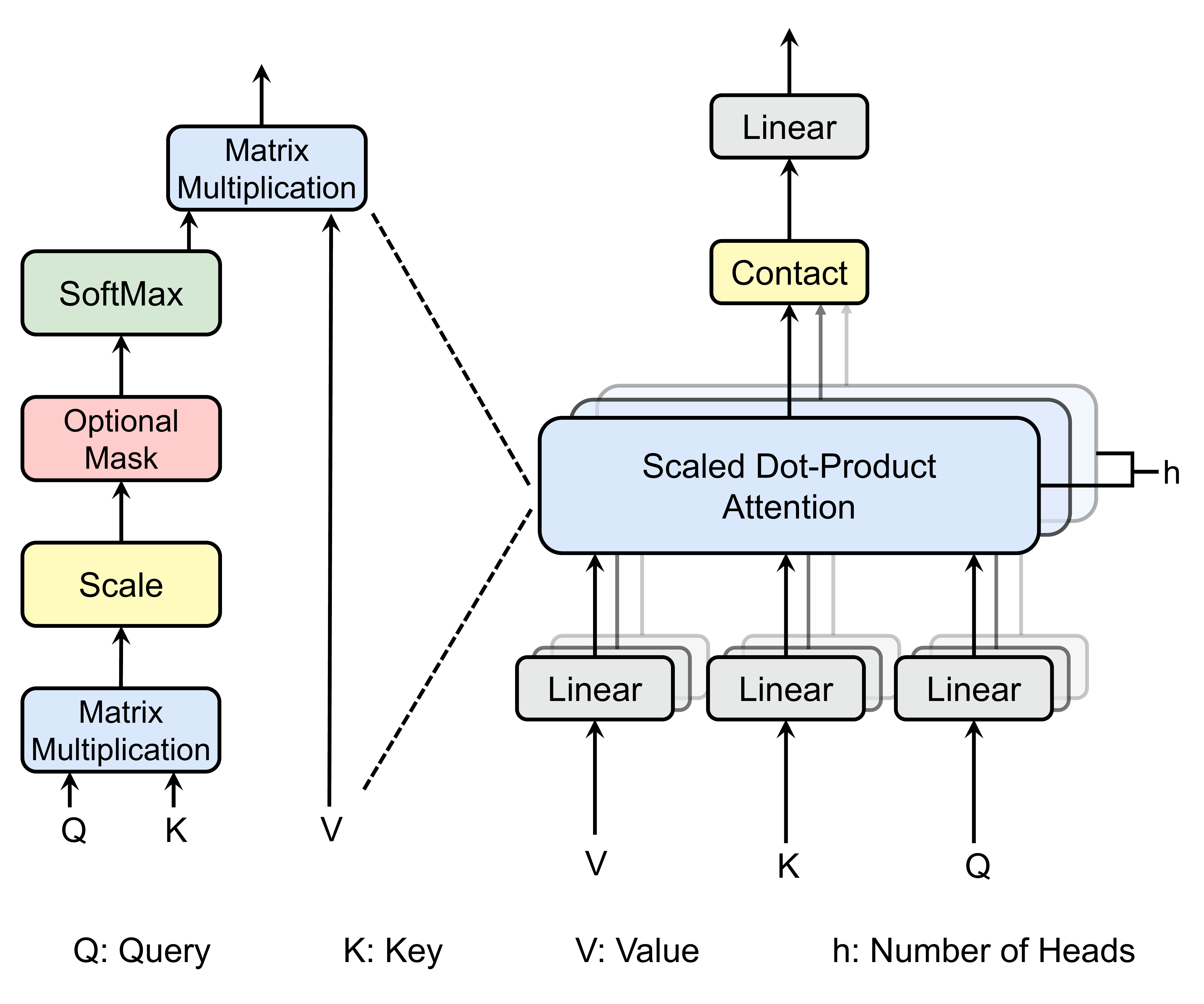}
    \caption{Multi-head attention mechanism in Transformer.}
    \label{fig:transformer_mechanism}
\end{figure}

Vaswani~\etal~\cite{vaswaniAttentionAllYou2017} introduced the Transformer model to address sequence-to-sequence tasks like 
machine translation, text summarization, and speech recognition~\cite{wangLearningDeepTransformer2019}. Unlike traditional recurrent 
neural networks, Transformer possesses a unique self-attention mechanism that effectively 
captures long-term dependencies while enabling more efficient parallel computing~\cite{karitaComparativeStudyTransformer2019,wangRTransformerRecurrentNeural2019}. 
\autoref{fig:transformer_mechanism} illustrates Transformer's multi-head attention 
mechanism, including the matrix multiplication (MatMul) of query ($Q$), key ($K$), and 
value ($V$) matrices, scaling, optional masking, and SoftMax operations. 
\minew{The linear 
transformations of $Q$, $K$, and $V$ are processed through $h$ parallel scaled dot-product attention 
heads, which are then concatenated and linearly transformed to generate the final output representations.}
The self-attention mechanism can be formulated as:
\begin{equation}
\text{Attention}(Q, K, V) = \text{softmax}\left(\frac{QK^T}{\sqrt{d_k}}\right)V.
\end{equation}

Within this foundational formula, the variables $Q$, $K$ and $V$ correspond to query, key, 
and value matrices extracted from the input sequence, while $d_k$ indicates the 
dimensionality of the key vectors. This distinctive network architecture has 
achieved exceptional results in natural language processing and has been widely applied 
across various domains, including \minew{computer vision (CV)}, marking the beginning of Transformer 
applications in multimodal tasks~\cite{gilliozOverviewTransformerbasedModels2020,jamshedNLPMeetsVision2021,chenVisionStatusResearch2022}.

Vision Transformer (ViT)~\cite{dosovitskiy2021transformers} represents the beginning of 
Transformer applications in computer vision by dividing images into fixed-size patches and 
processing the flattened sequence of patches using Transformer. Given an input image 
$\mathbf{x} \in \mathbb{R}^{H \times W \times C}$ where $H$ and $W$ are the height and width, 
and $C$ is the number of channels, the image is divided into $N = HW/P^2$ non-overlapping 
patches of size $P \times P$. This sequence conversion can be represented as:
\begin{equation}
\mathbf{z}_0 = [\mathbf{x}_\text{class}; \mathbf{x}_p^1\mathbf{E}; \mathbf{x}_p^2\mathbf{E}; 
\cdots; \mathbf{x}_p^N\mathbf{E}] + \mathbf{E}_{pos}.
\end{equation}

Breaking down the components of this equation: each patch $\mathbf{x}_p^n \in \mathbb{R}^{P^2C}$ 
is obtained by flattening the $n$-th image patch, where $n \in \{1, 2, ..., N\}$. The trainable 
linear projection matrix $\mathbf{E} \in \mathbb{R}^{(P^2 \cdot C) \times D}$ maps each 
flattened patch to a $D$-dimensional embedding space, resulting in patch embeddings 
$\mathbf{x}_p^n\mathbf{E} \in \mathbb{R}^{D}$. The learnable class token 
$\mathbf{x}_\text{class} \in \mathbb{R}^{D}$ serves as a global image representation similar 
to BERT's [CLS] token. The position embeddings $\mathbf{E}_{pos} \in \mathbb{R}^{(N+1) \times D}$ 
are added to retain positional information, where the first row corresponds to the class token 
and the remaining $N$ rows correspond to the $N$ patches. The final sequence 
$\mathbf{z}_0 \in \mathbb{R}^{(N+1) \times D}$ is then processed by the Transformer encoder.
This approach has even surpassed traditional convolutional neural networks in multiple image 
classification benchmarks~\cite{sunClassificationThyroidNodule2023, 
kimViTBasedMultiScaleClassification2024}.

However, ViT~\cite{dosovitskiy2021transformers} has relative limitations, such as consuming 
excessive resources when processing high-resolution images~\cite{safaeiNovelExperimentaltheoreticalMethod2023,sahinMultiobjectiveOptimizationViT2024}. 
Subsequent researchers introduced hierarchical structures and sliding window self-attention mechanisms, 
effectively reducing computational complexity while maintaining excellent performance, 
enhancing Transformer's practicality in CV tasks and laying the foundation 
for its expansion into more complex applications~\cite{aleissaeeTransformersRemoteSensing2023,wangTransformersRemoteSensing2024}.

Remote sensing images typically possess high dimensionality and complex spatial features, 
presenting challenges in computational resources and modeling capabilities when processing 
large-scale remote sensing data~\cite{baziVisionTransformersRemote2021,maRemoteSensingBig2015,chiBigDataRemote2016}. 
To address these challenges, researchers began introducing 
Transformer's self-attention mechanism into remote sensing image processing to better 
capture spatial and spectral features~\cite{guoTransformerBasedChannelspatial2022}. Classic Transformer models like Swin Transformer~\cite{chen2022swinstfm} 
have been widely applied to remote sensing image classification, object detection, and 
other tasks~\cite{liangEnhancedSelfAttentionNetwork2023}. As technology evolved, more innovative Transformer models emerged, driving 
applications in spatiotemporal fusion and other tasks, improving fusion accuracy while 
providing greater adaptability and robustness in computational complexity and detail 
preservation~\cite{qiExploringReliableInfrared2024, wangUnveilingMultiDimensionalSpatioTemporal2024}.

\minew{Building on the transformer architecture's success in computer 
vision, Chen~\etal proposed SwinSTFM~\cite{chen2022swinstfm}, which 
adapts the Swin Transformer for remote sensing spatiotemporal fusion 
through innovative architectural modifications. The model consists of 
two main modules: the Feature Extraction Module (FEM) that employs 
shifted window-based self-attention with learnable relative position 
bias for hierarchical feature extraction, and the Multilevel Fusion 
Module (MFM) that addresses the mixed pixel problem in coarse resolution 
imagery. The key innovation lies in the multi-head unmixing attention 
(MUA) module, which generates queries and keys from different sources 
to learn adaptive weights based on spectral mixing relationships between 
coarse and fine pixels, unlike conventional attention mechanisms. This 
unmixing-based fusion approach, combined with a correction factor for 
land-cover changes and hierarchical multi-scale processing through 
successive Swin extraction blocks, enables the model to simultaneously 
consider spatial relationships and temporal dynamics while preserving 
both fine spatial details and temporal consistency across different 
resolution levels.}

Subsequently, models like CFFormer~\cite{zhao2024cfformer} further improved the handling of spatiotemporal 
conflicts by combining GAN and Transformer architectures. Accurate modeling of temporal 
dynamics while maintaining high spatial resolution requires complementary advantages, 
leading researchers to introduce generative modules specifically designed to enhance 
spatial detail preservation. Through multi-scale feature fusion, these models improved 
adaptability to temporal changes in complex environments and heterogeneous data, thus 
balancing temporal and spatial information processing. 

MGSFformer~\cite{yu2025mgsfformer} represents the latest 
approach to resolving spatiotemporal conflicts, not viewing spatial and temporal 
dimensions as opposing elements but introducing an integrated framework containing 
residual redundancy removal modules, spatiotemporal attention modules, and dynamic 
fusion modules. This framework simulates interactions between spatial and temporal 
features at multiple granularity levels, establishing a more harmonious relationship 
between these traditionally conflicting dimensions. MGSFformer~\cite{yu2025mgsfformer} has demonstrated 
outstanding performance in fusion experiments across multiple datasets, better addressing 
the spatiotemporal conflict problem that has long plagued spatiotemporal fusion tasks.

\minew{
Transformer architectures bring significant advantages to spatiotemporal fusion through their global receptive fields 
and dynamic attention mechanisms. As demonstrated by SwinSTFM, the self-attention mechanism captures long-range 
dependencies that CNNs miss, while the shifted window scheme maintains computational efficiency for large satellite 
imagery~\cite{chen2022swinstfm}. The integration of unmixing theories within the attention framework effectively 
combines data-driven learning with domain knowledge. These capabilities make Transformers ideal for climate change 
monitoring applications where capturing global atmospheric patterns and long-term temporal trends is essential, and 
for ecosystem monitoring where complex interactions between distant regions need to be modeled~\cite{zhao2024enso,mu2024spatiotemporal}. 
The superior temporal modeling also benefits crop phenology tracking across entire growing seasons, enabling more 
accurate yield predictions and agricultural management decisions~\cite{wang2024stinet}.

However, transformers require more computational resources and larger training datasets than CNNs, which can be 
limiting in remote sensing applications where high-quality paired data is scarce. This becomes particularly 
problematic in rapid disaster response scenarios where immediate processing is required without time for extensive 
model training~\cite{kim2023multisource}. Additionally, window-based attention may sacrifice some global context, 
potentially missing critical large-scale patterns in applications like ocean current monitoring or continental-scale 
vegetation dynamics. Current position encodings may not fully capture the complex multi-sensor relationships in 
spatiotemporal fusion tasks, leading to suboptimal performance when integrating data from sensors with vastly 
different characteristics (e.g., combining SAR and optical imagery for flood mapping). The high memory requirements 
also limit their deployment on edge devices for in-situ agricultural monitoring systems where computational 
resources are constrained~\cite{chen2022swinstfm}.
}

\subsubsection{Generative Models}
Generative models work by learning the underlying distribution of input data to create new 
data samples~\cite{ruthottoIntroductionDeepGenerative2021,cuiReconstructionLargeScaleMissing2024}. 
Their main goal is to simulate real-world data distributions and generate new 
data with similar statistical properties~\cite{shahamSinGANLearningGenerative2019}. With the rapid development of deep learning, 
generative models have evolved from traditional probabilistic graphical models to deep 
generative networks~\cite{oussidiDeepGenerativeModels2018}. In their early stages, generative models relied on traditional methods 
like Hidden Markov Models (HMM)~\cite{soruriDesignFabricationGaN2023} and Gaussian Mixture Models (GMM)~\cite{huangGANbasedGaussianMixture2021}, capturing data 
distributions and structures to generate new samples~\cite{xieGenerativeLearningImbalanced2019}. These approaches were primarily used 
for simpler data generation tasks because they were limited by data dimensionality and complexity~\cite{manduchiChallengesOpportunitiesGenerative2024}.

The emergence of deep learning enabled further advancement of generative models~\cite{correiaEvolutionaryGenerativeModels2024}. Deep 
generative networks leverage neural networks to model complex data distributions, evolving 
into more powerful and flexible data generation methods. Typical examples include GANs and Variational Autoencoders (VAEs). Introduced by 
Goodfellow~\cite{goodfellow2014gan}, GANs revolutionized generative modeling through adversarial training 
between a generator and a discriminator~\cite{goodfellowDistinguishabilityCriteriaEstimating2015}. Their strength lies in generating 
highly realistic and diverse samples, greatly satisfying requirements for tasks such as image generation, image restoration, and style transfer. 
VAEs~\cite{gravingVAESNEDeepGenerative2020}, on the other hand, learn latent distributions 
of data by maximizing the variational lower bound, proving highly effective in unsupervised 
learning and generative tasks. Although VAEs~\cite{gravingVAESNEDeepGenerative2020} have more stable training processes than GANs, 
the quality and diversity of their generated samples are typically lower~\cite{caoSurveyGenerativeDiffusion2023, croitoruDiffusionModelsVision2023b}.

Diffusion models have brought new prospects to generative modeling in recent years. By 
simulating gradual ``noising'' of data and reverse denoising, these models can generate 
high-quality images while improving generation efficiency. These new models address some 
challenges faced by traditional generative models in detail recovery and training stability, 
demonstrating exceptional performance in areas such as image generation, image restoration, 
and super-resolution tasks~\cite{liuReviewRemoteSensing2021, wangRemoteSensingImage2019}. The continuous advancement of generative models has provided 
new solutions for data augmentation, image reconstruction, and fine-grained feature recovery, 
driving technological development in fields like CV, RS, and 
medical imaging.

\begin{figure}[t]
    \centering
    \includegraphics[width=.5\textwidth]{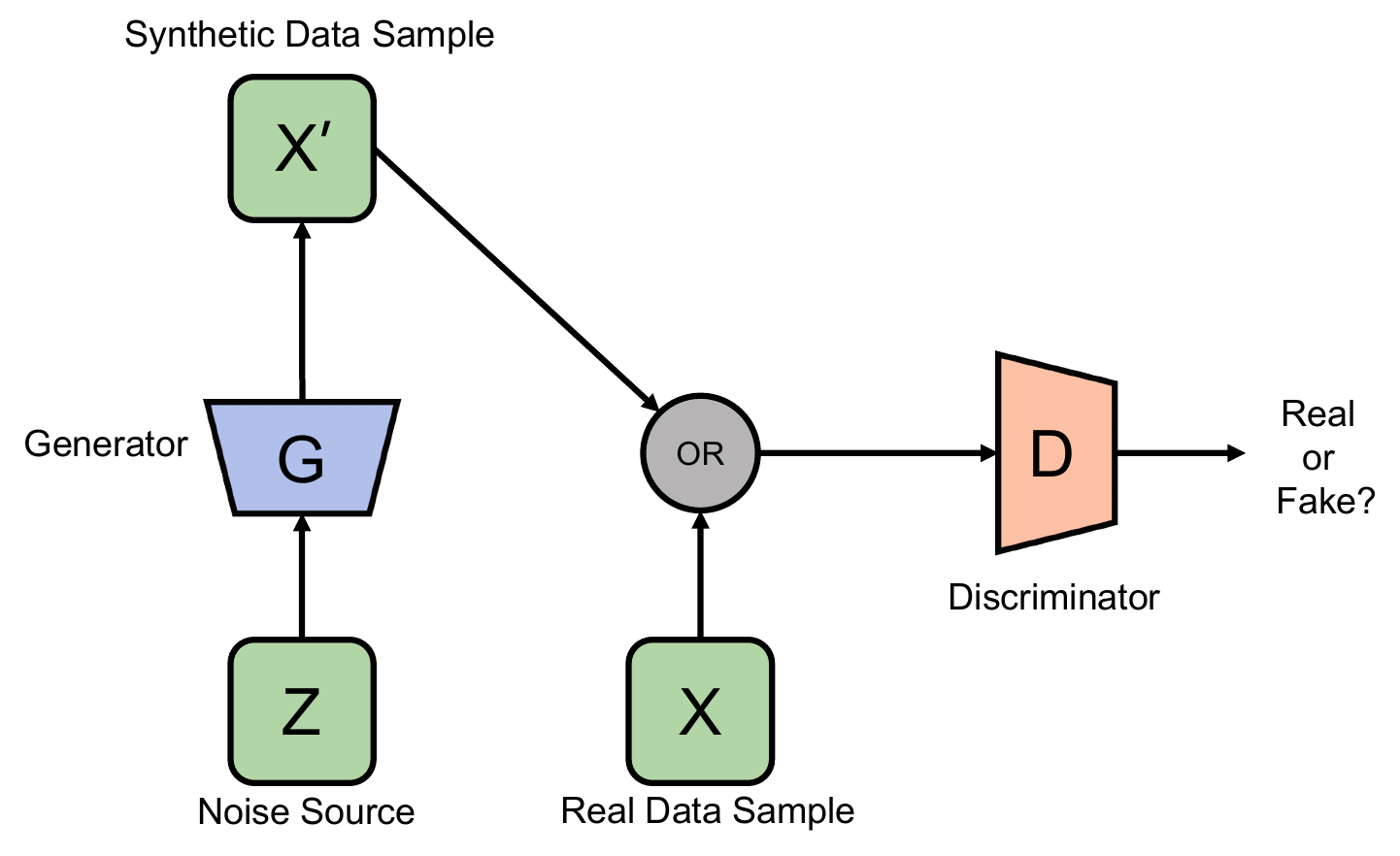}
    \caption{\minew{Basic Generative Adversarial Network framework with generator (G) and discriminator (D).}}
    \label{fig:gan_framework}
\end{figure}

\noindent\textbf{Generative Adversarial Networks}. GANs were first introduced by Goodfellow~\etal~\cite{goodfellow2014gan}, 
achieving high-quality data generation through adversarial training 
between a generator and a discriminator. As shown in 
\autoref{fig:gan_framework}, the GAN framework consists of two competing neural 
networks: a generator ($G$) that transforms random noise $\boldsymbol{z}$ from a 
noise source into synthetic data samples $\boldsymbol{x}'$, and a discriminator ($D$) 
that attempts to distinguish between real data samples $\boldsymbol{x}$ and generated 
samples. The inputs are alternately fed to the discriminator through an OR gate, 
which then outputs a probability verdict of ``Real or Fake ?'' for the incoming sample. 
Through this adversarial process, the generator progressively improves its ability 
to create realistic samples that can fool the discriminator. The core of this approach 
is formalized as a two-player minimax game, with the following value function $V(G, D)$:
\begin{equation}
\begin{aligned}
\min_{G} \max_{D} V(D, G) &= \mathbb{E}_{\boldsymbol{x}\sim p_{\text{data}}(\boldsymbol{x})}[\log D(\boldsymbol{x})] \\
&\quad + \mathbb{E}_{\boldsymbol{z}\sim p_{\boldsymbol{z}}(\boldsymbol{z})}[\log(1 - D(G(\boldsymbol{z})))].
\end{aligned}
\end{equation}Within this framework, the generator $G$ transforms random noise $\boldsymbol{z}$ 
into synthetic data through the mapping $G(\boldsymbol{z}; \theta_g)$, 
with $\theta_g$ comprising the generator's learnable parameters. Conversely, 
the discriminator $D$ calculates the likelihood that an input originated from 
actual data rather than being artificially generated. These expectations 
$\mathbb{E}$ are calculated across the distribution of authentic data 
$p_{\text{data}}(\boldsymbol{x})$ and the distribution of noise inputs 
$p_{\boldsymbol{z}}(\boldsymbol{z})$.

Undeniably, the GAN architecture pioneered a new era in generative models, though 
it often faces instability and mode collapse issues during training.

To address these limitations, Radford~\etal proposed DCGAN~\cite{radford2016dcgan}, 
which introduced a fully convolutional network structure that made generators and 
discriminators more suitable for image data, improving the quality and stability 
of generated images. Subsequently, Arjovsky~\etal 
introduced WGAN~\cite{arjovsky2017wasserstein}, effectively alleviating gradient vanishing problems by 
incorporating the Earth Mover's distance, further enhancing GAN training stability and 
generative performance. 

In StyleGAN~\cite{karras2020stylegan}, Karras~\etal achieved 
more refined control over generated images by separating content and style information, 
significantly improving the diversity and quality of generated images. The researchers 
developed a novel architectural approach where stylistic elements are controlled 
through weight modulation in the convolutional layers:
\begin{equation}
w'_{ijk} = s_i \cdot w_{ijk}.
\end{equation}

This equation illustrates how the original weights $w_{ijk}$ are transformed 
into modulated weights $w'_{ijk}$ by applying the style-specific scaling 
factor $s_i$ to the $i$th input channel. The subscripts $j$ and $k$ 
identify the respective output channel and spatial position within 
the convolution operation. 

Following this modulation process, the resulting 
activations exhibit a standard deviation characterized by:
\begin{equation}
\sigma_j = \sqrt{\sum_{i,k} {(w'_{ijk})}^{2}}.
\end{equation}

To ensure computational stability and appropriate normalization, 
an additional scaling transformation is implemented:
\begin{equation}
w''_{ijk} = w'_{ijk} / \sqrt{\sum_{i,k} {(w'_{ijk})}^{2} + \epsilon},
\end{equation}
where the symbol $w''_{ijk}$ denotes the weights after complete normalization, 
while $\epsilon$ represents a small positive constant that prevents 
division by zero. Through this sophisticated adaptive normalization system, 
the network achieves fine-grained stylistic control while maintaining 
robust and consistent training behavior.

These improvements laid the foundation for GAN applications in CV, establishing 
them as excellent techniques for high-quality image generation~\cite{borjiProsConsGAN2019, caoRecentAdvancesGenerative2019}. Initially, GANs were 
primarily used for image generation tasks such as image synthesis and style transfer. GANs 
can meet the demands of image restoration, enhancement, and synthesis by training generators 
to produce images similar to real data~\cite{zhanSpatialFusionGAN2019}. The emergence of StyleGAN~\cite{karras2020stylegan} marked a major breakthrough 
in separating content and style in images, advancing fields like image editing and facial 
synthesis~\cite{richardsonEncodingStyleStyleGAN2021}. 

Additionally, GANs are widely applied to object detection and image classification 
tasks, where researchers combine generative networks with discriminators to enhance image 
feature learning, further improving model performance across various CV tasks~\cite{jolicoeur-martineauRelativisticDiscriminatorKey2018}.

\minew{Building on the success of generative adversarial networks in 
image generation and style transfer, Song~\etal proposed 
MLFF-GAN~\cite{song2022mlffgan}, which demonstrates how GANs can 
effectively address the unique challenges of remote sensing 
spatiotemporal fusion through architectural innovations and 
domain-specific adaptations. The MLFF-GAN generator employs a U-net-like~\cite{ronnebergerUNetConvolutionalNetworks2015}
architecture with three distinct stages: feature extraction using 
multilevel features (MLFs) to handle the substantial resolution 
difference between high-resolution (e.g., Landsat at 30m) and 
low-resolution (e.g., MODIS at 500m) imagery through hierarchical 
features extracted via multiple downsampling layers; feature fusion 
that addresses both global systematic differences and local abrupt 
changes through two complementary mechanisms—Adaptive Instance 
Normalization (AdaIN)~\cite{huang2017arbitrary} to handle global spectral differences and 
systematic errors between sensors by aligning the global distribution 
of high-resolution features with spectral characteristics at the 
prediction time while preserving spatial details, and the Attention 
Module (AM) that learns spatially-varying weights to handle local 
changes such as phenological variations or land cover changes through 
a dual-attention mechanism considering both spatial similarity and 
temporal changes; and image reconstruction. The discriminator adopts 
a PatchGAN~\cite{isola2017image} architecture that evaluates patches rather than the entire 
image, which is particularly effective for remote sensing imagery where 
local texture patterns are crucial, while the overall loss function 
combines adversarial loss with multiple content losses, including 
spectrum loss using cosine similarity to ensure spectral fidelity and 
structure loss employing MS-SSIM to preserve structural details.}

To solve spatial coherence problems that often occur when fusing images from different 
time points, Tan~\etal's GAN-STFM~\cite{tan2022flexible} offers a novel solution. Their seamless splicing mechanism 
specifically addresses the splicing gap issues common in traditional spatiotemporal fusion 
methods, significantly improving the visual quality and spectral consistency of generated 
images, demonstrating GAN's excellent performance in helping bridge spatial integrity and 
temporal dynamics.

Lei~\etal~\cite{lei2024hpltsgan} further refined the handling of spatiotemporal conflicts by introducing an 
Adaptive Spatial Distribution Transformation (ASDT)~\cite{jiGenerativeAdversarialNetworkBased2021}. In heterogeneous regions 
where balancing spatial details and temporal changes is particularly challenging, their 
model significantly improves spatiotemporal consistency by adaptively adjusting spatial 
distribution based on temporal context. This allows the model to maintain high spatial 
fidelity while effectively tracking temporal evolution, directly addressing the core 
contradiction in spatiotemporal fusion applications.

Zhang~\etal's DCDGAN-STF~\cite{zhang2024dcdgan} addresses spatiotemporal conflicts by introducing multi-scale 
deformable convolution distillation mechanisms and teacher-student correlation distillation 
mechanisms. Their deformable convolution mechanism captures spatiotemporal differences 
between multi-temporal images, enabling the model to flexibly extract features of irregularly 
shaped changes. Additionally, their knowledge distillation innovation uses KL divergence 
to measure feature similarity between teacher and student networks, significantly improving 
texture detail recovery while maintaining temporal prediction accuracy, demonstrating its 
balanced capability in processing spatiotemporal information.

\minew{
GAN-based spatiotemporal fusion methods excel at generating perceptually realistic images with sharp details, 
avoiding the over-smoothing of traditional methods. The adversarial training enables learning complex nonlinear 
relationships, while flexible architectures allow integration of domain-specific modules 
like AdaIN~\cite{huang2017arbitrary} for global 
corrections and attention mechanisms for local adaptations~\cite{song2022mlffgan}. These strengths make GANs 
particularly valuable for urban planning applications where high visual quality is essential for stakeholder 
communication and decision-making, and for detailed agricultural field boundary delineation where sharp edges 
are crucial for precision farming~\cite{tan2022flexible}. The ability to work with limited input pairs also 
benefits emergency response scenarios where reference images may be scarce due to cloud cover or rapid 
environmental changes~\cite{weng2024spatially}.

However, GANs face training instability, high computational costs, and may generate plausible but inaccurate 
details—a critical issue in scientific applications such as glacier monitoring or sea ice extent mapping where 
measurement accuracy is paramount~\cite{lei2024hpltsgan}. The black-box nature limits interpretability, which 
can be problematic for legal land use assessments or environmental compliance monitoring requiring traceable 
and defensible results. Training instability particularly affects operational deployment in national mapping 
agencies where consistent and reliable outputs are essential. The tendency to hallucinate details poses risks 
in disaster damage assessment, where generated features might be misinterpreted as actual structural damage 
or environmental changes, potentially leading to misallocation of relief resources~\cite{zhang2024dcdgan}.
}

\begin{figure*}[t]
    \centering
    \includegraphics[width=.75\textwidth]{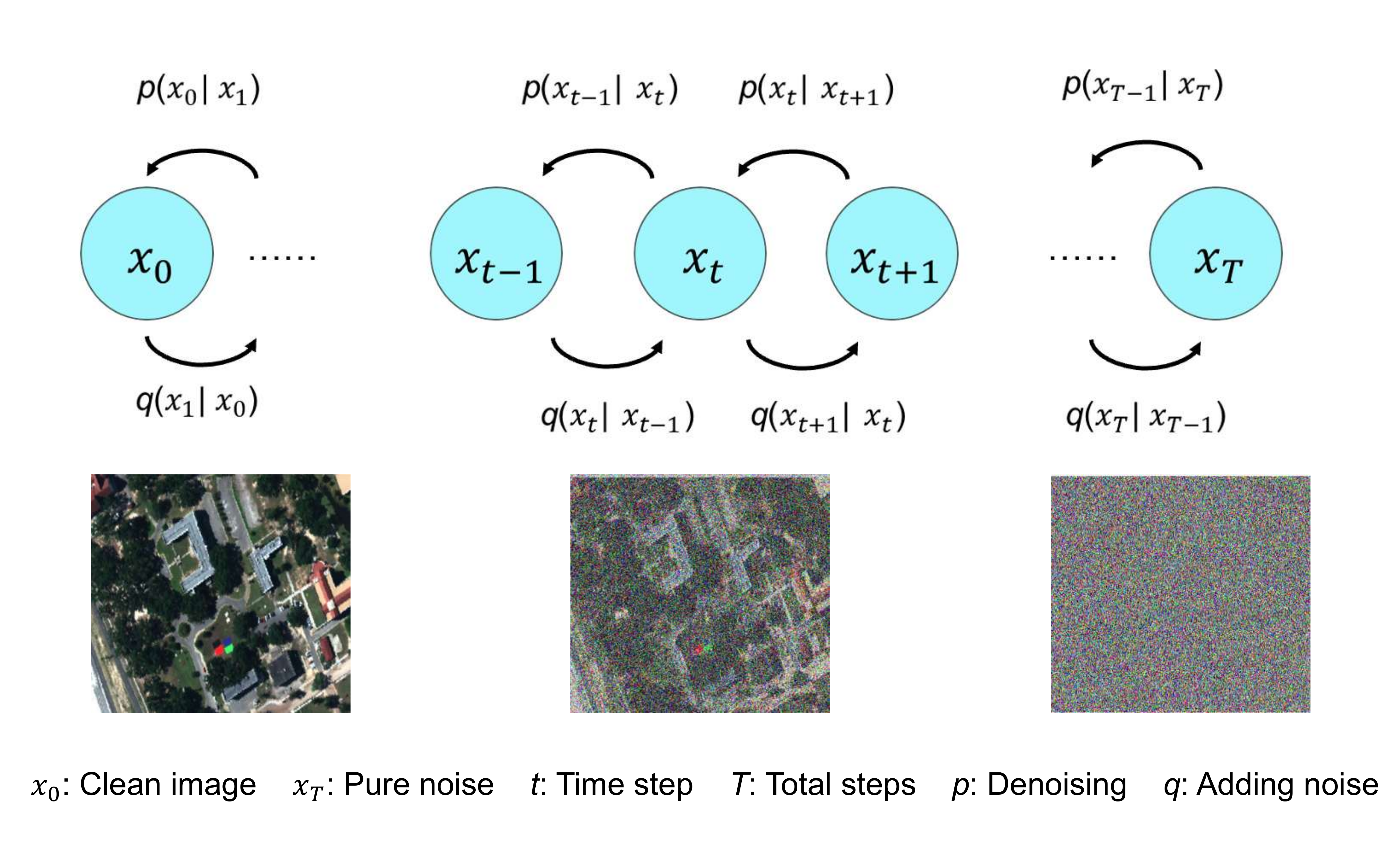}
    \caption{Forward and reverse process in diffusion models.}
    \label{fig:diffusion_process}
\end{figure*}

\noindent\textbf{Diffusion Models}. Diffusion models were actually developed 
inspired by diffusion processes in statistical physics, representing a 
breakthrough advancement in the field of generative modeling in recent years. 
As shown in \autoref{fig:diffusion_process}, diffusion models operate via 
two key processes: a forward process (denoted by $q(x_t|x_{t-1})$) that 
progressively adds noise to data samples, and a reverse process (denoted by 
$p(x_{t-1}|x_t)$) that learns to denoise and recover the original data. 
The figure shows the Markov chain from the clean image $x_0$ through 
intermediate noisy states ($x_{t-1}$, $x_t$, $x_{t+1}$) to the fully noisy 
state $x_T$. The bottom images demonstrate this process visually on a remote 
sensing example, where the left image shows the original clean data, the 
middle image shows a partially noised state, and the right image shows the 
completely noisy state resembling pure Gaussian noise. During training, the 
model learns to reverse this noise addition process to generate new data samples.

Watson~\etal used Denoising Diffusion Probabilistic Models (DDPM)~\cite{song2022denoising}, 
which established a forward diffusion process that systematically adds 
Gaussian noise to data through a Markov chain: 
\begin{equation}
q(x_t|x_{t-1}) = \mathcal{N}(x_t; \sqrt{1-\beta_t}x_{t-1}, \beta_t\mathbf{I}).
\end{equation}

Within this elegant framework, $q(x_t|x_{t-1})$ expresses the probability 
distribution of $x_t$ conditional on $x_{t-1}$, while $x_t$ indicates the 
noise-corrupted sample at time point $t$. The coefficient $\beta_t \in (0,1)$ 
serves as the noise intensity scheduler, and $\mathbf{I}$ represents the 
identity matrix. Here, $\mathcal{N}(\mu, \sigma^2)$ signifies a normal 
distribution characterized by mean $\mu$ and variance $\sigma^2$.

DDPM~\cite{song2022denoising} subsequently optimizes a neural architecture 
$\epsilon_\theta$ to reconstruct the reverse denoising trajectory by minimizing:
\begin{equation}
\mathcal{L} = \mathbb{E}_{t,x_0,\epsilon}[||\epsilon - \epsilon_\theta(x_t, t)||^2].
\end{equation}

In this optimization objective, $\mathcal{L}$ corresponds to the error metric, 
$\mathbb{E}$ signifies expectation, and $\epsilon_\theta$ denotes the neural 
predictor with trainable parameters $\theta$ designed to estimate noise 
components. The term $\epsilon \sim \mathcal{N}(0, \mathbf{I})$ refers to 
the random noise initially injected into the clean data $x_0$, while $t$ 
indicates the temporal position in the diffusion sequence, and $||\cdot||^2$ 
quantifies the squared Euclidean distance. This innovative approach has 
allowed diffusion-based architectures to surpass GANs in generation quality 
while avoiding their notorious training instabilities.

\minew{In the context of spatiotemporal fusion, DDPM's training process 
utilizes paired ground-truth data where high-resolution images at target 
dates serve as $x_0$~\cite{song2022denoising}. The model learns the noise distribution by comparing 
predicted noise $\epsilon_\theta(x_t, t)$ against actual noise applied to 
ground-truth samples, ensuring the reverse process generates outputs aligned 
with the true data distribution. This probabilistic framework naturally 
enables uncertainty quantification through multiple sampling runs, producing 
a distribution of possible fusion results. Model calibration is achieved by 
evaluating the mean of generated samples against held-out ground-truth using 
metrics like RMSE and SSIM, while sample variance provides uncertainty 
estimates across different spatial regions and temporal conditions.}

Although DDPM's~\cite{song2022denoising} generation quality is impressive, 
it requires multiple sampling steps, resulting in slower generation speed, 
which limits its use in practical applications. To address this issue, 
Song~\etal proposed Denoising Diffusion Implicit Models (DDIM)~\cite{song2022denoising}, 
which accelerated the sampling process through a non-Markovian diffusion process: 
\begin{equation}
\begin{aligned}
x_{t-1} &= \sqrt{\alpha_{t-1}}\left(\frac{x_t - \sqrt{1-\alpha_t}\epsilon_\theta(x_t, t)}{\sqrt{\alpha_t}}\right) \\
&\quad + \sqrt{1-\alpha_{t-1}-\sigma_t^2} \cdot \epsilon_\theta(x_t, t) + \sigma_t\epsilon
\end{aligned}
\end{equation}where $x_{t-1}$ represents the reconstructed sample at step $t-1$, $x_t$ 
denotes the current noisy input, and $\alpha_t = \prod_{i=1}^{t}(1-\beta_i)$ 
captures the cumulative effect of the noise schedule across timesteps. The 
function $\epsilon_\theta(x_t, t)$ embodies the neural network's prediction 
of noise components, while $\sigma_t$ functions as a control parameter 
governing the sampling process's stochastic properties, and 
$\epsilon \sim \mathcal{N}(0, \mathbf{I})$ introduces fresh Gaussian noise 
to maintain generation diversity, dramatically enhancing computational 
efficiency without sacrificing output quality.

Rombach~\etal introduced latent diffusion models~\cite{rombach2022latent}, 
further advancing the field by operating in a compressed latent space, 
greatly reducing computational complexity and making diffusion models more 
efficient in high-resolution image generation tasks.

In CV, the superior performance of diffusion models has gradually 
emerged~\cite{yangDiffusionModelsComprehensive2024, ulhaqEfficientDiffusionModels2024}. 
In tasks such as image restoration, style transfer, and super-resolution, 
diffusion models overcome some limitations of traditional generative models 
through their gradual denoising approach, generating clearer and more 
detailed images~\cite{bengesiAdvancementsGenerativeAI2024}. Subsequently, 
the efficiency and powerful detail recovery capabilities of diffusion models 
have also begun to be applied in other fields~\cite{hanEnhancingRemoteSensing2023,liuDiffusionModelsMeet2024}.

In remote sensing spatiotemporal fusion research, diffusion models provide 
new methods for resolving the fundamental conflict between temporal and 
spatial dimensions through gradual denoising. Previous methods typically 
struggled to maintain both detailed spatial features and accurate temporal 
dynamics when fusing HSLT and HTLS data, while diffusion models offer a 
natural framework to reconcile this contradiction. Ma~\etal proposed 
DiffSTF~\cite{ma2024conditional}, designing a conditional diffusion model 
specifically for spatiotemporal fusion to address spatiotemporal conflicts. 
They recognized that the inherent gradual denoising process of diffusion 
models provides a natural framework for harmonizing temporal and spatial 
information.

By combining an encoder-decoder structure with residual blocks and Transformer 
blocks, DiffSTF~\cite{ma2024conditional} effectively captures both global temporal patterns and local 
spatial features simultaneously, resolving the balance between spatial 
resolution and temporal dynamics more effectively than previous methods.

To further optimize the handling of spatiotemporal conflicts, Huang~\etal 
proposed the STFDiff~\cite{huang2024stfdiff}. During the noise prediction process, the tension 
between spatial and temporal information becomes particularly evident. They 
developed a dual-stream encoder specifically designed to process spatial 
and temporal information through independent but interconnected paths, 
allowing the model to maintain the features of each dimension while 
exchanging cross-dimensional information at key points in the processing pipeline. 

This approach combines innovative temporal embedding techniques 
that transform discrete time variables into continuous vector representations. 
By incorporating efficient noise prediction modules, STFDiff~\cite{huang2024stfdiff} achieves a more 
balanced fusion of spatial details and temporal dynamics, significantly 
improving the structural similarity index and spectral fidelity across 
multiple datasets.

Diffusion-based models optimize the efficiency and effectiveness of the 
fusion process by introducing multi-scale feature extraction and dynamic 
modules aimed at harmonizing spatial and temporal information. Unlike 
previous approaches that viewed spatiotemporal conflicts as limitations 
to be managed, these diffusion models consider this tension as a core part 
of their design, gradually harmonizing spatial and temporal features through 
an iterative denoising process. This undoubtedly demonstrates the powerful 
potential of diffusion models in resolving spatiotemporal conflicts in 
remote sensing spatiotemporal fusion.

\subsubsection{Sequence Models}
Sequence models can be used to efficiently simulate dynamic changes in time series 
through their recursive structure and memory mechanisms, making them widely used 
in spatiotemporal fusion research~\cite{tasnimNovelMultiModuleApproach2022}. The development of sequence models mainly includes 
Recurrent Neural Networks (RNNs)~\cite{li2018indrnn} and their enhanced versions, which include Gated 
Recurrent Units (GRU)~\cite{sherstinsky2020fundamentals} and Long Short-Term Memory (LSTM) networks~\cite{yaoImprovedLSTMStructure2018}. Such models, 
which are very effective in time series modeling, provide strong technical support 
for handling long-term dependencies and predicting dynamic changes.

\noindent\textbf{Recurrent Neural Network}. RNNs originated in the 1980s and are a form of neural 
network structure that recursively passes hidden layer states. The main advantage 
of RNNs lies in their capacity to capture time dependencies in input data. 
The typical RNN formulation can be written as:
\begin{equation}
h_t = \sigma(W_h h_{t-1} + W_x x_t + b).
\end{equation}

In this concise formulation, $h_t$ denotes the network's internal representation at temporal point $t$, 
while $h_{t-1}$ captures the memory state from the previous step. The variable $x_t$ encodes 
the current input information, and the transformation matrices $W_h$ and $W_x$ govern the 
influence of historical context and new data respectively. The vector $b$ introduces 
learning flexibility, and $\sigma$ represents the non-linear transformation function 
(commonly implemented as tanh or ReLU) that enables the network to capture complex patterns.
Traditional RNNs, however, struggle with processing 
long time series information due to issues such as vanishing and exploding gradients~\cite{li2018indrnn}.

With the effective application of RNNs in sectors such as natural language processing, 
scholars suggested many improved models to increase their computing efficiency and 
memory capabilities~\cite{chaitanyabharathiinstituteoftechnologyautonomousSurveyRecurrentNeural2017,reddyGeneratingNaturalLanguage2017}. 
For instance, Bidirectional RNNs (such as Bidirectional LSTM) 
enhance the model's understanding of sequence context by processing both forward 
and backward time dependencies. GRU~\cite{sherstinsky2020fundamentals} simplifies the gating structure, reducing model 
parameters and improving computational efficiency~\cite{kashid2023bilstm}. These improvements not only increase 
the model's performance in sequence modeling activities but also set the basis for 
their use in CV tasks such as video analysis and action recognition~\cite{sonderby2020metnet}.

In remote sensing, RNNs and their variants offer a distinctive approach to the 
conflict between temporal and spatial dimensions. Unlike other models that often 
struggle to balance high spatial resolution with accurate temporal dynamics, RNNs 
naturally prioritize the temporal aspect through their sequential processing 
architecture. This temporal focus provides an alternative perspective on the HSLT 
and HTLS tension inherent in remote sensing data. 

Models like STUNNER~\cite{fang2023stunner} address this 
time-space conflict by introducing a dual-stream structure combined with Time 
Difference Networks (TDN)~\cite{suttonTemporalDifferenceNetworks2004} and 
Spatiotemporal Trajectory Networks (STTN)~\cite{ivanovicTrajectronProbabilisticMultiAgent2019}.
This architecture effectively reconciles temporal and spatial priorities by separating 
their processing pathways, allowing STUNNER to efficiently model non-stationary 
sequences and short-term instantaneous changes while preserving spatial context. 
In particular, the TDN~\cite{suttonTemporalDifferenceNetworks2004} component employs a stacked TDiff-LSTM structure to model 
stationarity in time series through layer-by-layer feature differencing. 

This dual-stream approach creates a more balanced relationship between temporal evolution 
and spatial integrity, significantly outperforming mainstream models such as TrajGRU~\cite{zhang2023graphatnet}, 
PredRNN~\cite{wang2017predrnn}, and MetNet~\cite{fang2023stunner}. These advances in RNN architectures provide effective guidance for 
addressing the fundamental temporal-spatial conflict in remote sensing. 
By optimizing temporal information flow while preserving spatial context, 
these models enable the development of more accurate and robust fusion 
technologies that better balance the competing requirements of temporal 
dynamics and spatial detail preservation.

\noindent\textbf{Long Short-Term Memory}. Long Short-Term Memory (LSTM) networks have significantly surpassed traditional
Recurrent Neural Networks (RNNs) in handling long-term dependencies, thereby enhancing
the performance of sequence models~\cite{yaoImprovedLSTMStructure2018}. Their advanced
gating mechanisms enable efficient processing of extended sequences, overcoming the
limitations of conventional RNNs~\cite{hochreiter1997lstm}. Continuous refinements in
architecture have expanded their applicability to more complex tasks.
LSTM and its variants offer a sophisticated solution to the inherent temporal-spatial
conflict in remote sensing spatiotemporal fusion. Their selective memory capabilities
provide a distinct advantage in accurately capturing temporal evolution, which helps
reconcile the challenge of maintaining high spatial resolution with temporal precision.
These models excel at balancing competing demands in fusion tasks.

LSTM networks are mainly composed of three dedicated gate units: input gate, forget gate, 
and output gate, which control the flow of information in a selective manner. 
The architectural brilliance of LSTM networks emerges from their innovative triple-gate 
mechanism (input ($i_t$), forget ($f_t$), and output ($o_t$) gates), orchestrating precise 
information management. The mathematical machinery of LSTM with peephole connections 
unfolds through these relationships:
\begin{gather}
f_t = \sigma(W_{xf}x_t + W_{hf}H_{t-1} + W_{cf} \odot C_{t-1} + b_f), \\
C_t = f_t \odot C_{t-1} + i_t \odot \tanh(W_{xc}x_t + W_{hc}H_{t-1} + b_c),
\end{gather}

where $f_t$ embodies the forgetting mechanism determining which historical 
information should be retained or discarded, while $i_t$ functions as the information gatekeeper 
controlling the flow of new data into the memory. The term $C_t$ represents the updated memory 
reservoir that preserves long-term dependencies, building upon its previous state $C_{t-1}$. 
The activation function $\sigma$ constrains outputs to the range (0,1), creating continuous 
gating behaviors, while $\tanh$ normalizes values between -1 and 1. The notation $\odot$ 
indicates element-wise multiplication, enabling selective information filtering. The parameters 
$W_{xf}$, $W_{hf}$, $W_{cf}$, $W_{xc}$, and $W_{hc}$ constitute learnable transformation matrices 
for their respective pathways, with $x_t$ representing the current input vector, $H_{t-1}$ 
capturing previous output state, and $b_f$ and $b_c$ providing adjustable bias values. 
Particularly noteworthy is the peephole connection component ($W_{cf} \odot C_{t-1}$), which 
creates direct memory access pathways, substantially enhancing the network's capacity to 
capture precise temporal dependencies.

This somewhat intricate gating mechanism proves ideal for spatiotemporal fusion tasks that require a careful balance 
between spatial and temporal resolutions. L-UNet~\cite{sun2022lunet} combines convolutional layers, which are 
responsible for extracting spatial features, with LSTM layers, merging complementary architectural strengths to achieve 
effective spatiotemporal modeling while maintaining a delicate equilibrium.

ConvLSTM~\cite{cui2024soil} fuses the LSTM framework with convolutional operations. By preserving LSTM's robust 
temporal memory and leveraging convolutions to capture spatial dependencies, it offers a unified solution for challenging 
applications such as soil freeze/thaw monitoring. As dataset sizes increase and environmental complexity intensifies, 
however, ConvLSTM~\cite{cui2024soil} faces practical issues related to computational resources and model intricacy when 
managing highly dynamic spatiotemporal interactions.

Geo-BiLSTM~\cite{jia2024forecasting} combines direct LSTM modules with an attention mechanism to fine-tune the balance 
between spatial and temporal priorities. Its attention process assigns adaptive weights to spatial features based on 
the relevance of different temporal orders, while a bidirectional design captures context from both earlier and later 
time steps. This composite approach cultivates a refined equilibrium between dimensions, thereby boosting the model's 
capacity to manage intricate spatiotemporal relationships while preserving all essential details. This advantage proves 
significant for predicting remote sensing data.

\subsubsection{Other Models}
Beyond these mainstream models, various architectures have demonstrated unique 
advantages in remote sensing spatiotemporal fusion, including Graph Neural 
Networks (GNN)~\cite{liMultiLabelRemoteSensing2020}, Multilayer Perceptrons 
(MLP)~\cite{chen2023stfmlp}, and Dual-branch Fusion Networks~\cite{sun2023dbfnet}. 
These alternative approaches complement the capabilities of sequence-based 
models, offering specialized solutions for particular spatiotemporal fusion 
challenges.

GNNs can model spatial topological relationships between data through graph 
structures, making them suitable for tasks involving regional interactions 
and spatiotemporal dependencies~\cite{shi2020pointgnn}. 
STHGCN~\cite{wang2022sthgcn} proposed by Wang~\etal combines 
multiple graph convolution modules with dynamic higher-order temporal difference 
convolution modules, effectively extracting higher-order spatial and temporal 
dependencies and greatly improving traffic flow prediction accuracy. Gao~\etal's 
SMA-Hyper~\cite{gao2024smahyper} addresses the limitations of STHGCN~\cite{wang2022sthgcn} in handling multi-view relationships 
and sparse data. SMA-Hyper~\cite{gao2024smahyper} adopts an adaptive multi-view hypergraph architecture 
with attention mechanisms to dynamically model higher-order cross-regional 
dependencies, significantly improving prediction performance on sparse data 
and enhancing model generalization capability and prediction accuracy.

As lightweight neural network architectures, MLPs offer extremely high computational 
efficiency with fewer parameter requirements. The STFMLP~\cite{chen2023stfmlp} proposed by Chen~\etal 
replaces traditional CNNs for feature extraction with multi-layer perceptrons, 
combining Feature Pyramid Networks (FPN)~\cite{linFeaturePyramidNetworks2017} and temporal difference constraints to 
effectively enhance multi-scale feature extraction and prediction performance. 
This approach significantly reduces computational complexity, overcoming the 
computational resource consumption problems of traditional CNNs when modeling 
complex spatiotemporal relationships, providing a solid foundation for MLP-based 
complex tasks~\cite{yanMultihourMultisiteAir2021, zhangMLPSTMLPAll2023}.

Dual-branch Fusion Networks achieve exceptional modeling accuracy in multi-source 
remote sensing data fusion by processing features from different dimensions (such 
as space and time) separately. The DSTFNet~\cite{cai2023dualbranch} proposed by Cai~\etal combines very high 
resolution (VHR)~\cite{zhang2024wuhan} images with medium resolution (MRSITS) data~\cite{radeloffNeedVisionGlobal2024}, using independent 
spatial and temporal branches to extract dynamic texture and spectral features, 
while optimizing the feature fusion process through attention mechanisms. This 
dual-branch structure effectively integrates multi-source data, overcoming the 
limitations of single-branch models in feature extraction and fusion, greatly 
improving the accuracy of farmland boundary detection and zoning tasks, and 
demonstrating its generalization capability and robustness across different regions~\cite{sun2023dbfnet}.

\vspace{0.5cm}
To better showcase the advantages, disadvantages, and potential improvements of 
various deep learning architectures in remote sensing spatiotemporal fusion tasks, 
\autoref{tab:network_comparison} provides a comprehensive comparison.

\minew{
\begin{table*}[htb!]
\caption{Comparison of network architectures for remote sensing spatiotemporal fusion.}
\label{tab:network_comparison}
\small
\begin{tabular*}{\textwidth}{@{\extracolsep{\fill}} 
p{0.15\textwidth} 
>{\raggedright\arraybackslash}p{0.25\textwidth} 
>{\raggedright\arraybackslash}p{0.25\textwidth} 
>{\raggedright\arraybackslash}p{0.25\textwidth} @{}}
\toprule
\textbf{Network} & \textbf{Advantages} & \textbf{Disadvantages} & \textbf{Potential Improvements} \\
\midrule
CNN & 
Strong spatial feature extraction. Efficient parameter sharing. Well-established in remote sensing. & 
Limited for long-range dependencies. Poor with non-linear changes. High computational costs for high-resolution images. & 
Integrate attention mechanisms. Improve multi-scale feature fusion. Introduce deformable convolutions. Develop specialized loss functions. \\
\midrule
Transformer & 
Excellent for long-range dependencies. Effective global context processing. Superior in multi-temporal data fusion. & 
Quadratic complexity with image size. High memory requirements. Requires more data than CNNs. & 
Develop hierarchical processing. Create hybrid CNN-Transformer models. Design efficient attention computation. Improve detail preservation. \\
\midrule
GAN & 
High-quality image generation. Handles complex transformations. Strong adaptability to heterogeneous data. & 
Training instability issues. High computational demands. Often prioritizes visual over spectral quality. & 
Implement stable training strategies. Improve spectral fidelity. Combine with attention mechanisms. Develop distillation approaches. \\
\midrule
Diffusion Models & 
Superior detail restoration. Stable training. High-quality outputs with better structure preservation. & 
Slow generation process. High computational demands. Limited real-time applicability. Complex parameter tuning. & 
Develop accelerated sampling. Design latent-space models. Combine with encoder-decoder structures. Research conditional models. \\
\midrule
Sequence Models & 
Excellent time series modeling. Effective for temporal predictions. Strong in change detection with efficient training. & 
Inefficient with spatial data. Limited capacity for long sequences. Less effective with complex spatial structures. & 
Combine with convolutions. Introduce bidirectional structures. Integrate attention mechanisms. Develop multi-stream architectures. \\
\bottomrule
\end{tabular*}
\end{table*}
}

\subsection{Evaluation Metrics}
Evaluation metrics measure the accuracy of model predictions and provide benchmarks 
for comparing different STF methods~\cite{li2024incorporating,lyu2024multistream,cui2024novel}. 
\autoref{tab:evaluation_metrics} lists some commonly used evaluation metrics in this field.

\begin{table*}[htb!]
\caption{Common Evaluation Metrics for Remote Sensing Spatiotemporal Fusion.}
\label{tab:evaluation_metrics}
\normalsize   
\begin{tabular*}{\textwidth}{@{\extracolsep{\fill}} 
p{0.08\textwidth} 
>{\raggedright\arraybackslash}p{0.32\textwidth} 
>{\raggedright\arraybackslash}p{0.25\textwidth} 
>{\raggedright\arraybackslash}p{0.25\textwidth} @{}}
\toprule
\textbf{Metric} & \textbf{Description} & \textbf{Advantages} & \textbf{Limitations} \\
\midrule
MSE & 
Average squared differences between predicted and actual values for regression tasks. & 
Simple calculation, penalizes large errors effectively. & 
Overly sensitive to outliers, unintuitive units. \\
\midrule
RMSE & 
Square root of MSE, used in fusion quality assessment. & 
Same units as original data, intuitive interpretation. & 
Disproportionately emphasizes large errors. \\
\midrule
NRMSE & 
RMSE normalized by data range for cross-dataset comparison. & 
Enables comparison across different scales and sensors. & 
Depends on maximum and minimum values. \\
\midrule
PSNR & 
Logarithmic signal-to-noise ratio for image quality assessment. & 
Standard metric for image quality with established thresholds. & 
Poor correlation with human visual perception. \\
\midrule
MAE & 
Average absolute differences, suitable for outlier-prone data. & 
More robust to outliers than RMSE, equal error weighting. & 
Gradient issues during optimization. \\
\midrule
SSIM & 
Structural similarity based on luminance, contrast, and structure. & 
Aligned with human perception of image quality. & 
Computationally complex, less sensitive to certain degradations. \\
\midrule
PCC & 
Linear correlation measurement for image similarity assessment. & 
Clear [-1,1] range, captures global correlation patterns. & 
Only detects linear relationships. \\
\midrule
$R^2$ & 
Proportion of variance explained in regression evaluation. & 
Direct interpretation of model fit quality. & 
Unstable with small samples, misleading when assumptions violated. \\
\midrule
SAM & 
Spectral angle measurement for spectral analysis. & 
Preserves spectral signatures, invariant to illumination scaling. & 
Ignores magnitude differences between spectra. \\
\bottomrule
\end{tabular*}
\end{table*}

\subsubsection*{2.4.1. Root Mean Square Error}
\begin{equation}
RMSE = \sqrt{\frac{1}{n}\sum_{i=1}^n(y_i - \hat{y_i})^2},
\end{equation}
where $y_i$ stands for the actual value, $\hat{y_i}$ indicates the predicted value, 
and $n$ signifies the total number of samples considered.

RMSE quantifies the difference between predicted and actual values, measuring the 
magnitude of prediction errors~\cite{wu2022dasftot,cheng2022stfegfa}. Lower RMSE values indicate more accurate predictions, 
with values approaching zero signifying minimal differences between predicted and 
actual values~\cite{wangLandsat8Sentinel2Image2024,luppino2024codealigned}.

\subsubsection*{2.4.2. Peak Signal-to-Noise Ratio}
\begin{equation}
PSNR = 10 \cdot \log_{10}\left(\frac{MAX_I^2}{MSE}\right),
\end{equation}
where $MAX_I$ denotes the maximum possible pixel value 
in the image, while $MSE$ refers to the Mean Squared Error between the reconstructed and 
reference images.

PSNR represents the ratio between the maximum possible power of a signal and the power of 
corrupting noise affecting its representation quality~\cite{chen2022swinstfm,chen2022multiscale_srs}. This metric is frequently employed 
to measure signal reconstruction quality in domains such as image compression~\cite{zhu2022hcnn,ao2022deep}.

\subsubsection*{2.4.3. Mean Absolute Error}
\begin{equation}
MAE = \frac{1}{n}\sum_{i=1}^n|y_i - \hat{y_i}|,
\end{equation}
where $|y_i - \hat{y_i}|$ signifies the absolute difference between the actual 
value $y_i$ and the predicted value $\hat{y_i}$ for the $i$-th sample, while $n$ represents 
the total number of samples.

MAE represents the average of absolute errors between predicted and observed values~\cite{su2022transformer}. 
Unlike RMSE, MAE exhibits lower sensitivity to outliers, making it particularly suitable 
for datasets containing a limited number of extreme values~\cite{song2022spatiotemporal,cuiPixelWiseEnsembledMasked2024}.

\subsubsection*{2.4.4. Structural Similarity Index}
\begin{equation}
SSIM(x,y) = \frac{(2\mu_x\mu_y + C_1)(2\sigma_{xy} + C_2)}{(\mu_x^2 + \mu_y^2 + C_1)
(\sigma_x^2 + \sigma_y^2 + C_2)}.
\end{equation}

The parameters of this equation are as follows: $\mu_x$ and $\mu_y$ represent the mean 
intensities, $\sigma_x^2$ and $\sigma_y^2$ correspond to the variances, and $\sigma_{xy}$ 
describes the covariance between images $x$ and $y$. The constants $C_1$ and $C_2$ are 
incorporated to maintain numerical stability.

SSIM evaluates the similarity between two images and is commonly used to measure the 
similarity between images before and after distortion, as well as to assess the 
authenticity of model-generated images~\cite{sun2022lunet,tan2022flexible,yang2022msfusion}.

\subsubsection*{2.4.5. Spectral Angle Mapper}
\begin{equation}
SAM = \arccos\left(\frac{\sum_{i=1}^n x_iy_i}{\sqrt{\sum_{i=1}^n x_i^2}
\sqrt{\sum_{i=1}^n y_i^2}}\right).
\end{equation}

For this metric, $x_i$ and $y_i$ represent the spectral vectors of the predicted and reference 
images, respectively, while $n$ indicates the number of spectral bands considered in the 
analysis.

SAM is applicable for evaluating spectral consistency in multispectral and hyperspectral 
remote sensing applications~\cite{shang2022spatiotemporal,ao2021constructing}.

\subsubsection*{2.4.6. Relative Dimensionless Global Error in Synthesis}
\begin{equation}
ERGAS = 100 \cdot \frac{h}{l} \sqrt{\frac{1}{N}\sum_{i=1}^N\left(\frac{RMSE_i}{\mu_i}\right)^2}.
\end{equation}

The equation utilizes several variables: $h$ and $l$ denote the spatial resolution of high 
and low-resolution images respectively, $N$ symbolizes the number of spectral bands, $RMSE_i$ 
represents the Root Mean Square Error for band $i$, and $\mu_i$ corresponds to the mean value 
of the $i$-th band in the reference image.

ERGAS is frequently used to assess remote sensing image quality, typically expressed as a 
percentage~\cite{chen2021cyclegan,jia2021hybrid}. Lower ERGAS values indicate higher image 
quality~\cite{kattenborn2021cnn,li2021multi}. This metric considers mean 
square error, root mean square error, and brightness information to provide a comprehensive 
assessment of image processing or compression algorithm performance~\cite{ma2021explicit,wei2021enblending}.

When comparing and optimizing model performance, researchers rely on evaluation metric systems~\cite{simonyan2015very,tan2022robust}. 
By selecting and combining these common evaluation metrics, researchers can comprehensively 
assess a model's spatiotemporal reconstruction capabilities from multiple perspectives, 
compare the advantages and disadvantages of different models, and support 
further optimization of spatiotemporal fusion algorithms~\cite{song2018spatiotemporal,chen2022swinstfm,tan2022robust}.

\subsection{Applications of Spatiotemporal Fusion}
Spatiotemporal fusion combines images of varying spatial and temporal resolutions to predict high-resolution outputs, a 
capability that is critical for remote sensing data processing in urban planning, disaster assessment, ecological monitoring, 
and climate change research~\cite{belgiu2019spatiotemporal,zhuSpatiotemporalFusionMultisource2018}.

\subsubsection{{Ecology Monitoring}}
In agricultural monitoring, Xiao~\etal~\cite{xiao2023review} describe how this technique integrates MODIS's low spatial 
but high temporal resolution data with Landsat's high spatial but low temporal resolution imagery, yielding outputs that 
capture both fine spatial detail and dynamic temporal changes. This integration proves invaluable for tracking crop 
growth and developmental stage transitions~\cite{liuHybridSpatiotemporalFusion2024a}. Li~\etal~\cite{li2020overview} further highlight its efficacy in capturing 
the dynamic evolution of agricultural land, particularly when monitoring crop cycles and land-use modifications.
Focusing on broader ecosystem surveillance, shifts in forest cover and species distribution often command primary 
attention. Belgiu and Stein~\cite{belgiu2019spatiotemporal} assert that spatiotemporal fusion methods adeptly bridge 
data gaps caused by cloud cover, a benefit that is indispensable in tropical regions where such obstructions are frequent.

\subsubsection{{Urban Planning}}
Spatiotemporal fusion methods have also gained attention in urban monitoring 
and planning. Since satellite images in urban areas require high spatial and 
temporal resolution, spatiotemporal fusion can provide more precise data for 
urban expansion monitoring. By combining HTLS data with HSLT data, urban land use and expansion can be tracked more effectively, 
providing decision support for urban planning~\cite{zhang2024wuhan}.

\subsubsection{{Disaster Monitoring and Assessment}}
Spatiotemporal fusion technology demonstrates exceptional value throughout the 
full cycle of natural disaster monitoring and assessment, particularly in the refined monitoring 
of sudden disasters such as floods, droughts, and forest fires~\cite{guo2024obsum}. Through intelligent 
coordination of heterogeneous spatiotemporal resolution data, this technology 
rapidly reconstructs high-precision impact assessment maps during the critical 
post-disaster period. In forest fire ecological recovery monitoring, Xiao and 
colleagues have verified that fused high-resolution remote sensing images can 
precisely track vegetation succession trajectories in burned areas~\cite{xiao2023review}. 
Even more groundbreaking is the technology's ability to construct dynamic damage assessment heat maps based on 
coordinated observation systems, providing spatial decision support for flood emergency 
response at the minute level. This represents a revolutionary advancement that has significantly enhanced the 
efficiency of traditional damage assessment~\cite{chen2023robot, chenTerraMultimodalSpatioTemporal2024}.

\subsubsection{{Climate Change Research}}
Against the backdrop of accelerating global climate system evolution, spatiotemporal 
fusion technology has become a crucial key to decoding climate-environment feedback 
mechanisms~\cite{zhao2024enso}. Its distinctive advantage lies in reconstructing global environmental 
monitoring sequence data with both spatial and temporal precision, thereby providing multidimensional 
observational benchmarks for climate sensitivity analysis. Recent studies have shown that this technology 
not only captures the spatiotemporal heterogeneity of large-scale temperature and humidity fields but also 
constructs ecological response fingerprint maps. This ability 
to decouple climate signals from environmental feedback provides a novel perspective for 
revealing the nonlinear interaction mechanisms between climate and ecosystems in the 
context of the geologic era dominated by contemporary human activities~\cite{kim2023multisource}.

\section{Experiments}\label{sec:experiments}
\begin{figure*}[htb!]
\centering
\includegraphics[width=\textwidth]{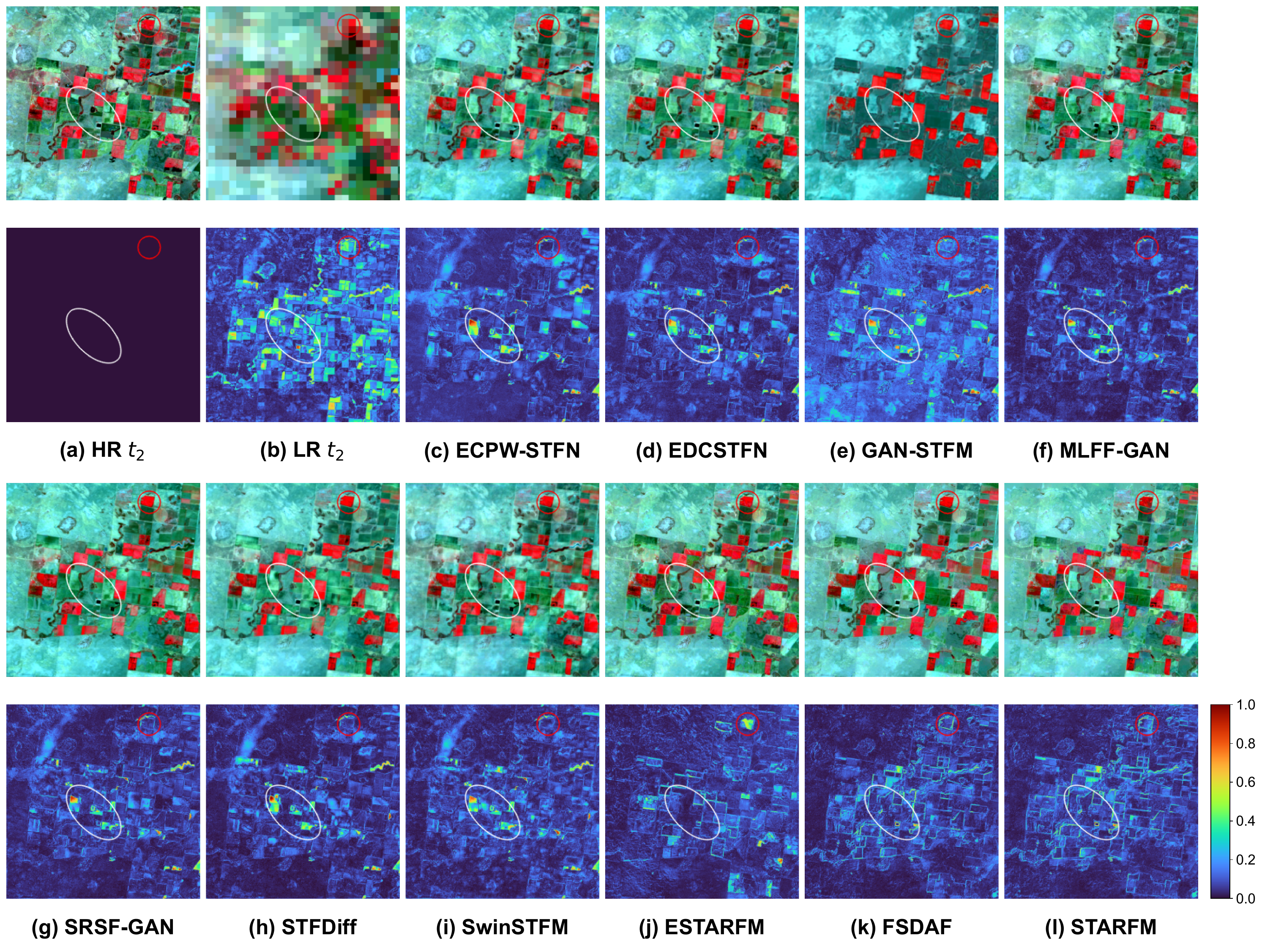}
\caption{Visual comparison of different spatiotemporal fusion methods 
on the CIA dataset. The first and third rows show ground truth and 
predicted false color composite images, while the second and fourth 
rows display difference maps.}
\label{fig:cia_visual_comparison}
\end{figure*}

\begin{figure*}[htb!]
\centering
\includegraphics[width=\textwidth]{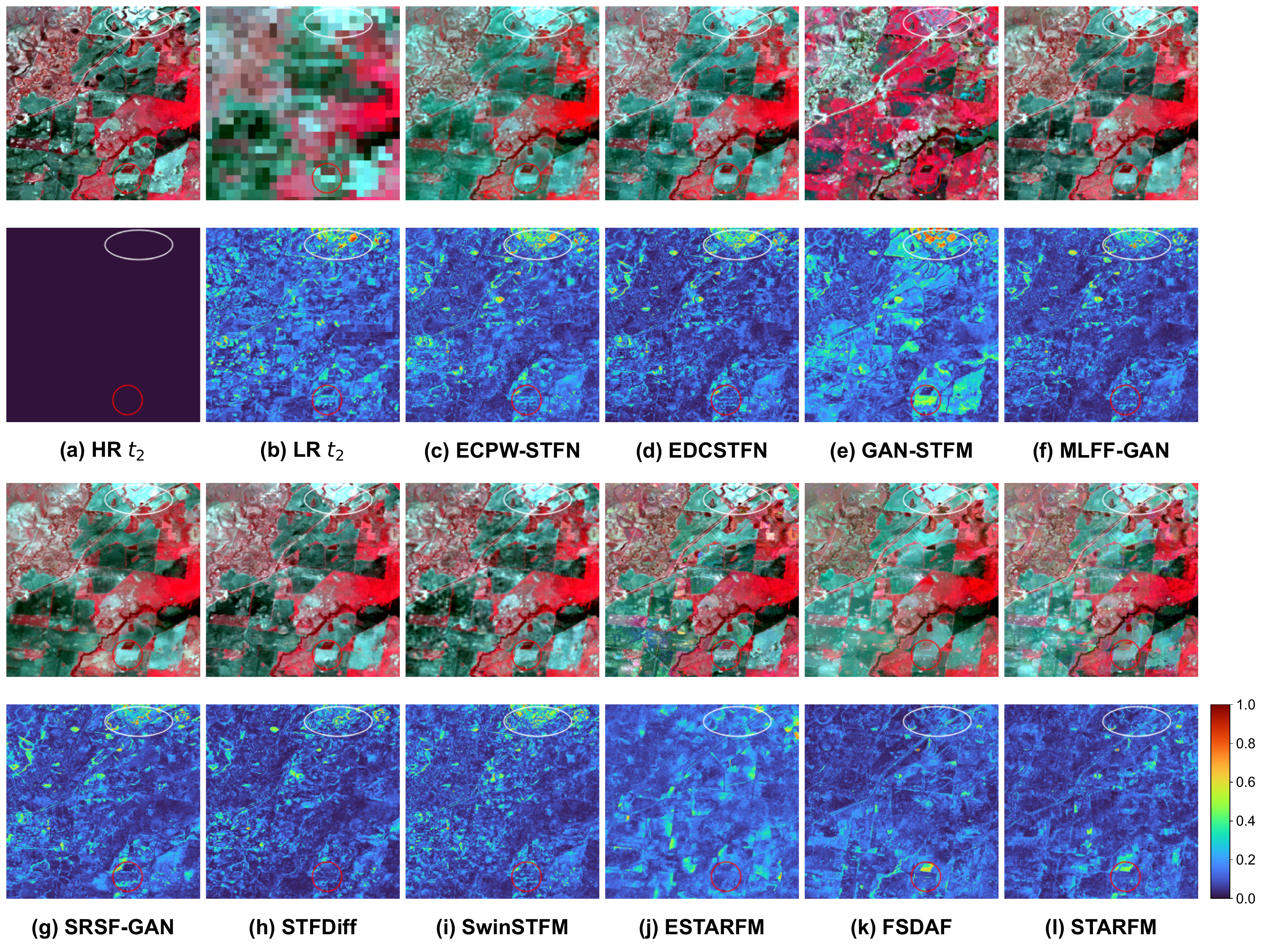}
\caption{Visual comparison of different spatiotemporal fusion methods 
on the LGC dataset. The first and third rows show ground truth and 
predicted false color composite images, while the second and fourth 
rows display difference maps.}
\label{fig:lgc_visual_comparison}
\end{figure*}

\begin{figure*}[htb!]
\centering
\includegraphics[width=\textwidth]{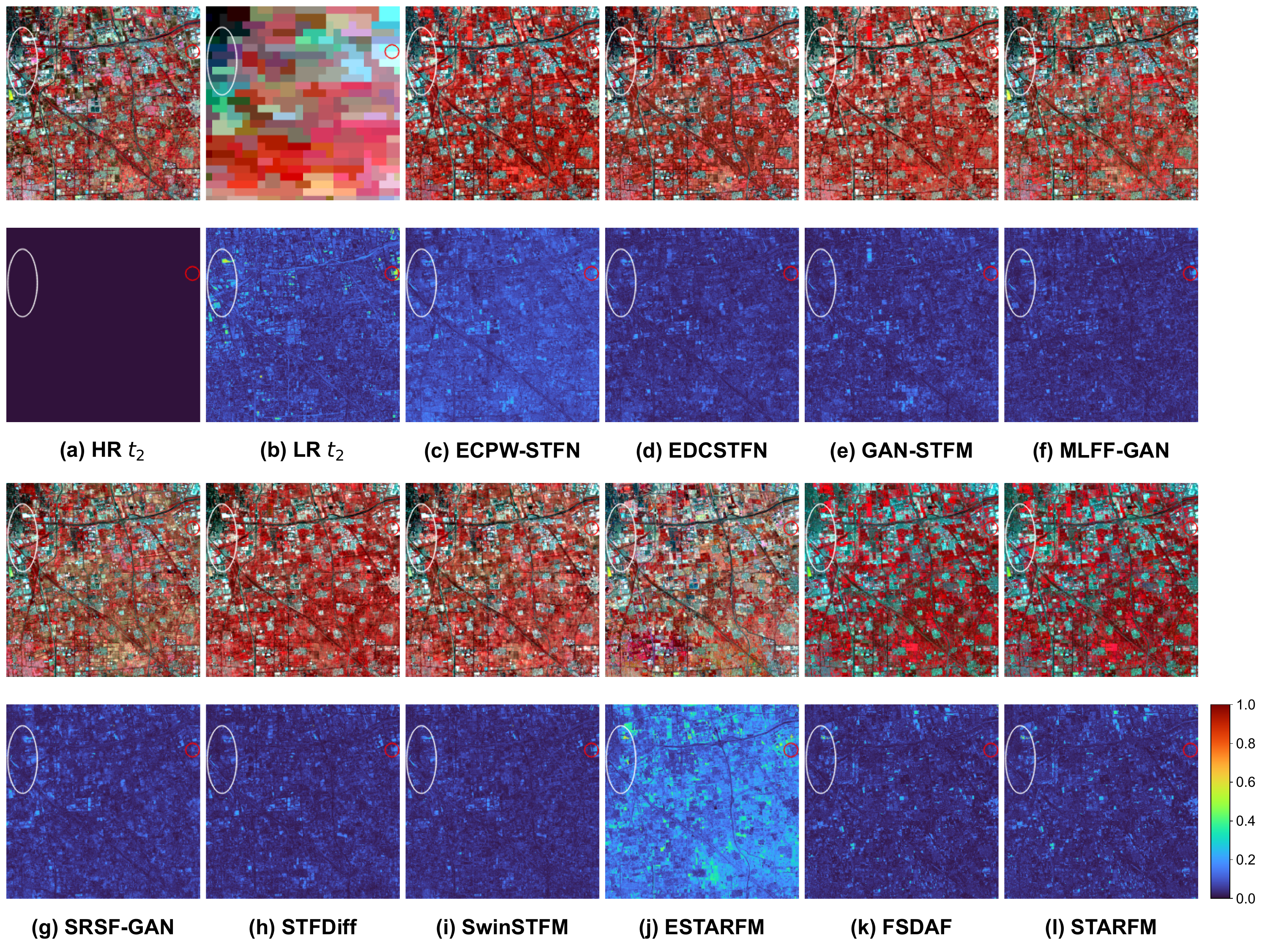}
\caption{Visual comparison of different spatiotemporal fusion methods 
on the DX dataset. The first and third rows show ground truth and 
predicted false color composite images, while the second and fourth 
rows display difference maps.}
\label{fig:dx_visual_comparison}
\end{figure*}

To comprehensively evaluate the performance of deep learning methods in remote 
sensing spatiotemporal fusion, we conducted systematic comparative experiments 
on ten representative methods across seven benchmark datasets.

\subsection{Experimental Setup}\label{subsec:experimental_setup}

\subsubsection{Datasets}\label{subsubsec:datasets}

The experiments utilized seven widely-used spatiotemporal fusion datasets 
covering diverse geographical regions and land cover types: CIA (Coleambally 
Irrigation Area)~\cite{zhang2024enhanced}, LGC (Lower Gwydir Catchment)~\cite{ma2024conditional}, 
DX (Daxing)~\cite{li2020overview}, TJ (Tianjin)~\cite{li2020overview}, 
WH (Wuhan)~\cite{zhang2024dcdgan}, BC (Butte County)~\cite{guo2024obsum}, 
and IC (Imperial County)~\cite{guo2024obsum}. These datasets vary 
in scale and complexity, with the number of image pairs ranging from 5 (IC) 
to 29 (DX). Detailed descriptions of these datasets are provided in 
Section~\ref{subsec:benchmark_datasets}.

For each dataset, we partitioned the data into training and testing sets 
following temporal order to simulate real-world application scenarios. 
\autoref{tab:dataset_split} summarizes the partition configuration for 
each dataset.

\begin{table}[htbp!]
\centering
\captionsetup{justification=centering}
\caption{Dataset partition configuration.}
\label{tab:dataset_split}
\small
\begin{tabular}{lccccccc}
\toprule
\textbf{Dataset} & \textbf{CIA} & \textbf{LGC} & \textbf{DX} & 
\textbf{TJ} & \textbf{WH} & \textbf{BC} & \textbf{IC} \\
\midrule
Total Pairs & 17 & 14 & 29 & 27 & 8 & 6 & 5 \\
Training Set & 12 & 10 & 21 & 21 & 6 & 4 & 4 \\
Testing Set & 5 & 4 & 8 & 6 & 2 & 2 & 1 \\
\bottomrule
\end{tabular}
\end{table}

\subsubsection{Implementation Details}\label{subsubsec:implementation_details}

All deep learning models were implemented using PyTorch 2.4.1 and trained 
and tested on an NVIDIA GeForce RTX 4090 GPU. To ensure reproducibility 
and statistical significance, each deep learning model was trained with 
five different random seeds, and the final test results were averaged. 
Traditional methods (ESTARFM~\cite{zhuEnhancedSpatialTemporal2010a}, 
FSDAF~\cite{zhu2016flexible}, STARFM~\cite{fenggaoBlendingLandsatMODIS2006a}) 
were implemented using their 
official code with recommended parameter settings.

For deep learning methods, we employed the Adam optimizer with an initial 
learning rate of $1 \times 10^{-3}$. A step decay strategy was applied every 100 epochs 
to reduce the learning rate by a factor of 0.8. All models were trained 
for a maximum of 500 epochs. During training, input images were cropped 
into 512×512 patches, while a sliding window strategy was adopted to 
process complete images during testing.


\begin{table*}[!htbp]
{\scriptsize     
\setlength{\tabcolsep}{2pt}
\centering
\caption{Performance comparison of different models across all datasets.}
\label{tab:all_results}
\begin{tabular*}{\textwidth}{@{\extracolsep{\fill}} l l cccccccccc @{}}
\toprule
& \textbf{Model} & \scalebox{0.7}{\textbf{ECPWSTFN~\cite{zhang2024enhanced}}} & \scalebox{0.7}{\textbf{EDCSTFN~\cite{tan2019enhanced}}} & 
\scalebox{0.7}{\textbf{GANSTFM~\cite{tan2022flexible}}} & \scalebox{0.7}{\textbf{MLFFGAN~\cite{song2022mlffgan}}} & \scalebox{0.7}{\textbf{SRSFGAN~\cite{zhaoSRSFGANSuperResolutionBasedSpatial2023}}} & 
\scalebox{0.7}{\textbf{STFDiff~\cite{huang2024stfdiff}}} & \scalebox{0.7}{\textbf{SwinSTFM~\cite{chen2022swinstfm}}} & \scalebox{0.7}{\textbf{ESTARFM~\cite{zhuEnhancedSpatialTemporal2010a}}} & 
\scalebox{0.7}{\textbf{FSDAF~\cite{zhu2016flexible}}} & \scalebox{0.7}{\textbf{STARFM~\cite{fenggaoBlendingLandsatMODIS2006a}}} \\
\midrule

\multirow{7}{*}{\rotatebox{90}{\textbf{CIA}}} 
& RMSE & 0.0233 & 0.0211 & 0.0253 & \textbf{0.0201} & 0.0221 & 0.0205 & 
0.0207 & 0.0202 & 0.0216 & 0.0219 \\
& R$^2$ & 0.9059 & 0.7363 & 0.8856 & \textbf{0.9283} & 0.9140 & 0.9182 & 
0.9245 & 0.9259 & 0.9169 & 0.9131 \\
& SSIM & 0.8673 & \textbf{0.8798} & 0.8587 & 0.8735 & 0.8735 & 
0.8785 & 0.8658 & 0.8661 & 0.8608 & 0.8549 \\
& SAM & 0.0860 & 0.0722 & 0.0932 & \textbf{0.0707} & 0.0808 & 0.0727 & 
0.0727 & 0.0797 & 0.0751 & 0.0784 \\
& PSNR & 30.679 & 31.367 & 29.813 & \textbf{31.801} & 31.050 & 
31.285 & 31.613 & 31.631 & 31.193 & 31.020 \\
& ERGAS & 0.3218 & 0.2835 & 0.3433 & \textbf{0.2638} & 0.3003 & 0.2783 & 
0.2761 & 0.2735 & 0.2897 & 0.2975 \\
& Time(s) & 0.6417 & 0.1644 & \textbf{0.0090} & 0.0134 & 0.0118 & 
1.3100 & 0.0874 & 2.8657 & 3.5054 & 3.7456 \\
\midrule

\multirow{7}{*}{\rotatebox{90}{\textbf{LGC}}} 
& RMSE & 0.0251 & 0.0177 & 0.0261 & 0.0165 & 0.0176 & 0.0173 & 
\textbf{0.0160} & 0.0163 & 0.0174 & 0.0178 \\
& R$^2$ & 0.8563 & 0.9293 & 0.8348 & 0.9409 & 0.9323 & 0.9342 & 
\textbf{0.9440} & 0.9353 & 0.9285 & 0.9244 \\
& SSIM & 0.9085 & 0.9278 & 0.9024 & 0.9280 & 0.9284 & \textbf{0.9314} & 
0.9302 & 0.9296 & 0.9222 & 0.9161 \\
& SAM & 0.0926 & 0.0672 & 0.1019 & 0.0628 & 0.0675 & 0.0610 & 
\textbf{0.0573} & 0.0663 & 0.0614 & 0.0649 \\
& PSNR & 30.280 & 32.986 & 29.507 & 33.729 & 33.143 & 
33.363 & \textbf{34.028} & 33.867 & 33.093 & 32.814 \\
& ERGAS & 0.3150 & 0.2236 & 0.3279 & 0.2052 & 0.2221 & 0.2204 & 
\textbf{0.2022} & 0.2187 & 0.2100 & 0.2227 \\
& Time(s) & 0.4424 & \textbf{0.0016} & 0.0084 & 0.0129 & 0.0113 & 
1.3056 & 0.0887 & 17.1251 & 17.7439 & 18.2814 \\
\midrule

\multirow{7}{*}{\rotatebox{90}{\textbf{DX}}} 
& RMSE & 0.0490 & 0.0317 & 0.0429 & \textbf{0.0292} & 0.0312 & 
0.0314 & 0.0304 & 0.0375 & 0.0340 & 0.0340 \\
& R$^2$ & 0.7171 & 0.8894 & 0.7984 & \textbf{0.9054} & 0.8921 & 
0.8892 & 0.9009 & 0.8183 & 0.8697 & 0.8683 \\
& SSIM & 0.7126 & 0.7752 & 0.7298 & 0.7805 & \textbf{0.7806} & 
0.7687 & 0.7627 & 0.7503 & 0.7395 & 0.7436 \\
& SAM & 0.1896 & 0.1141 & 0.1561 & \textbf{0.1012} & 0.1102 & 
0.1117 & 0.1087 & 0.1297 & 0.1213 & 0.1213 \\
& PSNR & 25.364 & 28.699 & 26.429 & \textbf{29.352} & 28.756 & 
28.703 & 29.117 & 27.500 & 28.102 & 28.040 \\
& ERGAS & 0.6779 & 0.4241 & 0.5806 & \textbf{0.3964} & 0.4111 & 
0.4297 & 0.4131 & 0.5087 & 0.4577 & 0.4505 \\
& Time(s) & 0.0092 & \textbf{0.0017} & 0.0092 & 0.0130 & 0.0116 & 
1.3059 & 0.0875 & 5.0935 & 5.8040 & 5.3991 \\
\midrule

\multirow{7}{*}{\rotatebox{90}{\textbf{TJ}}} 
& RMSE & 0.0382 & \textbf{0.0343} & 0.0531 & 0.0360 & 0.0345 & 
0.0359 & \textbf{0.0343} & 0.0396 & 0.0440 & 0.0443 \\
& R$^2$ & 0.8250 & 0.8607 & 0.6315 & 0.8521 & 0.8592 & 0.8450 & 
\textbf{0.8646} & 0.7809 & 0.7766 & 0.7742 \\
& SSIM & 0.7487 & 0.7707 & 0.7012 & 0.7431 & \textbf{0.7719} & 
0.7489 & 0.7590 & 0.7392 & 0.7011 & 0.6987 \\
& SAM & 0.1553 & 0.1342 & 0.2251 & 0.1357 & 0.1313 & 
\textbf{0.1308} & 0.1359 & 0.1611 & 0.1859 & 0.1874 \\
& PSNR & 27.099 & 27.857 & 24.121 & 27.619 & 27.783 & 
27.470 & \textbf{27.998} & 26.832 & 26.114 & 26.049 \\
& ERGAS & 0.5760 & 0.5145 & 0.7621 & 0.5830 & \textbf{0.5120} & 
0.5544 & 0.5372 & 0.5546 & 0.6706 & 0.6714 \\
& Time(s) & 0.5859 & \textbf{0.0016} & 0.0092 & 0.5830 & 0.0113 & 
1.3062 & 0.0876 & 6.7969 & 7.9555 & 7.3832 \\
\midrule

\multirow{7}{*}{\rotatebox{90}{\textbf{WH}}} 
& RMSE & 0.0282 & 0.0296 & 0.0325 & 0.0380 & 0.0418 & 0.0301 & 
0.0454 & \textbf{0.0170} & 0.0240 & 0.0247 \\
& R$^2$ & 0.7355 & 0.7283 & 0.6485 & 0.6123 & 0.3856 & 0.7193 & 
0.5035 & \textbf{0.8864} & 0.8105 & 0.7875 \\
& SSIM & 0.8175 & 0.8125 & 0.7628 & 0.7726 & 0.7222 & 0.7889 & 
0.5557 & \textbf{0.9237} & 0.8704 & 0.8614 \\
& SAM & 0.1418 & 0.1261 & 0.1910 & 0.1525 & 0.2611 & 0.1313 & 
0.2261 & \textbf{0.0887} & 0.1215 & 0.1362 \\
& PSNR & 29.126 & 29.095 & 27.734 & 27.623 & 25.517 & 
28.911 & 26.325 & \textbf{32.697} & 30.456 & 30.081 \\
& ERGAS & 0.5687 & 0.5871 & 0.6293 & 0.7693 & 0.8160 & 0.6189 & 
0.9397 & \textbf{0.3748} & 0.4877 & 0.4961 \\
& Time(s) & 0.1821 & 0.1101 & 0.0635 & \textbf{0.0230} & 0.1258 & 
1.4497 & 0.2681 & 0.4997 & 0.5471 & 0.5162 \\
\midrule

\multirow{7}{*}{\rotatebox{90}{\textbf{BC}}} 
& RMSE & 0.0466 & 0.0495 & 0.0589 & 0.0440 & 0.0635 & 0.0489 & 
0.0468 & 0.0417 & 0.0303 & \textbf{0.0296} \\
& R$^2$ & 0.8286 & 0.8013 & 0.7148 & 0.8438 & 0.7005 & 0.8093 & 
0.8422 & 0.8198 & 0.9173 & \textbf{0.9238} \\
& SSIM & 0.7624 & 0.7233 & 0.6517 & 0.7378 & 0.7365 & 0.7230 & 
0.5892 & 0.8137 & 0.8455 & \textbf{0.8506} \\
& SAM & 0.1508 & 0.1665 & 0.2119 & 0.1255 & 0.2105 & 0.1717 & 
0.1584 & 0.1475 & 0.0860 & \textbf{0.0835} \\
& PSNR & 25.458 & 25.167 & 23.794 & 25.752 & 22.002 & 
24.877 & 25.296 & 26.804 & 28.241 & \textbf{28.748} \\
& ERGAS & 0.6862 & 0.7281 & 0.8935 & 0.6081 & 0.9433 & 0.7406 & 
0.6786 & 0.6165 & \textbf{0.4203} & 0.4260 \\
& Time(s) & 0.5487 & 0.0331 & \textbf{0.0234} & 0.0291 & 0.0272 & 
1.3008 & 0.0981 & 2.2444 & 2.2092 & 2.1577 \\
\midrule

\multirow{7}{*}{\rotatebox{90}{\textbf{IC}}} 
& RMSE & 0.0399 & 0.0365 & 0.0566 & 0.0339 & 0.0596 & 0.0387 & 
0.0398 & 0.0440 & 0.0329 & \textbf{0.0326} \\
& R$^2$ & 0.7993 & 0.8262 & 0.5883 & 0.8626 & 0.5602 & 0.8000 & 
0.8302 & 0.6772 & \textbf{0.8690} & 0.8639 \\
& SSIM & 0.8526 & 0.8718 & 0.7899 & 0.8518 & 0.7912 & 0.8717 & 
0.7161 & 0.8640 & 0.8682 & \textbf{0.8756} \\
& SAM & 0.1105 & 0.1105 & 0.2128 & \textbf{0.1003} & 0.2117 & 
0.1157 & 0.1375 & 0.1566 & 0.1048 & 0.1039 \\
& PSNR & 25.690 & 26.318 & 22.587 & 27.218 & 22.844 & 
25.706 & 26.260 & 23.652 & \textbf{27.378} & 27.216 \\
& ERGAS & 0.4181 & 0.3909 & 0.6594 & 0.3525 & 0.6583 & 0.4294 & 
0.4169 & 0.5076 & \textbf{0.3502} & 0.3557 \\
& Time(s) & 0.8680 & \textbf{0.0041} & 0.0138 & 0.0181 & 0.0164 & 
1.3030 & 0.0907 & 2.1966 & 2.3963 & 2.1117 \\

\bottomrule
\end{tabular*}
}
\end{table*}

\subsection{Comparison Methods}\label{subsec:comparison_methods}

We compared ten representative spatiotemporal fusion methods, 
covering both traditional and deep learning approaches: 
(a) \textbf{Deep Learning Methods}: ECPW-STFN~\cite{zhang2024enhanced}, 
EDCSTFN~\cite{tan2019enhanced}, GAN-STFM~\cite{tan2022flexible}, 
MLFF-GAN~\cite{song2022mlffgan}, SRSF-GAN~\cite{zhaoSRSFGANSuperResolutionBasedSpatial2023}, 
STFDiff~\cite{huang2024stfdiff}, SwinSTFM~\cite{chen2022swinstfm}; 
(b) \textbf{Traditional Methods}: ESTARFM~\cite{zhuEnhancedSpatialTemporal2010a}, 
FSDAF~\cite{zhu2016flexible}, STARFM~\cite{fenggaoBlendingLandsatMODIS2006a}.

\subsection{Evaluation Metrics}\label{subsec:evaluation_metrics}

We employed seven complementary metrics to comprehensively evaluate 
fusion quality: Root Mean Square Error (RMSE) for pixel-level accuracy, 
Coefficient of determination (R²) for overall fitting quality, 
Structural Similarity Index (SSIM) for perceptual quality, 
Spectral Angle Mapper (SAM) for spectral fidelity, 
Peak Signal-to-Noise Ratio (PSNR) for reconstruction quality, 
Relative Dimensionless Global Error (ERGAS), and average inference 
time per image (Time) for algorithm efficiency assessment.

\subsection{Results and Analysis}\label{subsec:results_analysis}

\subsubsection{Overall Performance Comparison}
\label{subsubsec:overall_performance}

\autoref{fig:cia_visual_comparison}, \autoref{fig:lgc_visual_comparison}, 
and \autoref{fig:dx_visual_comparison} present the visual comparison 
results of different methods on the CIA~\cite{zhang2024enhanced}, LGC~\cite{ma2024conditional}, 
and DX~\cite{li2020overview} datasets, 
respectively. In each figure, the first and third rows show the ground 
truth images and predicted false color composite images, while the 
second and fourth rows display the corresponding difference maps. 
From left to right: (a) ground truth high-resolution (HR) image HR $\textit{t}_2$,
(b) low-resolution (LR) input image LR $\textit{t}_2$, and prediction results from 
various methods: (c) ECPW-STFN~\cite{zhang2024enhanced}, (d) EDCSTFN~\cite{tan2019enhanced}, 
(e) GAN-STFM~\cite{tan2022flexible}, (f) MLFF-GAN~\cite{song2022mlffgan}, 
(g) SRSF-GAN~\cite{zhaoSRSFGANSuperResolutionBasedSpatial2023}, (h) STFDiff~\cite{huang2024stfdiff}, 
(i) SwinSTFM~\cite{chen2022swinstfm}, (j) ESTARFM~\cite{zhuEnhancedSpatialTemporal2010a}, 
(k) FSDAF~\cite{zhu2016flexible}, (l) STARFM~\cite{fenggaoBlendingLandsatMODIS2006a}. 
White ellipses highlight key comparison regions. It can be observed that deep 
learning methods generally outperform traditional methods in preserving 
spatial details and reducing differences, particularly in building edges and 
agricultural textures within the highlighted regions. Visual comparison 
results for the remaining datasets (TJ~\cite{li2020overview}, WH~\cite{zhang2024dcdgan}, 
BC~\cite{guo2024obsum}, and IC~\cite{guo2024obsum}) are provided 
in the appendix.

\autoref{tab:all_results} presents the comprehensive performance comparison 
across all datasets. It should be noted that the metrics shown in 
this table represent the average performance indicators 
across all test image pairs. Detailed performance metrics for individual 
test image pairs on each dataset are provided in the appendix to allow 
readers to understand the stability and variation of method performance.

\noindent\textbf{Traditional Method Performance:} Among traditional methods, 
ESTARFM~\cite{zhuEnhancedSpatialTemporal2010a} maintains strong competitiveness across various datasets, particularly 
excelling on less commonly used datasets. STARFM~\cite{fenggaoBlendingLandsatMODIS2006a} also demonstrates robust 
performance in specific scenarios, achieving optimal results on certain 
evaluation metrics. This indicates that traditional methods retain their 
value despite the emergence of deep learning approaches.

\noindent\textbf{Deep Learning Advantages:} Deep learning methods show 
significant advantages on widely-used spatiotemporal fusion datasets. 
MLFF-GAN~\cite{song2022mlffgan} and SwinSTFM~\cite{chen2022swinstfm} consistently achieve excellent results on benchmark 
datasets like CIA and LGC, where abundant training data and established 
evaluation protocols favor neural network approaches. These methods benefit 
from extensive optimization on such standard benchmarks.

\noindent\textbf{Architecture-Specific Advantages:} CNN-based methods like 
EDCSTFN~\cite{tan2019enhanced} and ECPW-STFN~\cite{zhang2024enhanced} demonstrate exceptional computational efficiency while 
maintaining competitive accuracy. GAN-based approaches excel in preserving 
fine spatial details and generating visually coherent results. Transformer-based 
SwinSTFM~\cite{chen2022swinstfm} leverages attention mechanisms for comprehensive feature modeling. 
Diffusion-based STFDiff~\cite{huang2024stfdiff}, despite higher computational requirements, shows 
promise in specific evaluation metrics.

\subsubsection{Dataset Specific Analysis}
\label{subsubsec:dataset_specific_analysis}

\noindent\textbf{Commonly-Used Datasets (CIA~\cite{zhang2024enhanced}, LGC~\cite{ma2024conditional}):} On these frequently 
utilized benchmarks in spatiotemporal fusion research, deep learning methods 
dominate the performance rankings. The abundance of research and optimization 
efforts on these datasets has enabled neural networks to learn effective 
representations, resulting in superior performance across most metrics.

\noindent\textbf{Less Common Datasets (WH~\cite{zhang2024dcdgan}, BC~\cite{guo2024obsum}, IC~\cite{guo2024obsum}):} Interestingly, traditional 
methods like ESTARFM~\cite{zhuEnhancedSpatialTemporal2010a} and STARFM~\cite{fenggaoBlendingLandsatMODIS2006a} achieve remarkable results on datasets that 
are less frequently used in the spatiotemporal fusion community. This 
phenomenon suggests that deep learning models may suffer from limited 
generalization when applied to data distributions they were not extensively 
trained on, while traditional methods' physics-based assumptions remain 
universally applicable.

\noindent\textbf{Balancing Accuracy and Generalization:} While achieving high 
accuracy on standard benchmarks like CIA~\cite{zhang2024enhanced} and LGC~\cite{ma2024conditional} remains important for 
demonstrating method effectiveness, we encourage researchers to avoid 
over-specialization to these datasets. Generalization capability across 
diverse scenarios is equally crucial for determining algorithm quality and 
practical applicability. An ideal approach would balance both in-domain 
performance and cross-dataset robustness, ensuring that proposed methods 
can adapt to varied real-world conditions. Future developments should 
consider incorporating diverse evaluation scenarios alongside traditional 
benchmarks to foster more generalizable spatiotemporal fusion solutions.

\subsubsection{Computational Efficiency Analysis}
\label{subsubsec:computational_efficiency}

The computational landscape reveals a clear divide: modern deep learning 
methods, particularly lightweight CNNs, achieve inference times in the 
millisecond range, while traditional pixel-wise methods require orders of 
magnitude more time. This efficiency gap becomes critical for real-time 
applications and large-scale processing tasks. Deep learning methods benefit 
significantly from GPU acceleration, leveraging parallel processing capabilities 
to handle entire image regions simultaneously through tensor operations that 
align perfectly with modern GPU architectures. In contrast, traditional methods 
like STARFM~\cite{fenggaoBlendingLandsatMODIS2006a} and ESTARFM~\cite{zhuEnhancedSpatialTemporal2010a} rely on iterative pixel-wise calculations that cannot 
fully exploit parallel computing resources. Additionally, deep learning models 
support efficient batch processing with minimal overhead, making them ideal 
for operational systems.

\subsection{Experimental Summary}
\label{subsec:experimental_summary}

Our comprehensive evaluation reveals that no single approach dominates across 
all scenarios. Deep learning methods excel on well-established benchmarks 
where they benefit from extensive optimization, while traditional methods 
maintain advantages on less common datasets where their model-free assumptions 
prove more robust. This suggests that the spatiotemporal fusion community 
should consider dataset characteristics when selecting methods.

The performance variations across datasets highlight important considerations 
for future research. Over-reliance on standard benchmarks may lead to methods 
that generalize poorly to new scenarios. The strong performance of traditional 
methods on certain datasets indicates that incorporating domain knowledge 
remains valuable. Future work should focus on developing adaptive frameworks 
that combine the efficiency of deep learning with the robustness of traditional 
approaches, ultimately creating more versatile spatiotemporal fusion solutions.

\section{Challenges}\label{sec:challenges}
This section examines various challenges encountered during the development and application of remote sensing 
spatiotemporal fusion technologies. Breaking through these limitations requires researchers to innovate existing technologies, 
develop deeper understanding of these problems, and provide solutions that support further advancement in spatiotemporal 
fusion applications.

\subsection{Conflict Between Time and Space as Opposites}

In the realm of remote sensing image fusion, mismatched temporal and spatial resolutions between 
sensors lead to fundamental conflicts in data harmonization. Inconsistencies in temporal resolution 
can lead to deviations in the fusion results along the time dimension, particularly in areas experiencing 
rapid surface changes, such as vegetation growth and urban expansion. In such cases, the fusion outcomes may 
fail to accurately capture these dynamic changes. Similarly, discrepancies in spatial resolution can result in 
errors in the spatial dimension, which is especially critical for applications requiring high spatial precision, 
such as urban mapping and agricultural monitoring. Beyond the fundamental conflicts arising from mismatched temporal 
and spatial resolutions, several other issues contribute to the complexity of spatiotemporal fusion. These include 
mismatched temporal scales, asynchronous data acquisition, differences in sensor characteristics (\eg, spectral bands, 
viewing geometry, radiometric calibration), incompatibilities in data processing methods (\eg, radiometric correction, 
geometric correction, and atmospheric correction), and surface heterogeneity, \etc In essence, harmonizing time-space 
conflict requires a holistic approach that not only optimizes the trade-off between temporal and spatial resolutions 
but also accounts for the entire data lifecycle—from acquisition and preprocessing to fusion, validation, and ultimately 
the production of high-quality fusion products for application.

\subsection{Generalization of Deep Models}

While numerous studies report impressive performance metrics on specific datasets, very few systematically evaluate 
how these models perform when applied to entirely new datasets or satellite pairs without re-training or fine-tuning. 
Deep models possess inherent limitations, particularly in terms of poor generalizability. Models that perform exceptionally well 
within their training distribution often experience significant performance degradation when confronted with data from different 
geographical regions, seasons, or imaging conditions. The generalization challenge extends beyond spectral differences to include 
variations in spatial resolution ratios, temporal gaps, and regional landscape characteristics~\cite{liuReviewRemoteSensing2021}. 
A model trained on agricultural landscapes may fail to generalize to urban environments. Similarly, models optimized for specific 
seasonal transitions might perform poorly during other seasonal changes. These limitations highlight the need for more robust 
approaches that can maintain performance across diverse scenarios. Furthermore, current deep architectures for spatiotemporal 
fusion typically operate with fixed input band configurations. This architectural rigidity means they cannot process data with 
different channel inputs than those used during training. This further restricts the practical application capabilities of these 
models, leading to a complete loss of generalization ability.

\subsection{Large Datasets at Scale}

The advancement of deep learning in remote sensing spatiotemporal fusion is significantly constrained by the limited 
scale and diversity of available datasets. Unlike the computer vision field, which benefits from large-scale and diverse 
datasets like ImageNet~\cite{ImageNet} and COCO~\cite{COCO} that provide extensive coverage and rich annotations, remote sensing 
spatiotemporal fusion datasets often fall short. These datasets typically lack the volume and variety needed to effectively train 
deep learning models that can generalize well across different regions and conditions. For example, while ImageNet~\cite{ImageNet} 
contains millions of images across thousands of categories, supporting robust feature learning, remote sensing datasets such as 
Landsat and MODIS, though valuable, offer fewer samples and less diversity in terms of geographic coverage, environmental conditions, 
and sensor types. This limitation results in models that perform well on specific datasets but struggle when applied to new or varied 
scenarios. Additionally, the high cost and complexity of acquiring and processing remote sensing data further hinder the creation of larger 
and more diverse datasets. The absence of comprehensive benchmarks comparable to those in computer vision limits the ability of researchers 
to fully leverage the potential of deep learning, emphasizing the urgent need for more extensive and varied datasets to drive innovation 
in spatiotemporal fusion technology.

\minew{
Beyond dataset scale limitations, the practical application of deep learning models in remote sensing spatiotemporal fusion also faces 
challenges in processing image sizes. In conventional computer vision tasks, the image sizes used are relatively small 
(such as $224 \times 224$ pixels), but remote sensing applications require larger image dimensions to preserve spatial 
details and contextual information~\cite{kimResolutionAwareDesignAtrous2023}. However, deep learning spatiotemporal fusion models often work with constrained 
image patch sizes due to computational limitations. For instance, MLFF-GAN processes images of $256 \times 256$ pixels 
with six bands during training~\cite{song2022mlffgan}, while StfNet tailors input images into tiles of $250 \times 250$ 
pixels for prediction, with adjacent tiles having overlaps to avoid boundary artifacts~\cite{liu2019stfnet}. These 
size constraints can significantly impact model performance, as smaller patches may lose important contextual information 
and spatial relationships that are crucial for accurate spatiotemporal fusion. The trade-off between computational 
feasibility and preservation of spatial context remains a key challenge, requiring careful consideration of patch size 
selection and overlap strategies to maintain fusion quality while ensuring practical deployment.
}

\subsection{Multi-source Heterogeneous Fusion}
The diversification of technologies and observation platforms has led to remote sensing data from different sources exhibiting 
significant heterogeneity across spatial resolution, spectral range, band configuration, and viewing 
angles~\cite{li2022deeplearning,huang2017research,sdraka2022deep,niu2023heterogeneous}. 
These differences pose substantial barriers to effective information integration. During practical spatiotemporal fusion operations, 
even spectrally similar images generate unavoidable pixel-level disparities due to variations in spatial resolution, band width, 
and atmospheric interference. 
With the recent emergence of drone technology, attempts to merge drone imagery with satellite imagery have increased. However, 
the spatial scale and observational condition differences between these cross-scale images far exceed those among data from similar 
platforms, further intensifying the complexity of multi-source heterogeneous 
data fusion~\cite{zhuSpatiotemporalFusionMultisource2018,belgiu2019spatiotemporal,huangEvaluationPlanetScopeImages2022}. 
Addressing the proper integration of multi-modal, multi-scale data has become an urgent challenge in remote sensing data processing tasks.

\minew{
Recent years have witnessed the emergence of several innovative solutions for multi-scale data fusion, exemplified by 
the Spatiotemporal Random Forest (STRF) method~\cite{wangFillingGapsCloudy2024}. This approach incorporates 2km-resolution 
Geostationary Operational Environmental Satellite (GOES) land surface temperature data as auxiliary information, performing multi-scale spatiotemporal data fusion with 100m-resolution 
Landsat data, thereby effectively addressing the limitations of traditional fusion methods when handling large gaps and dramatic 
change scenarios~\cite{belgiu2019spatiotemporal,zhuSpatiotemporalFusionMultisource2018}. STRF leverages spatially complete 
high-resolution imagery at known time points to establish a baseline, subsequently integrating coarse-resolution auxiliary 
observations from different timestamps to capture temporal variation patterns, and ultimately exploiting the model's inherent 
nonlinear mapping capabilities to achieve effective integration of cross-scale information.

Nevertheless, current multi-scale fusion approaches still encounter numerous challenges. When the spatial scale of target changes 
is smaller than the pixel size of coarse-resolution auxiliary data, models often struggle to capture fine-scale variation information; 
systematic biases and data quality discrepancies between different sensors may introduce additional uncertainties; furthermore, 
the spatial structural consistency requirements between training and prediction regions constrain the applicability of 
these methods~\cite{wangFillingGapsCloudy2024}. Future research directions may focus on several promising avenues: developing 
more sophisticated cross-sensor calibration models to mitigate systematic biases; incorporating uncertainty quantification 
mechanisms to enhance reliability assessment of fusion results; and exploring adaptive fusion frameworks that integrate deep learning 
techniques to strengthen adaptability to complex scenarios.
}

\subsection{Computational Efficiency}
In practical applications, spatiotemporal fusion technology faces data processing speed bottlenecks~\cite{wu2021land}. 
Most fusion methods rely on pixel-level calculations or moving window strategies, which, though accurate, are too slow. 
For large-scale imagery, personal computers can take hours to a full day, which is adequate for small-scale 
areas but insufficient for larger regions or global tasks~\cite{zhuSpatiotemporalFusionMultisource2018}. 

\minew{
To improve efficiency, one can optimize computational unit granularity to balance pixel-level and image-level calculations and employ 
more efficient frameworks such as GPU parallel acceleration or cloud computing~\cite{zhuSpatiotemporalFusionMultisource2018, belgiu2019spatiotemporal}.
Reducing model complexity is also a feasible approach, which can be specifically referenced from the cuFSDAF model~\cite{gaoCuFSDAFEnhancedFlexible2022b}. Originally, 
the FSDAF algorithm employed Thin Plate Spline (TPS) interpolators with a time complexity of O($n^3$), resulting in relatively high complexity. 
The cuFSDAF model abandons the TPS interpolator and instead uses Inverse Distance Weighting (IDW) interpolators with O($n$) complexity, significantly reducing computational complexity 
while maintaining fusion accuracy. Furthermore, cuFSDAF parallelizes computation-intensive processes (such as interpolation, pixel homogeneity 
calculation, etc.) through Compute Unified Device Architecture (CUDA). CUDA distributes computational tasks across multiple 
GPU threads for processing, thereby greatly improving computational efficiency and enabling cuFSDAF to efficiently handle 
large-scale datasets while avoiding performance bottlenecks inherent in traditional CPU computing.
}

\section{Opportunities}\label{sec:opportunities}

\subsection{Balancing Data-driven and Model-driven Approaches}
Balancing data-driven and model-driven approaches stems from different emphases on generalization and 
precision in remote sensing image processing. As mentioned before, generalization refers to a model's ability to perform well on unfamiliar training 
data, essentially its adaptability or transferability~\cite{zhuSpatiotemporalFusionMultisource2018, liu2024semiblind}. Precision measures how well a model's predictions match actual results when 
processing known data. Data-driven methods like deep learning demonstrate exceptional self-adaptive capabilities, excelling at extracting 
complex non-linear features from remote sensing imagery, yet heavily depend on data scale and diversity. When datasets lack sufficient coverage, data-driven methods experience a significant decline in generalization ability for unknown scenarios. 
This leads to an unacceptable risk in practical remote sensing applications.
Conversely, model-driven approaches provide deeper understanding 
of the physical mechanisms behind imaging processes and exhibit stable generalization performance, but struggle with complex non-linear 
relationships in remote sensing images. Combining data-driven and model-driven methods represents 
a popular coupling approach that reduces deep learning networks' dependence on large-scale data while enhancing model interpretability and generalizability. 
For instance, in synthetic aperture radar (SAR) image despeckle tasks, variational models constrain deep networks, aligning learning 
processes with actual physical characteristics~\cite{shenCouplingModelDrivenDataDriven2022}. In hyperspectral image denoising, embedding model-driven constraints 
into network structures maintains effective generalization capabilities even with incomplete data~\cite{yangResearchProgressChallenges2022,shenCouplingModelDrivenDataDriven2022}. 
Nevertheless, the complexity and diversity of remote sensing data, coupled with inherent limitations in model-driven approaches' modeling 
capabilities, means this coupling technology rarely achieves optimal performance in practical applications. The form and degree of integration 
still require exploration and optimization by researchers.

\subsection{Unsupervised Learning and Self-supervised Learning}
In situations where remote sensing data is difficult to acquire or existing samples lack cleanliness, unsupervised learning proves valuable 
as it completely eliminates dependence on labeled data. A typical approach leverages GAN's unique adversarial loss and competitive 
training mechanism to capture high-level abstract features when reference images are insufficiently clean, achieving high-quality image 
restoration tasks~\cite{wangReviewPixellevelRemote2023}. Researchers can subsequently focus on integrating remote sensing observation 
models with unsupervised network structures, enhancing the robustness and generalization of corresponding methods to achieve more stable 
and reliable spatiotemporal fusion technology~\cite{zhuDeepLearningRemote2017,wangReviewPixellevelRemote2023}.

\minew{
Self-supervised learning has emerged as a promising approach to address the limitations of supervised learning in spatiotemporal fusion tasks. 
Unlike supervised methods that rely heavily on auxiliary date data and may perform poorly when significant spatiotemporal changes occur between 
auxiliary and prediction dates, self-supervised strategies can focus more on the characteristics of prediction dates during training. 
Sun et al. demonstrated the effectiveness of combining supervised and self-supervised learning strategies in a dual-stage cascade framework, 
where supervised learning extracts rich spatial features from original-scale data, while self-supervised learning excavates spatiotemporal 
features at prediction dates~\cite{sunSupervisedSelfsupervisedLearningbased2023b}. This innovative approach addresses the key challenge that 
traditional self-supervised methods typically train on down-sampled data with insufficient spatial information. Furthermore, the integration 
of temporal consistency loss functions that exploit temporal correlations between multiple prediction dates can significantly improve fusion 
accuracy. The self-supervised learning paradigm shows particular promise for remote sensing applications where temporal dynamics are critical, 
such as land surface temperature monitoring and vegetation phenology tracking, enabling more temporally accurate and spatially detailed fusion results.
}

\subsection{Potential of Multi-task Learning}
Multi-task Learning (MTL) offers the core advantage of effectively integrating associated features between multiple tasks, thereby 
improving overall processing efficiency and accuracy. This characteristic aligns perfectly with several remote sensing image processing 
requirements. For instance, remote sensing image fusion frequently encounters distribution feature differences caused by seasonal changes, 
while MTL can alleviate accuracy losses resulting from insufficient feature migration by sharing information between different tasks. 
When high-quality paired samples are scarce, MTL can also leverage its characteristics to address such data scarcity issues~\cite{chengMultitaskMultisourceDeep2020}.
In practical applications, MTL can configure image fusion tasks from different periods as a group of interconnected subtasks sharing 
feature spaces, achieving personalized processing of individual tasks while utilizing task synergies to enhance overall fusion performance. 
For example, treating the fusion of Landsat and MODIS images from different time points as interconnected prediction tasks effectively 
captures the continuity of seasonal changes and spatial details, enhancing spatiotemporal consistency in fused images~\cite{chengMultitaskMultisourceDeep2020,leiva-murilloMultitaskRemoteSensing2013}. Looking forward, researchers can further enhance model generalization and robustness by integrating MTL with deep learning frameworks, 
improving both processing efficiency and prediction accuracy.

\subsection{Foundation Models for Spatiotemporal Fusion}
Remote sensing foundation models (RSFMs) can integrate multi-source, multi-temporal, and multi-resolution remote sensing data through a unified framework to generate 
surface observation data with high spatial continuity, showing potential for spatiotemporal fusion tasks~\cite{xiaoFoundationModelsRemote2024}. 
SatMAE~\cite{cong2022satmae} can capture spatiotemporal dependencies in multispectral images through temporal mask reconstruction pretraining, 
effectively combining high and low-resolution images from different time periods. Scale-MAE~\cite{reedScaleMAEScaleAwareMasked2023} enhances 
detail consistency by combining multi-scale reconstruction and land feature distance encoding, fusing remote sensing data from different 
sensors. CROMA~\cite{fuller2023croma} extracts invariant spatiotemporal features through cross-modal contrastive learning, thereby analyzing 
heterogeneous data sources. Generative models similar to DiffusionSat~\cite{khannaDiffusionSatGenerativeFoundation2024} can utilize diffusion 
processes to reconstruct continuous dynamic land cover, cloud-contaminated or missing temporal images for practical applications. RSFMs are suitable for large-scale spatiotemporal analysis and can resolve spatiotemporal conflicts caused by sensor differences through 
multi-modal alignment capabilities, certainly applicable to specific spatiotemporal fusion tasks in the future~\cite{xiaoFoundationModelsRemote2024}.

\minew{
Large foundation models such as extensively pre-trained language models and vision-language models are being widely deployed 
in spatiotemporal prediction tasks, achieving remarkable results. Spatiotemporal forecasting tasks focus on domains including climate 
change, urban expansion, and traffic flow prediction~\cite{pathakFourCastNetGlobalDatadriven2022, jinLargeModelsTime2023, xueLeveragingLanguageFoundation2022}. 
Such tasks require processing complex temporal sequence data and spatial information. The prominent large models that have gained popularity 
in recent years can effectively learn spatiotemporal relationships by integrating data from different temporal nodes and spatial locations, 
thereby enabling precise predictions. Pangu-Weather~\cite{bi2023accurate} successfully modeled global climate through multimodal model 
combinations, significantly enhancing weather forecasting accuracy; the Climax model~\cite{nguyenClimaXFoundationModel2023} applied 
pre-trained deep learning networks to spatiotemporal analysis of climate data, successfully predicting variation trends across different 
climatic variables. These vivid examples undoubtedly demonstrate the potential of foundation models in spatiotemporal prediction 
tasks~\cite{maiOpportunitiesChallengesFoundation2024}.

Foundation models also possess tremendous potential in remote sensing spatiotemporal fusion tasks. Large vision-language models (VLFMs), 
such as OpenCLIP~\cite{Ilharco_Open_Clip_2021} and BLIP~\cite{liBLIPBootstrappingLanguageImage2022}, can combine remote sensing imagery 
with other geospatial data to effectively perform land cover classification and change detection. These models with powerful multimodal 
learning capabilities can efficiently fuse remote sensing images from different time periods with auxiliary data (such as climatic information), 
thereby enhancing the model's understanding and predictive capacity for surface changes. This indicates that large foundation models can extract 
valuable information from blurred, missing, or low-quality data under conditions of relatively low spatiotemporal resolution. This capability 
proves particularly crucial in large-scale remote sensing image prediction tasks.

Therefore, further investigating the application of large foundation models in remote sensing spatiotemporal fusion, exploring how 
they can better capture and integrate information across spatiotemporal dimensions, will bring significant 
breakthroughs to the advancement of remote sensing technology.
}

\minew{
\subsection{Hybrid Architecture Integration}

Hybrid models that combine different deep learning architectures can compensate for potential deficiencies that may exist in single 
architectures to some extent. The hybrid deep learning model proposed by Yang et al.~\cite{yangRobustHybridDeep2021a} ingeniously 
combines two different deep learning architectures: Super-Resolution Convolutional Neural Network (SRCNN)~\cite{dongImageSuperResolutionUsing2016} 
and Long Short-Term Memory networks (LSTM). Specifically, SRCNN is responsible for learning the spatial relationships between high and low-resolution 
images, recovering blurred spatial details through feature extraction, non-linear mapping, and reconstruction; while LSTM specializes in 
processing temporal sequence information, utilizing its gating mechanism to learn complex phenological change patterns. This combined 
design enables the model to achieve higher prediction accuracy when dealing with rapid phenological change scenarios compared to single-architecture models.

In the future, researchers can focus on these development directions: (1) Hierarchical architecture design: decompose complex 
problems into spatial and temporal dimensions according to specific task requirements, and adopt the most suitable network 
architectures respectively (e.g., CNN for spatial information processing, RNN/LSTM for time series processing); (2) Progressive 
information transfer: design reasonable data flows so that the output of the previous stage serves as the input of the next stage, 
ensuring effective information transfer between different architectures; (3) Scenario-based evaluation strategies: establish test 
scenarios with different levels of change (e.g., rapid, moderate, minimal changes) to systematically evaluate the robustness of 
hybrid models under various conditions.
}

\minew{
\subsection{Multimodal Data Fusion}

Recent advances in multimodal fusion techniques have demonstrated significant 
potential for enhancing spatiotemporal analysis capabilities. 
Choi et al.~\cite{choiAttentionBasedMultimodalImage2022} developed 
attention-based fusion modules that effectively integrate complementary 
spectral information from different imaging modalities, providing valuable 
insights for remote sensing applications.

The integration of multi-spectral observations offers promising opportunities 
to address inherent limitations in single-modal spatiotemporal fusion. 
Different spectral bands capture distinct surface properties and respond 
differently to environmental conditions, enabling more robust monitoring 
across varying temporal and spatial scales. For instance, combining visible 
spectrum observations with thermal measurements can enhance land cover 
classification accuracy and improve change detection capabilities in 
dynamic environments.

Future research directions include: (1) developing adaptive fusion 
architectures that leverage attention mechanisms for multi-temporal and 
multi-spectral remote sensing data; (2) creating universal frameworks 
compatible with diverse sensor configurations and observation platforms; 
(3) exploring integration with emerging data sources such as LiDAR and 
hyperspectral imagery to expand the scope of spatiotemporal fusion 
applications.

\subsection{Integration with 3D Reconstruction and Multi-dimensional Analysis}

Recent advances in satellite constellation technology present new opportunities 
for spatiotemporal fusion through 3D reconstruction capabilities. The emergence 
of high-frequency imaging systems like PlanetScope, with over 130 satellites 
providing daily global coverage at 3-5m resolution, enables multi-view stereo 
(MVS) reconstruction for generating digital surface models (DSMs) with 4-6m 
vertical accuracy~\cite{huangEvaluationPlanetScopeImages2022, nohAnalysisPlanetScopeDove2023}. This capability 
opens unprecedented possibilities for spatiotemporal fusion by incorporating 
elevation information as an additional constraint in the fusion process. The 
integration of 2D spectral-temporal fusion with 3D geometric reconstruction 
can significantly enhance change detection accuracy, particularly for 
volumetric analysis in disaster monitoring, urban development tracking, and 
ecological assessment. Future research directions include developing 
multi-dimensional fusion frameworks that simultaneously optimize spatial, 
temporal, and elevation dimensions, leveraging the abundant multi-view imagery 
from modern satellite constellations to create more comprehensive Earth 
observation products~\cite{huangEvaluationPlanetScopeImages2022, zhuSpatiotemporalFusionMultisource2018}.

}

\section{Conclusion}\label{sec:conclusion}
\minew{This survey provides the first comprehensive review of deep learning applications 
in remote sensing spatiotemporal fusion over the past decade, establishing a 
systematic taxonomy of five main architectures: CNNs, Transformers, GANs, diffusion 
models, and sequence models. Our analysis reveals that CNN-based approaches dominate 
spatial feature extraction, while Transformer architectures excel in modeling 
long-range temporal dependencies, and GAN/diffusion models demonstrate superior 
generative capabilities for high-quality image synthesis, substantially outperforming 
traditional methods. Through systematic experiments on seven benchmark datasets, we 
empirically validate these observations and provide quantitative comparisons of ten 
representative methods. We identify five critical challenges limiting current progress: 
the inherent time-space conflict, poor cross-dataset generalization, computational 
inefficiency for large-scale processing, multi-source heterogeneous fusion 
difficulties, and insufficient benchmark diversity. 

Based on our comprehensive analysis, we highlight promising future opportunities 
including the integration of data-driven and model-driven approaches, self-supervised 
learning methods to address data scarcity, multi-task learning frameworks, emerging 
foundation models for large-scale analysis, hybrid architectures combining multiple 
paradigms, and multimodal data integration. As deep learning continues to mature, 
we anticipate the development of more robust, generalizable, and computationally 
efficient spatiotemporal fusion systems that will enable real-time global-scale 
monitoring capabilities, supporting critical applications in climate change research, 
disaster response, agricultural management, and urban planning. We hope this 
comprehensive survey serves as a valuable resource for researchers and practitioners, 
providing both a solid foundation for understanding current capabilities and clear 
guidance for future innovations in remote sensing spatiotemporal fusion technologies.}

\section*{Declaration of Competing Interest}
The authors declare that they have no known competing financial interests or personal relationships that could have appeared 
to influence the work reported in this paper.


\printcredits

\bibliographystyle{model1-num-names}
\bibliography{cas-refs}

\end{sloppypar}
\end{document}